\documentclass{article}

\usepackage[margin=2.5cm]{geometry}

\usepackage{tikz}
\usetikzlibrary{arrows.meta, positioning, shapes.geometric}

\usepackage{amsmath,amssymb,amsfonts,mathtools,mathrsfs,bm}
\usepackage{tabularx,booktabs,array,longtable,multirow,threeparttable}
\usepackage{graphicx,epsfig,subfigure}
\usepackage{color,,colortbl}
\usepackage{caption}
\usepackage{rotating}
\usepackage{diagbox}
\usepackage{pdflscape,lscape}
\usepackage{float}
\usepackage{longtable}
\usepackage[dvipsnames]{xcolor}

\usepackage{cite,url,hyperref}
\hypersetup{colorlinks=true, linkcolor=blue, urlcolor=blue, citecolor=blue}

\usepackage{algorithm,algpseudocode}

\usepackage{pifont}
\usepackage{tcolorbox}
\tcbuselibrary{skins}
\usepackage{multicol}

\definecolor{lightblue}{RGB}{70,130,180} 
\hypersetup{
  colorlinks=true,
  linkcolor=lightblue,
  urlcolor=lightblue,
  citecolor=lightblue
}

\newcommand{\RMSection}[2]{\textbf{\hyperref[#2]{#1}}\par}
\newcommand{\RMSub}[2]{\hspace*{1.5em}{\color{gray}\footnotesize$\triangleright$}~\hyperref[#2]{#1}\par}

\newtheorem{theorem}{Theorem}
\newtheorem{corollary}{Corollary}

\newcommand{\red}[1]{{\color{red} #1 }}

\title{A Practitioner's Guide to Kolmogorov--Arnold Networks}

\author{
Amir Noorizadegan$^{1,*}$, Sifan Wang$^{2}$, Leevan Ling$^{1}$, and Juan P. Dominguez-Morales$^{3}$ \\[4pt]
{\small $^{1}$Department of Mathematics, Hong Kong Baptist University, Hong Kong SAR, China} \\
{\small $^{2}$Institution for Foundations of Data Science, Yale University, New Haven, CT 06520, USA}\\
{\small $^{3}$Robotics and Technology of Computers Lab., Universidad de Sevilla, Av. Reina Mercedes s/n, Sevilla 41012, Spain} \\[6pt]
{\small *Corresponding author: \texttt{amir\_noori@hkbu.edu.hk}}
}

\begin{document}
\maketitle

\begin{abstract}
\noindent
Kolmogorov--Arnold Networks (KANs), whose design is inspired—rather than dictated—
by the Kolmogorov superposition theorem, have emerged as a
structured alternative to  MLPs.
This review provides a systematic and comprehensive overview of the rapidly
expanding KAN literature.
The review is organized around three core themes:
(i) clarifying the relationships between \textbf{KANs} and
\textbf{Kolmogorov superposition theory (KST)}, \textbf{MLPs}, and
\textbf{classical kernel methods};
(ii) analyzing \textbf{basis functions} as a \textbf{central design axis}; and
(iii) summarizing recent advances in \textbf{accuracy}, \textbf{efficiency},
\textbf{regularization}, and \textbf{convergence}.

\noindent
Finally, we provide a practical \textbf{``Choose--Your--KAN'' guide} and outline
\textbf{open research challenges and future directions}.
The accompanying \textbf{GitHub repository}
(\url{https://github.com/AmirNoori68/kan-review})
serves as a \textbf{structured reference} for ongoing KAN research.

\noindent\small
The following compact, \emph{clickable} overview summarizes the main topics and
organizational structure covered in this review.

\begin{center}
\begin{tcolorbox}[
  colback=blue!3,
  colframe=RoyalBlue,
  arc=3mm,
  boxrule=1pt,
  width=0.99\linewidth
]

\small
\begin{multicols}{2}

\RMSection{\textcolor{black}{1 Introduction}}{Introduction}

\RMSection{\textcolor{red}{2 Bridging KANs and KST}}{KAT}
\RMSection{\textcolor{red}{3 Bridging KANs and Kernel Methods}}{KANtoKernel}
\RMSection{\textcolor{red}{4 Bridging KANs and MLPs}}{KANtoMLP}
\RMSub{\textcolor{red}{4.1 How KANs Extend the MLPs}}{PINNtoPIKAN}

\RMSection{\textcolor{blue}{5 Basis Functions}}{Basis}
\RMSub{\textcolor{blue}{5.1 B-spline}}{spline}
\RMSub{\textcolor{blue}{5.2 Chebyshev Polynomial}}{Cheby}
\RMSub{\textcolor{blue}{5.3 ReLU}}{Relu}
\RMSub{\textcolor{blue}{5.4 Jacobi and General Polynomials}}{Jacobi}
\RMSub{\textcolor{blue}{5.5 Gaussian RBF}}{gaussiankan}
\RMSub{\textcolor{blue}{5.6 Fourier}}{fourierkan}
\RMSub{\textcolor{blue}{5.7 Wavelet}}{Wavelet}
\RMSub{\textcolor{blue}{5.8 Finite-Basis}}{Finite}
\RMSub{\textcolor{blue}{5.9 SincKAN}}{sinckan}

\RMSection{\textcolor{green!60!black}{6 Accuracy Improvement}}{sec:accuracy-improvement}
\RMSub{\textcolor{green!60!black}{6.1 Physics Constraints \& Loss Design}}{physics}
\RMSub{\textcolor{green!60!black}{6.2 Adaptive Sampling \& Grids}}{sample}
\RMSub{\textcolor{green!60!black}{6.3 Domain Decomposition}}{decomposition}
\RMSub{\textcolor{green!60!black}{6.4 Function Decomposition}}{function_decompose}
\RMSub{\textcolor{green!60!black}{6.5 Hybrid, Ensemble \& Integrated Models}}{hybrid}
\RMSub{\textcolor{green!60!black}{6.6 Sequence and Attention Hybrids}}{seq-hybrids}
\RMSub{\textcolor{green!60!black}{6.7 Discontinuities \& Sharp Gradients}}{subsec:discsharp}
\RMSub{\textcolor{green!60!black}{6.8 Optimization \& Adaptive Training}}{optmization}

\RMSection{\textcolor{green!60!black}{7 Efficiency Improvement}}{sec:efficiency-improvement}
\RMSub{\textcolor{green!60!black}{7.1 Parallelism \& GPU Engineering}}{parallel}
\RMSub{\textcolor{green!60!black}{7.2 Matrix \& Bases Optimization}}{matrix}

\RMSection{\textcolor{violet}{8 Sparsity \& Regularization}}{Regularization}
\RMSection{\textcolor{violet}{9 Scaling Laws \& Convergence}}{convergence}
\RMSub{\textcolor{violet}{9.1 Theoretical Approximation Rates}}{theory}
\RMSub{\textcolor{violet}{9.2 Spectral Bias}}{bias}
\RMSub{\textcolor{violet}{9.3 NTK-Based Convergence}}{ntk}
\RMSub{\textcolor{violet}{9.4 Practical Trade-offs}}{practical}

\RMSection{\textcolor{orange}{10 Practical ``Choose--Your--KAN''}}{sec:choose-kan}
\RMSection{\textcolor{orange}{11 Current Gaps and Path Forward}}{future}
\RMSection{\textcolor{black}{12 Conclusion}}{Conclude}

\end{multicols}
\end{tcolorbox}
\end{center}

\end{abstract}


\section{Introduction}\label{Introduction}

\noindent
Within the modern machine learning landscape, multilayer perceptrons (MLPs) remain a foundational architecture for function approximation and data-driven modeling. Recently, Kolmogorov--Arnold Networks (KANs), introduced by Liu et al.~\cite{Liu24,Liu24b}, have emerged as a promising alternative that rethinks the role of nonlinearities in neural networks. While MLPs rely on fixed activation functions applied at nodes, KANs instead place learnable univariate functions along network edges, a design loosely inspired by the Kolmogorov--Arnold representation theorem. This structural shift has been argued to improve interpretability and adaptability, although empirical studies suggest that its advantages over well-designed MLPs depend strongly on the problem setting and architectural choices. This emphasis on interpretability has also extended to symbolic regression and equation discovery, where KAN-based formulations aim to produce sparser, more structured, and more human-readable representations~\cite{Howard26,Faroughi26_}. To place these developments in context, we first review the role of MLPs and their extensions in scientific machine learning before introducing the KAN framework, whose conceptual origins arise from a different theoretical tradition.

\noindent
\textbf{MLPs.}
Multilayer perceptrons are universal approximators and remain a standard building block for regression, function approximation, and pattern recognition across scientific and engineering applications. A major extension of this paradigm is the physics-informed neural network (PINN) framework of Raissi et al.~\cite{Raissi19}, which has rapidly grown and enabled large-scale studies. 

\noindent
Since then, numerous advances have broadened the scope of MLP-based PINNs. Fractional operators were addressed by fPINNs~\cite{pang2019fpinns}, while uncertainty quantification for forward and inverse problems was introduced in~\cite{zhang2019quantifying}. Extensions such as VPINNs~\cite{kharazmi2019}, XPINNs for domain decomposition~\cite{jagtap2020extended}, and multi-fidelity PINNs~\cite{Meng20} improved accuracy and scalability. The framework also inspired convergence analyses~\cite{shin2020convergence}, NTK-based diagnostics~\cite{Wang22}, and practical toolkits like DeepXDE~\cite{lu2021deepxde}. More recent contributions include multi-stage training with near machine-precision accuracy~\cite{wang2023multi}, adaptive residual weighting~\cite{mcclenny2020self}, and distributed solvers for extreme-scale PDEs~\cite{Shukla21,hu2024tackling,Sifan25,Sifan24Pirate}. Collectively, these developments establish MLP–PINNs as a mature reference standard in scientific machine learning. At the same time, KANs arise from a distinct line of thought rooted in functional representations \textbf{inspired} by the Kolmogorov superposition theorem, rather than as a direct modification of the MLP paradigm.

\noindent
Despite their flexibility, MLPs face well-documented 
limitations. Fixed activation functions restrict adaptability~\cite{Rossi05}. Network behavior can be 
difficult to interpret~\cite{Cranmer23} and to attribute causally~\cite{Cunningham23}. Achieving high accuracy 
often entails large parameter counts, which can hinder 
efficient updates~\cite{Amir24,Amir24a,Amir26_reg}. Generalization 
robustness can degrade in challenging regimes~\cite{Guilhoto24,Sifan25a}. Optimization itself 
can be fragile or stiff, depending on the task and scaling~\cite{Sifan21}.
MLPs also exhibit spectral bias: a tendency to learn low frequencies faster than high frequencies~\cite{Rahaman19, Xu20}. The effect slows convergence and reduces accuracy for oscillatory or sharp-gradient solutions \cite{Cai19}. This motivates investigations into alternatives with explicit basis control.

\noindent
\textbf{KANs.}
Kolmogorov--Arnold Networks  in their modern \emph{multilayer} form were introduced by Liu et al.~\cite{Liu24} as a structured alternative to standard MLPs, in which fixed pointwise activations are replaced by \emph{learnable univariate functions} defined on network edges and composed across layers.  
This design is \emph{motivated} by the Kolmogorov representation theorem, but departs from its classical construction by learning the inner functions directly from data.

\noindent
The original KAN architecture employs B-spline bases; however, the framework is inherently modular, allowing alternative basis families—such as Chebyshev polynomials, Gaussian kernels, Fourier features, and others—to be substituted without changing the overall architecture.  
This modularity is fundamental, as the choice of basis directly governs smoothness, locality, spectral content, and numerical stability, thereby influencing expressivity, interpretability, and optimization behavior.  
As a result, adaptive basis learning combined with interchangeable basis families constitutes a defining feature of KANs, alongside a growing collection of architectural variants, optimization strategies, and theoretical analyses.

\noindent
In parallel with these methodological developments, KAN implementations have rapidly expanded across open-access platforms, with diverse variants shared by the research community.  
Curated resources such as \cite{awesomeKAN}\footnote{\url{https://github.com/mintisan/awesome-kan}} provide a centralized overview of recent developments across regression, function approximation, and PDE-solving tasks.  
Table~\ref{githubs} summarizes representative GitHub repositories illustrating this fast-growing ecosystem, which are also referenced throughout this review when discussing specific methods and design choices.

\noindent
\textbf{MLPs vs.\ KANs.}
When it comes to deciding whether MLP\hbox{-}PINNs or KAN\hbox{-}PIKANs are the better choice, the comparison is far from straightforward, with studies that emphasize fairness often reaching divergent or contradictory conclusions~\cite{Yu24_fairer}\footnote{\url{https://github.com/yu-rp/KANbeFair}},
 \cite{KAN_pde_Shukla24, Yang24_Comp}. A key reason is that KANs are not a monolithic architecture; their performance is critically influenced by the choice of basis function. For example, a comprehensive comparison by Farea and Celebi~\cite{Farea25_BasisComp}\footnote{\url{https://github.com/afrah/pinn_learnable_activation}} found that the optimal basis (e.g., B\hbox{-}spline, Fourier, Gaussian) varies significantly with the PDE being solved, making a simple ``KAN vs.\ MLP'' verdict insufficient.

\noindent
Despite these complexities, Table~\ref{KAN_vs_MLP} summarizes reported outcomes across regression, function approximation, and PDE-solving tasks. The entries reflect prevailing tendencies in the literature rather than definitive results for every case. A consistent pattern is that KANs—particularly when equipped with specialized basis functions—tend to match or outperform vanilla MLPs and PINNs in terms of accuracy and convergence speed. 

\medskip
\noindent
\textbf{Related Surveys.}
A number of recent surveys have appeared on KANs, each highlighting different
facets of this emerging architecture.
Notable contributions include those by Andrade et al.~\cite{Andrade25},
Basina et al.~\cite{Basina24_interp_review}, Beatrize et al.~\cite{Beatrize25},
Faroughi et al.~\cite{Faroughi25_review}, Kilani et al.~\cite{Kilani25},
Hou et al.~\cite{Ji24}, Dutta et al.~\cite{Dutta25_review}, Essahraui et al.~\cite{Essahraui25},
and Somvanshi et al.~\cite{Somvanshi25}, which together provide valuable entry
points into the literature.
Our review builds on these efforts by seeking to offer a more systematic and
comprehensive perspective: rather than cataloguing studies in isolation, we
integrate theoretical, architectural, optimization, and application viewpoints
into a structured roadmap.
By combining comparative analysis, methodological insights, and practical
guidance, our goal is to provide a resource that complements existing surveys and
serves readers aiming to both understand and apply KANs.

\medskip
\noindent
\textbf{Contributions.}
This review makes the following contributions:
\begin{itemize}
  \item it systematically connects KAN architectures with the Kolmogorov superposition theorem;
  \item it comprehensively bridges KANs with classical kernel methods;
  \item it compiles and compares reported accuracy, convergence behavior, and computational cost between KANs and MLPs across tasks;
  \item it provides a comprehensive and unified overview of basis functions used in KANs;
  \item it collects and organizes a wide range of open-source KAN implementations;
  \item it offers a structured and integrative perspective on KANs by unifying theoretical foundations, architectural designs, optimization strategies, and applications into a coherent roadmap for both understanding and practice.
\end{itemize}

\noindent
\noindent
\textbf{Organization and Key Insights.}
This review is structured to build a comprehensive understanding of Kolmogorov-Arnold Networks, moving from foundational theory to practical application.
Section \ref{KAT} establishes that KANs are not exact realizations of the Kolmogorov Superposition Theorem, but rather data-driven approximations inspired by its compositional structure, using learned smooth functions and modern deep architectures. Section \ref{KANtoKernel} demystifies their behavior by showing that shallow, one-dimensional KANs are mathematically equivalent to classical kernel methods, differing only in how their coefficients are optimized. Section \ref{KANtoMLP} demonstrates the formal equivalence between KANs and MLPs, highlighting that KANs achieve superior parameter efficiency and different optimization dynamics by relocating nonlinearities from nodes to edges.

\noindent
Building on this, Section \ref{Basis} presents the choice of basis function as a central design axis, providing a guide for selecting bases (e.g., splines, polynomials, Fourier) to match the properties of the target function (smooth, periodic, or discontinuous). Sections \ref{sec:accuracy-improvement} and \ref{sec:efficiency-improvement} then catalog practical techniques—such as physics-informed constraints, adaptive grids, decomposition strategies, and parallel implementations—that systematically enhance KAN accuracy and efficiency. Section \ref{Regularization} covers methods like sparsity, pruning, and Bayesian formulations that improve interpretability and generalization.

\noindent
Finally, Section \ref{convergence} explains the theoretical advantages of KANs, showing they achieve faster convergence rates, suffer less from spectral bias, and benefit from better-conditioned NTK dynamics. These insights are distilled into a practical "Choose-Your-KAN" guide in Section \ref{sec:choose-kan}. We conclude by identifying open research challenges in Section \ref{future} and summarizing our findings in Section \ref{Conclude}.

\newpage
\begin{longtable}{|l|l|l|l|}
\caption{Representative performance trends comparing KAN-based models with MLPs.
Results are aggregated from the literature and indicate qualitative trends.
Reported accuracy and convergence metrics differ across studies (e.g., relative $L_2$, RMSE).
“Slower training’’ refers to higher per-iteration cost; total time may vary with epochs.
All rows compare KANs with plain MLP/PINN baselines.}
\label{KAN_vs_MLP} \\

\hline
\textbf{Ref.} & \textbf{Accuracy} & \textbf{Convergence / Time (per iter.)} & \textbf{Basis Functions} \\
\hline
\multicolumn{4}{|c|}{\textbf{Regression \& Symbolic Representation}} \\
\hline
\cite{Ta24} & BSRBF-KAN $\approx$ MLP & Faster convergence; slower training & B-spline + Gaussian \\
\cite{He24} & MLP-KAN $>$ MLP  & -- & B-Spline  \\
\cite{Pal25} & KAN $<$ MLP  & Slower training; less generalizable & B-Spline   \\
\cite{Li25_DEKAN} & DE-KAN $>$ MLP & Faster convergence & B-spline   \\
\cite{pde_Koeing24} & KAN-ODE $>$ MLP-ODEs & Faster convergence; slower training & Gaussian   \\
\cite{Mallick25_battery} & KAN-Therm $>$ MLP & Faster convergence  & B-spline  \\
\hline
\multicolumn{4}{|c|}{\textbf{Function Approximation}} \\
\hline
\cite{Liu24} & KAN $\geq$ MLP  & Faster convergence; slower training & B-spline    \\
\cite{Yu24_fairer} & KAN $>$ MLP (symbolic) & Slower training & B-spline  \\
\cite{Actor25} & KAN $\approx$ MLP  & Faster training & Free-knot B-spline  \\
\cite{KAN_pde_Zeng24} 
& KAN $\ge$ MLP (noisy) & Faster convergence; slower training & B-spline\\
\cite{KAN_pde_Zeng24}  & KAN $\le$ MLP (non-smooth) & Faster convergence; slower training
& B-spline \\
\cite{Qiu25} & PowerMLP $>$ MLP  & Slower training & Power-ReLU  \\
\cite{Yu24} & SincKAN $>$ MLP & Faster convergence; slower training & Sinc   \\
\cite{Mahmoud25}& ChebyKAN $>$ MLP & Faster convergence; slower training & Shifted Chebyshev   \\
\cite{Jiang25_quantum} & QKAN $>$ MLP & Faster convergence & Quantum variational \\

\hline
\multicolumn{4}{|c|}{\textbf{PDE Solving}} \\
\hline
\cite{Toscano24_kkan} & KKAN $>$ PINN & Faster convergence & Various Basis \\
\cite{Wang25} & PIKAN $\gg$ PINN  & Faster convergence; slower training & B-spline  \\
\cite{Liu24} & PIKAN $\gg$ PINN  & slower training (10$\times$) & B-spline   \\
\cite{Abueidda25} & DeepOKAN $>$ DeepONet & Faster convergence; slower training & Gaussian    \\
\cite{Zhang25_comp} & PIKAN $>$ PINN  & Faster convergence & B-Spline     \\
\cite{Yang25} & KAN-MHA $>$ PINN  & Faster convergence; comparable time & B-spline + Attention   \\
\cite{Guo25} & Res-KAN $\gg$ PINN  & Faster convergence; better generalization & B-Spline + Residual \\
\cite{Xu25} & HPKM-PINN $>$ PINN  & Faster convergence; slower training & B-spline \\
\cite{Khedr25} & PI-KAN $\gg$ PINN  & Faster convergence & Spline  \\
\cite{Lei25} & DPINN $\gg$ PINN & Faster convergence; slower training &  B-spline + Fourier \\
\cite{Kalesh25} & PIKAN $\approx$ PINN & Slower training & B-spline    \\
\cite{KAN_pde_Shukla24} & PIKAN $\approx$ PINN  & slower training & Various Basis  \\
\cite{Mostajeran24} & EPi-cKAN $\gg$ PINN & Slower training; better generalization & Chebyshev \\
\cite{Mostajeran25} & Scaled-cPIKAN $\gg$ PINN & Faster convergence & Chebyshev  \\
\cite{Faroughi25} & PIKAN $\gg$ PINN  & Faster convergence & Chebyshev \\
\cite{Daryakenari25} & tanh-PIKAN $>$ PINN & -- & Chebyshev    \\
\cite{pde_Zhang24} & AL-PKAN $>$ PINN  & Faster convergence & B-spline decoder \\
\cite{Aghaei24_kantorol} & KANtrol $>$ PINN & Slower training & B-spline  \\
\cite{pde_shuai24} & PIKAN $>$ PINN  & Slower training & B-spline \\
\cite{pde_Wang24} & KINN $>$ PINNs  & Slower training & B-spline \\
\cite{Kashefi25} & PIKAN $\approx$ PINNs  & Slower training & Chebyshev \\
\cite{Zhang25} & PIKAN $\approx$ PINNs  & Slower training &  Fourier-based\\
\cite{Yang25_multiScale} &  MR-PIKAN $\gg$ PINN  & Slower training & Chebyshev \\ 
\cite{Farea25_BasisComp} &  PIKAN $\gtrless$ PINN (Prob-dep) & Faster convergence; slower training & Various Bases \\
\cite{Xiong25} & J-PIKAN $\gg$ PINN & Faster convergence; slower training & Jacobi (orthogonal) \\
\cite{Zhang26_legendre} & Legend-KINN $>$ MLP, KAN & Faster convergence; slower training & Legendre  \\
\hline
\end{longtable}

\begin{longtable}{|l|l|l|}
\caption{GitHub repositories related to KANs for regression, function approximation, or PDEs.}\label{githubs} \\
\hline
\textbf{Repository} & \textbf{Description} & \textbf{Ref.} \\
\hline
\href{https://github.com/KindXiaoming/pykan}{.../pykan} & Official PyKAN for ``KAN'' and ``KAN 2.0''. & \cite{Liu24b} \\
\hline
\href{https://github.com/AmirNoori68/PU-GKAN}{\ldots/\allowbreak{}PU-\allowbreak{}GKAN} 
& Partition-of-Unity Gaussian KAN. 
& \cite{Amir_PUGKAN}\\\hline
\href{https://github.com/AmirNoori68/Gaussian-KAN}{\ldots/\allowbreak{}Gaussian-\allowbreak{}KAN} 
& Scaling of Gaussian KANs. 
& \cite{Amir_GKAN}\\
\hline
\href{https://github.com/ParamIntelligence/Anant-Net}{.../Anant-Net} &
High-dimensional PDE solver with tensor sweeps. &
\cite{Menon25} \\
\hline
\href{https://github.com/srigas/RGA-KANs}{.../RGA-KANs} &
Deep cPIKANs with variance-preserving initialization. &
\cite{Rigas25_deep} \\
\hline
\href{https://github.com/afrah/pinn_learnable_activation}{.../pinn\_learnable\_activation} & Compares various KAN bases vs.\ MLP on PDEs & \cite{Farea25_BasisComp} \\
\hline
\href{https://github.com/1ssb/torchkan}{.../torchkan} & Simplified PyTorch KAN with variants & \cite{TorchKAN} \\
\hline
\href{https://github.com/mintisan/awesome-kan}{.../awesome-kan} & Curated list of KAN resources, projects, and papers. & \cite{awesomeKAN} \\
\hline
\href{https://github.com/sidhu2690/Deep-KAN}{.../Deep-KAN} &  Spline-KAN examples and a PyPI package. & \cite{DeepKAN} \\
\hline
\href{https://github.com/sidhu2690/RBF-KAN}{.../RBF-KAN} & RBF-KAN examples  & \cite{RBFKAN} \\
\hline
\href{https://github.com/yu-rp/KANbeFair}{.../KANbeFair} & Fair benchmarking of KANs vs MLPs. & \cite{Yu24_fairer} \\
\hline
\href{https://github.com/Blealtan/efficient-kan}{.../efficient-kan} & Efficient PyTorch implementation of KAN. & \cite{EfficientKAN} \\
\hline
\href{https://github.com/pnnl/spikans}{.../spikans} &
Separable PIKAN (SPIKAN). &
\cite{pde_jacob24} \\
\hline
\href{https://github.com/srigas/jaxKAN}{.../jaxKAN} & JAX-based KAN package with grid extension support. & \cite{pde_Rigas24f} \\
\hline
\href{https://github.com/ZiyaoLi/fast-kan}{.../fast-kan} & FastKAN using RBFs. & \cite{Li24} \\
\hline
\href{https://github.com/AthanasiosDelis/faster-kan}{.../faster-kan} &  Using SWitch Activation Function & \cite{Athanasios2024} \\
\hline
\href{https://github.com/Indoxer/LKAN}{.../LKAN} & Implementations of KAN variations. & \cite{liu2024kan} \\
\hline
\href{https://github.com/pnnl/neuromancer/tree/feature/fbkans}{.../neuromancer$->$fbkans} & Parametric constrained optimization. & \cite{pde_fbkan_Howard24} \\
\hline
\href{https://github.com/quiqi/relu_kan}{.../relu\_kan} & Minimal ReLU-KAN. & \cite{Qiu24} \\
\hline
\href{https://github.com/OSU-STARLAB/MatrixKAN}{.../MatrixKAN} & Matrix-parallelized KAN. & \cite{Coffman25} \\
\hline
\href{https://github.com/Iri-sated/PowerMLP}{.../PowerMLP} & MLP-type network with KAN-level expressiveness. & \cite{Qiu25} \\
\hline
\href{https://github.com/GistNoesis/FourierKAN}{.../FourierKAN} & Fourier-based KAN layer. & \cite{FourierKAN} \\
\hline
\href{https://github.com/GistNoesis/FusedFourierKAN}{.../FusedFourierKAN} &  Optimized FourierKAN with fused GPU kernels & \cite{FusedFourierKAN} \\
\hline
\href{https://github.com/alirezaafzalaghaei/fKAN}{.../fKAN} & Fractional KAN using Jacobi functions. & \cite{Aghaei24_fkan} \\
\hline
\href{https://github.com/alirezaafzalaghaei/rKAN}{.../rKAN} & Rational KAN (Padé/Jacobi rational designs). & \cite{Aghaei24_rkan} \\
\hline
\href{https://github.com/M-Wolff/CVKAN}{.../CVKAN} & Complex-valued KANs. & \cite{Wolff25} \\
\hline
\href{https://github.com/DUCH714/SincKAN}{.../SincKAN} & Sinc-based KAN with PINN applications. & \cite{Yu24} \\
\hline
\href{https://github.com/SynodicMonth/ChebyKAN}{.../ChebyKAN} & Chebyshev polynomial-based KAN variant. & \cite{ChebyKAN} \\
\hline
\href{https://github.com/Boris-73-TA/OrthogPolyKANs}{.../OrthogPolyKANs} & Orthogonal polynomial-based KAN implementations. & \cite{OrthogPolyKANs} \\
\hline
\href{https://github.com/kolmogorovArnoldFourierNetwork/kaf_act}{.../kaf\_act} & PyTorch activation combining with RFF. & \cite{kaf_act} \\
\hline
\href{https://github.com/kolmogorovArnoldFourierNetwork/KAF}{.../KAF} & Kolmogorov-Arnold Fourier Networks. & \cite{Zhang25} \\
\hline
\href{https://github.com/kelvinhkcs/HRKAN}{.../HRKAN} & Higher-order ReLU-KANs. & \cite{KAN_pde_So24} \\
\hline
\href{https://github.com/yizheng-wang/Research-on-Solving-Partial-Differential-Equations-of-Solid-Mechanics-Based-on-PINN}{.../KINN} & PIKAN for solid mechanics PDEs. & \cite{pde_Wang24} \\
\hline
\href{https://github.com/Ali-Stanford/Physics_Informed_KAN_PointNet}{.../PIKAN\_PointNet} & Chebushev for solving time-independent inverse problems. & \cite{Kashefi25} \\
\hline
\href{https://github.com/Jinfeng-Xu/FKAN-GCF}{.../FKAN-GCF} & FourierKAN-GCF for graph filtering. & \cite{Xu25_fourier} \\
\hline
\href{https://github.com/jdtoscano94/KKANs_PIML}{.../KKANs\_PIML} & Kurkova-KANs combining MLP with basis functions. & \cite{Toscano24_kkan} \\
\hline
\href{https://github.com/Zhangyanbo/MLP-KAN}{.../MLP-KAN} & MLP-augmented KAN activations. & \cite{MLP-KAN} \\
\hline
\href{https://github.com/Adamdad/kat}{.../kat} & Kolmogorov-Arnold Transformer. & \cite{Yang25_transformer} \\
\hline
\href{https://github.com/YihongDong/FAN}{.../FAN} & Fourier Analysis Network (FAN). & \cite{Dong25_FAN} \\
\hline
\href{https://github.com/seydi1370/Basis_Functions}{.../Basis\_Functions} & Polynomial bases for KANs (comparative study). & \cite{Seydi24a} \\
\hline
\href{https://github.com/zavareh1/Wav-KAN}{.../Wav-KAN} & Wav-KAN: wavelet-based KANs. & \cite{Bozorgasl24} \\
\hline
\href{https://github.com/Jim137/qkan}{.../qkan} & Quantum variational activations and pruning tools. & \cite{Jiang25_quantum} \\
\hline
\href{https://github.com/liouvill/KAN-Converge}{.../KAN-Converge} & 
Additive \& hybrid KANs for convergence-rate experiments  & \cite{Liu25_convergence} \\
\hline
\href{https://github.com/hoangthangta/BSRBF_KAN}{.../BSRBF\_KAN} & Combines B-splines and radial basis functions. & \cite{Ta24} \\
\hline
\href{https://github.com/wmdataphys/Bayesian-HR-KAN}{.../Bayesian-HR-KAN} & Introduces Bayesian higher-order ReLU-KANs. & \cite{pde_bayesian_Giroux24} \\
\hline
\href{https://github.com/zhang-zhuo001/Legend-KINN}{.../Legend-KINN} & Legendre polynomial--based KAN. & \cite{Zhang26_legendre} \\
\hline
\href{https://github.com/DiabAbu/DeepOKAN}{.../DeepOKAN} & Deep Operator Network based on KAN. & \cite{Abueidda25} \\
\hline
\href{https://github.com/DENG-MIT/LeanKAN}{.../LeanKAN} & A memory-efficient Kolmogorov--Arnold Network. & \cite{Koenig25} \\
\hline
\href{https://github.com/Aqasch/KANQAS_code}{.../KANQAS\_code} & KANQAS: KAN for Quantum Architecture Search. & \cite{Kundu24_quantom} \\
\hline
\href{https://github.com/andrewpolar/pkan}{.../pkan} & Probabilistic KAN via divisive data re-sorting. & \cite{Polar25} \\
\hline
\href{https://github.com/teocala/pihnn}{.../pihnn} & Holomorphic KAN/PIHNN framework for solving PDEs. & \cite{Clafa25} \\
\hline
\href{https://github.com/schwallergroup/lmkan}{.../lmkan} &
Lookup-based KAN for fast high-dim mappings. &
\cite{Pozdnyakov25} \\
\hline
\href{https://github.com/srigas/KAN_Initialization_Schemes}{.../KAN\_Initialization\_Schemes} &
Initialization schemes for spline-based KANs. &
\cite{Rigas25_init} \\
\hline
\href{https://github.com/geoelements-dev/mlp-kan}{.../mlp-kan} &
KAN vs.\ MLP for PDEs in DeepONet/GNS frameworks. &
\cite{Pant25} \\
\hline
\hline

\end{longtable}

\section{Bridging KANs and KST}\label{KAT}

\noindent
\textbf{Hilbert’s Question and Superpositions.}
At the 1900 International Congress of Mathematicians,
Hilbert posed what later became known as \emph{Hilbert’s 13th Problem}
\cite{Hilbert02}.
He asked whether every continuous function of three variables,
\[
f : [0,1]^3 \to \mathbb{R},
\]
can be represented as a finite superposition of continuous functions of at most
two variables.
Hilbert conjectured that such a representation was impossible, suggesting that
genuinely multivariate functions could not be decomposed into simpler lower-arity
components.

\medskip
\noindent
To clarify what is meant by a \emph{superposition} in this question—without
addressing the existence of such representations—consider a formal expression of
the form
\[
F(x,y,z)
=
f\!\bigl(g(x,y),\,h(x,\,k(y,z))\bigr),
\]
which illustrates how a function of three variables may be written as nested
compositions of bivariate and univariate functions.
Even the elementary identity
\[
x_1+\cdots+x_n
=
x_1+\bigl(x_2+\cdots+(x_{n-1}+x_n)\cdots\bigr)
\]
is a simple instance of a superposition, obtained by repeated binary operations.

\medskip
\noindent
Contrary to Hilbert’s conjecture, the seminal results of Kolmogorov and Arnol'd
showed that every continuous function of \(s\) variables can indeed be represented
using only sums and superpositions of univariate continuous functions.
This discovery, now known as the Kolmogorov–Arnold superposition theorem, radically
changed the understanding of functional representation and forms the theoretical
foundation underlying modern Kolmogorov–Arnold Networks.

\medskip
\noindent
\textbf{Kolmogorov and Arnol'd: resolving Hilbert’s 13th Problem.}
In a short 1956 note~\cite{Kolmogorov56}, \textbf{Kolmogorov} proposed the radical idea
that multivariate continuous functions might be representable through
superpositions of functions of fewer variables.
This was not yet a theorem, but a conceptual breakthrough supported by
constructive examples.
In 1957, \textbf{Arnol'd} confirmed the conjecture for the three-variable case~\cite{Arnold57},
providing an explicit representation of a general function \(f(x,y,z)\) as a
superposition of continuous bivariate functions.
Later that same year, Kolmogorov established the first complete
\emph{theorem}~\cite{Kolmogorov57}, proving that any continuous function of
\(n>2\) variables can be expressed using only univariate functions and addition.
This result definitively resolved Hilbert’s 13th Problem.

\medskip
\begin{theorem}[Kolmogorov Superposition Theorem (KST) {\cite{Kolmogorov57}}]
Let \(E_1=[0,1]\) and \(E_n=[0,1]^n\).  
For every integer \(n>2\), there exist continuous \emph{universal} inner functions
\[
\phi_{q,p}:E_1 \to \mathbb{R},
\qquad
p=1,\dots,n,\;\; q=0,\dots,2n,
\]
independent of the target function \(f\), such that every continuous function
\(f \in C(E_n)\) admits the representation
\begin{equation}
f(x_1,\ldots,x_n)
=
\sum_{q=0}^{2n}
\Phi_q\!\Bigg(
  \sum_{p=1}^{n}
  \phi_{q,p}(x_p)
\Bigg),
\qquad (x_1,\ldots,x_n)\in E_n,
\label{eq:KST_1}
\end{equation}
where the outer functions \(\Phi_q:\mathbb{R}\to\mathbb{R}\) are continuous and depend on \(f\), while the inner functions \(\phi_{q,p}\) are universal.
\end{theorem}

\noindent
The KST shows that multivariate continuity can be synthesized entirely from univariate nonlinearities and addition, a structural insight that later motivated two–layer interpretations and neural–network analogies.

\medskip
\noindent
\textbf{Proof strategies and later refinements.}
Two major approaches emerged in the decades following Kolmogorov’s work:

(i)~\emph{Baire–category proofs}:  
Kahane~\cite{Kahane75} and Hedberg~\cite{Hedberg71} showed that for generic tuples of continuous increasing functions, the superposition operator is surjective.  
Their arguments reveal that KST is not only true but \emph{abundant}: for almost all choices of inner functions, an appropriate outer function exists.

(ii)~\emph{Constructive proofs}:  
Lorentz~\cite{Lorentz62}, Sprecher~\cite{Sprecher65}, Arnol'd~\cite{Arnold58}, and others developed explicit (though highly involved) constructions.  
These proofs follow a common pattern: partition \(E_n\) into subcubes, map each subcube into a disjoint real interval via the inner functions, and build the outer functions iteratively to approximate the target \(f\).  
Kolmogorov’s original paper left several analytic details unstated; these were filled in by Arnol'd~\cite{Arnold58} and later authors, making the constructive picture complete.

\medskip
\noindent
This classical theory forms the starting point for the refinements described next: reductions in the number of unique functions, improvements in regularity, and the development of computationally usable versions that ultimately connect the superposition viewpoint to modern KAN architectures.

\medskip
\noindent
\textbf{Refinements to KST.}
Kolmogorov’s constructive formulation \eqref{eq:KST_1} established the universality of the inner functions \(\phi_{q,p}\), but it did so at considerable structural cost: the representation requires \(2n^2{+}n\) distinct inner functions and \(2n{+}1\) outer functions.  
While theoretically elegant, this level of redundancy makes the original form far too complex for computation or numerical approximation.  
The decades following Kolmogorov’s work therefore focused on two goals:  
(i) reducing the number of \emph{unique} functions needed, and  
(ii) improving their regularity.  
These efforts produced the modern, compressed understanding of KST that
underlies computational constructions and contemporary reinterpretations,
including recent analyses by Ismailov and Ismailov~\cite{Ismailov24}.

\medskip
\noindent
\textbf{(i) Reducing the number of functions.}
Sternfeld~\cite{Sternfeld79} showed that the number of summands \(2n+1\)
in the Kolmogorov superposition formula is optimal in a functional--analytic
sense and cannot be reduced within the class of continuous superpositions.
However, his functional–analytic framework shows that these \(2n{+}1\)
summands do \emph{not} require \(2n{+}1\) different inner or outer
functions: the same small collection of functions can be reused repeatedly.
Thus the number of \emph{distinct} functions can be much smaller, even
though the number of summands is fixed.

\smallskip
\noindent
\emph{Lorentz’ reduction of outer functions.}
Lorentz’s 1962 exposition~\cite{Lorentz62} clarified Kolmogorov’s argument and showed
that the \(2n{+}1\) outer functions \(\Phi_q\) can be replaced by a single continuous
function~\(\Phi\).

\begin{theorem}[Kolmogorov superposition in two variables {\cite{Lorentz62}}]
There exist continuous, monotone increasing functions
\[
\phi_{1,q},\, \phi_{2,q} : [0,1] \to [0,1],
\qquad q = 1,\ldots,5,
\]
such that every continuous function \(f : [0,1]^2 \to \mathbb{R}\) admits the
representation
\begin{equation}
f(x_1,x_2)
=
\sum_{q=1}^{5}
\Phi\!\bigl(\phi_{1,q}(x_1)+\phi_{2,q}(x_2)\bigr),
\label{eq:lorentz-2d}
\end{equation}
where the outer function \(\Phi : [0,2] \to \mathbb{R}\) is continuous and depends on
\(f\).
\end{theorem}

\noindent
In this formulation, the ten inner functions \(\phi_{1,q}\) and \(\phi_{2,q}\) arise
because the two-variable case requires five summands, each involving one univariate
function of \(x_1\) and one of \(x_2\).
These inner functions are universal and independent of \(f\), while the outer
function \(\Phi\) depends on the specific target.

\noindent
More generally, for \(s\) variables Lorentz obtained
\begin{equation}
f(x_1,\ldots,x_s)
=
\sum_{q=1}^{2s+1}
\Phi\!\left(
      \sum_{p=1}^{s} \phi_{p,q}(x_p)
  \right),
\label{eq:Lorentz_general_s}
\end{equation}
demonstrating that the dependence on \(f\) enters only through a single continuous
outer function~\(\Phi\).

\smallskip
\noindent
\emph{Sprecher’s reduction of inner functions.}
Sprecher~\cite{Sprecher65} achieved the corresponding simplification for the inner
layer. He showed that the entire family \(\{\phi_{q,p}\}\) can be generated from a
\emph{single} monotone universal function \(\psi:E_1\to E_1\), together with fixed
positive weights \(\{\lambda_p\}_{p=1}^n\) and uniform shifts
\(x\mapsto x+tq\), where \(t>0\) is an arbitrarily small rational parameter.

\begin{theorem}[Sprecher 1965 {\cite{Sprecher65}}]
For each integer \(N \ge 2\) there exists a monotone increasing function
\(\psi : E_1 \to E_1\) satisfying
\[
\psi \in \mathrm{Lip}\!\left( \frac{\ln 2}{\ln(2N-2)} \right),
\qquad
\psi(0)=0,\;\psi(1)=1,
\]
such that for every \(n \le N\), every \(f \in C(E_n)\), and every \(\delta>0\),
there exists a rational \(t\) with \(0 < t \le \delta\) for which
\begin{equation}
f(x_1,\ldots,x_n)
=
\sum_{q=1}^{2n+1}
g\!\left(
      \sum_{p=1}^{n} \lambda_p\,\psi(x_p + t q)
      + q
  \right),
\label{eq:KST_Sprecher}
\end{equation}
where \(g : \mathbb{R} \to \mathbb{R}\) is continuous, and the coefficients
\(\{\lambda_p\}_{p=1}^n\) together with the function \(\psi\) are universal and
independent of \(f\).
\end{theorem}

\noindent
Sprecher’s refinement thus collapses Kolmogorov’s family of inner functions to a
single universal function \(\psi\), and merges the family of outer functions
\(\{g_q\}\) into a single continuous \(g\).
In the representation~\eqref{eq:KST_Sprecher}, all significant structural complexity
lies in the universal components \(\psi\) and \(\{\lambda_p\}\), while the dependence
on the target function \(f\) is contained solely in the choice of \(g\).

\smallskip
\noindent
By combining Lorentz’ outer-function reduction and Sprecher’s inner-function compression, one obtains the modern compressed version of the superposition theorem:
\begin{equation}
f(x_1,\ldots,x_n)
=
\sum_{q=0}^{2n}
\Phi\!\Bigg(
  \sum_{p=1}^{n}
    \lambda_p\,\psi(x_p+\varepsilon q)
  +\delta q
\Bigg),
\label{eq:KST_modern}
\end{equation}
where \(\psi:E_1\to\mathbb{R}\) is a single universal inner function, \(\Phi:\mathbb{R}\to\mathbb{R}\) is a continuous outer function depending on \(f\), the constants \(\{\lambda_p\}\) are positive and rationally independent, \(\varepsilon>0\) is a uniform shift, and \(\delta\in\mathbb{R}\) is an offset introduced for separation of summands.  
\medskip
\noindent
\textbf{(ii) Improving regularity.}
The inner functions in Kolmogorov’s original proof are highly irregular—Hölder
continuous but nowhere differentiable.
A function \(u:E_1\to\mathbb{R}\) is Hölder continuous with exponent
\(\alpha\in(0,1]\) if
\[
|u(x)-u(y)|\le L|x-y|^\alpha,\qquad x,y\in E_1,
\]
for some \(L>0\).
The case \(\alpha=1\) corresponds to Lipschitz continuity.

\noindent
This lack of smoothness is not accidental but a fundamental requirement.
Vitushkin and Henkin
\cite{Vitushkin54_nonSmooth,Vitushkin77_nonSmooth,Henkin64_nonSmooth}
proved that no universal inner family can possess \(C^1\) (or higher)
smoothness.
Although \(C^1\) regularity implies Lipschitz continuity on compact sets,
differentiability itself destroys the Kolmogorov separation property required
for exact superposition.
As a result, smooth inner functions are incompatible with exact KST
representations.

\noindent
Within the theoretical development of KST, Fridman~\cite{Fridman67} showed that
the universal inner functions may be chosen Lipschitz continuous,
corresponding to the maximal regularity compatible with Kolmogorov-type
superpositions.
Sprecher’s compressed formulation~\cite{Sprecher65} can be realized within this
Lipschitz framework.
More recently, Actor and Knepley~\cite{Actor18,Actor19} formalized this
obstruction by proving that \(C^1\) inner maps cannot satisfy the Kolmogorov
separation property.
These results confirm that Lipschitz continuity is optimal, and that any attempt
to smooth the universal inner functions necessarily breaks exact KST.

\medskip
\noindent
\textbf{Computational KST.}
Algorithmic realizations of the Kolmogorov--Arnold superposition theorem may be
broadly categorized as follows:
\begin{itemize}
    \item \emph{Exact constructive KST}, where both the universal inner function and
    the outer functions are generated by an explicit recursive procedure;
    \item \emph{Approximate KST (outer-only)}, in which the inner function is fixed
    and only the outer functions are approximated or learned;
    \item \emph{Approximate KST (inner--and--outer)}, where both components are
    approximated within a superposition-based architecture.
\end{itemize}

\medskip
\noindent
\textbf{Exact constructions.}
Early digit-based constructions by Sprecher
\cite{Sprecher96,Sprecher97} provided one of the first fully computable realizations
of Kolmogorov’s superposition theorem.
Köppen~\cite{Koppen02} later identified flaws in the original recursion, which were
corrected in the rigorous formulation of Braun and Griebel~\cite{Braun09}.
Their construction guarantees continuity, monotonicity, and convergence, and shows
that the universal inner function \(\psi\) is not merely an abstract object, but can
be generated explicitly by a finite, algorithmic recursion.

\medskip
\noindent
From an intuitive standpoint, the inner function \(\psi\) can be viewed as a
\emph{universal nonlinear reparameterization} of the input.
In the Kolmogorov superposition theorem, the original input lies in \([0,1]^n\), so the
parameter \(n\) is the dimension of the input space.
The role of \(\psi\) is to encode information from this \(n\)-dimensional input into
a single scalar variable in a universal way.

\medskip
\noindent
The construction of \(\psi\) involves three parameters with clearly distinct roles.
The dimension parameter \(n\) determines how many input variables are being folded
into one and therefore directly controls how complex the encoding must be.
The base parameter \(\gamma>1\) denotes the scaling factor used in the digit-based
representation that separates successive refinement levels.
The refinement level \(k \in \mathbb{N}\) indicates the number of recursive
construction steps used to generate the finite approximation \(\psi_k\) of the
inner function.
Increasing \(k\) refines the same basic geometric pattern, while changing \(n\)
modifies the structure of that pattern itself.

\begin{figure}[t]
  \centering
  \includegraphics[width=\textwidth]{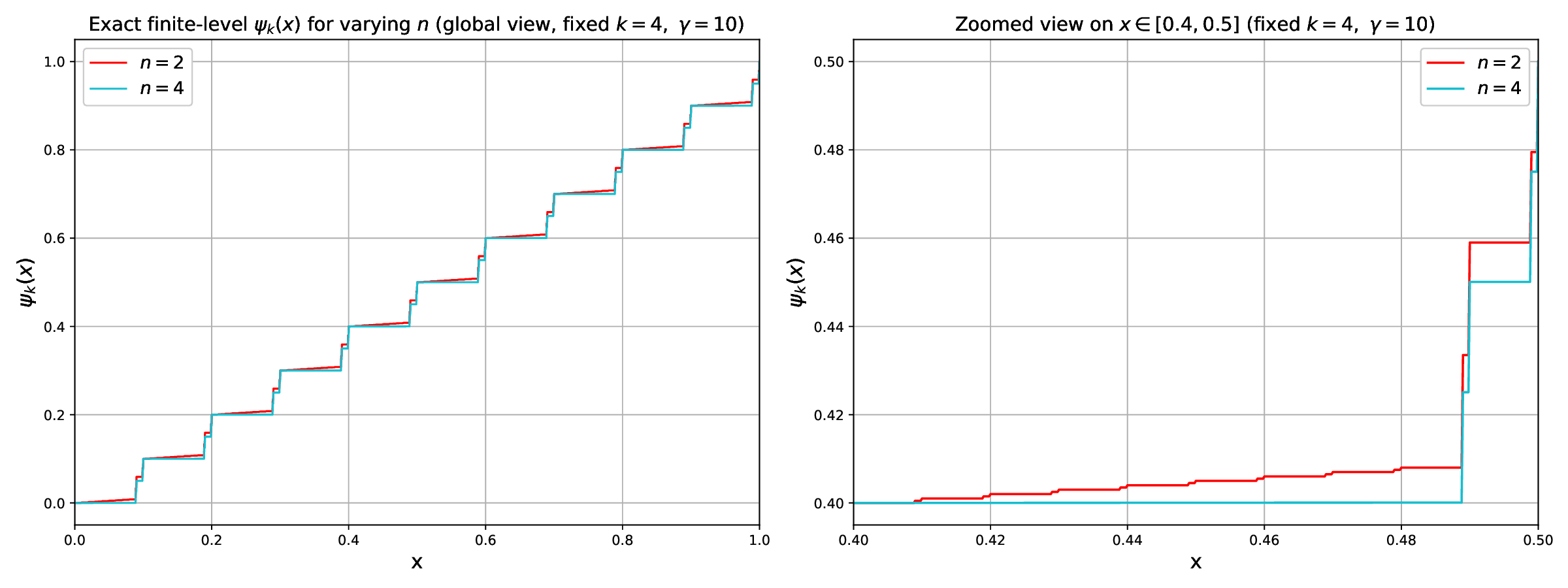}
\caption{Exact finite-level inner functions \(\psi_k(x)\) generated by the
constructive Kolmogorov--Köppen inner-function scheme for fixed refinement level
\(k\) and base parameter \(\gamma=10\), comparing the cases \(n=2\) and \(n=4\).
The case \(n=2\) exhibits many small visible steps, leading to a more oscillatory
appearance, whereas \(n=4\) shows flatter regions interspersed with fewer but larger
transitions due to stronger suppression of fine-scale increments.
Left: global 2D view on \([0,1]\).
Right: zoomed 2D view on \(x \in [0.4, 0.5]\), highlighting the contrast in local
regularity.}

  \label{fig:inner_all}
\end{figure}

\medskip
\noindent
Figure~\ref{fig:inner_all} illustrates this distinction by plotting exact finite-level
inner functions \(\psi_k(x)\) for fixed \(k\) and \(\gamma\), comparing the cases
\(n=2\) and \(n=4\).
When \(n=2\), the inner function displays many small visible steps, giving a more
oscillatory appearance and stronger apparent nondifferentiability at the plotted
scale.
When \(n=4\), the function appears flatter over longer intervals but exhibits fewer
and larger transitions.
This occurs because increasing \(n\) rapidly suppresses the amplitude of fine-scale
refinement steps, causing local irregularities to concentrate into larger but rarer
features.
As a result, varying \(n\) provides clearer insight into how multivariate input
dimension influences the geometry of the universal inner function than varying the
refinement level alone.

\noindent
Related constructive and guaranteed approaches include the work of Nakamura
et al.~\cite{Nakamura93}, who developed algorithmic constructions of both the
universal inner map~\(\psi\) and the outer correction map~\(\chi\), together with
certified approximation error bounds.
Together, these contributions clarify not only the theoretical existence of
Kolmogorov-type representations, but also their explicit realizability and geometric
structure.

\medskip
\noindent
\textbf{Approximate KST (outer-only; inner fixed).}
This line of work preserves the universal inner map and reduces learning to a one-dimensional outer problem.

\begin{itemize}
    \item De~Figueiredo (1980): fixes Sprecher’s universal inner function~$\psi$ and learns the outer function~$g$ using a Chebyshev basis---the first data-driven outer-fitting scheme in a KST layout~\cite{Figueiredo80}.
    \item Frisch et al.\ (1989): retain Lorentz’s universal inner maps and iteratively refine a single outer function~$g$~\cite{Frisch89}.
    \item Braun--Griebel inner + RKHS outer (2009): uses the corrected universal inner function~$\psi$ and replaces constructive outer windows by spline/Fourier RKHS approximants~\cite{Braun09a}.
    \item Lai--Shen (2025): again fix the Lorentz--Sprecher inner maps and reconstruct the outer function using linear B--splines with \(O(n^{-2})\) accuracy~\cite{Lai25_kst,Lai25_kst_a}.
\end{itemize}

\noindent
Outer-only methods preserve inner universality and thus avoid the curse of dimensionality: all problem dependence flows through a univariate outer approximation.

\medskip
\noindent
\textbf{Approximate KST (inner--and--outer).}
These approaches approximate or replace the universal inner functions of the
Kolmogorov superposition theorem, thereby departing from the exact Kolmogorov
formulation while enabling practical approximation and learning.
\begin{itemize}
    \item \textbf{Kurková (1991--1992):}
    initiated the approximation-theoretic study of the Kolmogorov superposition
    theorem by interpreting Kolmogorov’s structure in the sense of
    \emph{approximation} rather than exact representation.
    In the Kolmogorov--Sprecher form, the universal inner functions need only satisfy
    prescribed values on finitely many intervals and can therefore be approximated
    by staircase-like combinations of a sigmoidal activation.
    This yields a fully sigmoidal analogue of the Kolmogorov scheme and resolves
    earlier concerns regarding non-smooth inner maps and non-parametric outer
    functions~\cite{Kurkova91,Kurkova92}.

    \begin{theorem}[Kurková 1992~\cite{Kurkova92}]
    Let \(n \ge 2\), let \(\sigma : \mathbb{R}\to I\) be sigmoidal, and let
    \(f \in C(I^n)\).
    For every \(\varepsilon>0\) there exist \(k\in\mathbb{N}\) and functions
    \(\varphi_i, \psi_{p,i} \in S(\sigma)\) such that
    \[
    \Bigl|
     f(x_1,\ldots,x_n)
     -
     \sum_{i=1}^{k}
     \varphi_i\!\Bigl(\sum_{p=1}^{n}\psi_{p,i}(x_p)\Bigr)
    \Bigr|
    < \varepsilon,
    \qquad
    \forall (x_1,\ldots,x_n) \in I^n.
    \]
    \end{theorem}

    Thus, the Kolmogorov superposition \(\sum_{p}\phi_{p,q}(x_p)\) can be replaced by
    smooth sigmoidal compositions, yielding learnable inner and outer functions
    within a two-hidden-layer architecture.
    Kurková’s results provided the first rigorous bridge between Kolmogorov-type
    superpositions and standard neural network approximation, demonstrating both
    theoretical universality and practical realizability.

    \item \textbf{Nees (1994, 1996):}
    constructed explicit recursive piecewise-linear approximations of the inner
    maps \(\phi_q^{k}\) together with data-driven outer corrections \(g_r\), achieving
    geometric error decay \cite{Nees94,Nees96}.

    \item \textbf{Spline-based architectures:}
    The KSN model of Igelnik and Parikh (2003)~\cite{Igelnik03} uses cubic splines for
    \emph{both} inner and outer functions, while Coppejans (2005)~\cite{Coppejans05}
    introduces monotone inner splines for statistical identifiability.
    Related spline-based superpositional models, including ExSpliNet
    \cite{Fakhoury22,Fakhoury25}, KASAM~\cite{Deventer22}, and the Urysohn-operator
    framework of Polar and Poluektov~\cite{Polar21}, adopt additive or hierarchical
    structures reminiscent of KST but do not enforce universality of the inner layer.
\end{itemize}

\noindent
\emph{Caveat.}  
Exact KST requires a universal (target-independent) inner function; approximating or smoothing this function breaks universality and may violate separation.  
Classical results of Vitushkin and Henkin, and the modern analysis of Actor--Knepley, show that no \(C^1\) or overly smooth inner functions can satisfy the Kolmogorov separation property%
~\cite{Vitushkin54_nonSmooth,Henkin64_nonSmooth,Vitushkin77_nonSmooth,Actor19}.  
Thus inner approximations are meaningful only in the \emph{approximation-theoretic} sense, not as exact KST.

\medskip
\noindent
\textbf{Neural interpretations.}
De~Figueiredo (1980)~\cite{Figueiredo80} was the first to view the Kolmogorov–Sprecher representation as a computational architecture, expressing it in block-diagram form and proposing a data-driven procedure for learning the outer function \(g\) using Chebyshev bases.  
Later, Hecht--Nielsen (1987)~\cite{Nielsen87} interprets Sprecher’s representation as a three-stage mapping network with a fixed universal hidden layer and learned outer functions.  
Kurková’s approximation framework (1991–1992)~\cite{Kurkova91,Kurkova92} provides compatibility with standard trainable neural networks by replacing the inner staircase with smooth sigmoidal combinations while learning the outer layer.  
Spline-based models such as KSN (2003)~\cite{Igelnik03} and the shape-constrained variant of Coppejans (2005)~\cite{Coppejans05} train both inner and outer functions within spline families, yielding practical but non-KST architectures.  
See Table~\ref{history} for a chronological overview of these developments, tracing the progression from the original KST theory to its computational and neural interpretations.

\begin{table}[h!]
\centering
\scriptsize
\caption{Development of Kolmogorov’s Superposition Theorem towards Kolmogorov-inspired networks (1900–2024).}\label{history}
\begin{tabular}{|l|l|p{10.5cm}|l|}
\hline
\textbf{Year} & \textbf{Author(s)} & \textbf{Key Contribution} & \textbf{Focus on} \\
\hline

1900 & Hilbert \cite{Hilbert02} &
Poses Hilbert's 13th problem &
Theory \\

1956 & Kolmogorov \cite{Kolmogorov56} &
Preliminary idea of superpositions; first hint toward the theorem. &
Theory \\

1957 & Arnol'd \cite{Arnold57} &
First explicit 3-variable construction with 9 terms; provides concrete counterexample to Hilbert 13. &
Theory \\

1957 & Kolmogorov \cite{Kolmogorov57} &
Full Kolmogorov Superposition Theorem; first general $n$-D proof, &
Theory \\

1958 & Arnol'd \cite{Arnold58} &
Supplies missing lemmas; first to complete Kolmogorov’s proof. &
Theory \\

1962 & Lorentz \cite{Lorentz62} &
Gives simplified canonical form with \emph{one} outer function. &
Theory \\

1965 & Sprecher \cite{Sprecher65} &
First single universal inner function; reduces many $\phi_{p,q}$ to one $\psi$ with shifts; major structural simplification over Kolmogorov. &
Theory \\

1967 & Fridman \cite{Fridman67} &
Shows universal inner functions may be taken $\mathrm{Lip}(1)$; first optimal smoothness result; strengthens Sprecher–Kolmogorov scheme. &
Theory \\

1980 & de Figueiredo \cite{Figueiredo80} &
First network-like interpretation; introduces block diagram + data-driven outer fitting; operationalizes Sprecher’s form. &
Theory \\

1987 & Hecht--Nielsen \cite{Nielsen87} &
First explicit neural mapping theorem; views KST as two-hidden-layer network. &
Network \\

1989 & Girosi--Poggio \cite{Girosi89} &
First critical analysis of learnability; shows inner function must be non-smooth + outer non-parametric. &
Theory \\

1989 & Frisch et al.\ \cite{Frisch89} &
First computational implementation of Lorentz form; iterative learning of single outer function. &
Network \\

1991 & Kurková \cite{Kurkova91} &
First approximation-theoretic reinterpretation; ties network size to modulus of continuity. &
Network \\

1992 & Kurková \cite{Kurkova92} &
Two-hidden-layer sigmoidal approximants; universal inner parameters. &
Both \\

1993 & Sprecher \cite{Sprecher93} &
Single universal $\psi$ for \emph{all} $n$. &
Both \\

1993 & Nakamura et al.\ \cite{Nakamura93} &
First fully constructive + guaranteed-accuracy version. &
Theory \\

1994 & Nees \cite{Nees94} &
First use of spline (piecewise-linear) inner maps, introducing geometric error decay and a fully constructive, algorithmic approximation of KST with explicit error bounds; &
Both \\

1996 & Sprecher \cite{Sprecher96} &
First executable version of $\psi$ with separation guarantees. &
Both \\

1997 & Sprecher \cite{Sprecher97} &
Explicit algorithm for outer functions. &
Both \\

2002 & Köppen \cite{Koppen02} &
Corrected continuous monotone inner function of Sprecher \cite{Sprecher96}; first training-ready inner map. &
Network \\

2003 & Igelnik--Parikh \cite{Igelnik03} &
First fully trainable Kolmogorov Spline Network; replaces fractal inner functions with smooth cubic splines and provides learnable inner and outer maps. &
Network \\

2009 & Braun--Griebel \cite{Braun09} &
First correct constructive KST; fixes Sprecher’s errors \cite{Sprecher96}; modern constructive foundation. &
Theory \\

2019 & Actor--Knepley \cite{Actor19} &
Proves $C^1$ inner functions impossible; first rigorous smoothness obstruction. &
Theory \\

2024 & Liu \emph{et al.} \cite{Liu24} &
First deep network architecture inspired by the Kolmogorov–Arnold representation (KAN). &
Network \\

\hline
\end{tabular}
\end{table}

\subsubsection*{2024: Liu et al.\ --- Kolmogorov–Arnold Network}

Liu et al.~\cite{Liu24,Liu24b, Liu25_kanl} introduce the \emph{Kolmogorov–Arnold Network}, a
neural architecture explicitly \emph{motivated} by the structure of the
KST~\eqref{eq:KST_1}, though not intended as a
mathematical realization of it.  
The authors emphasize that KAN is \emph{not} an exact implementation of the Kolmogorov
Superposition Theorem, but rather a computational architecture ``inspired by''  the Kolmogorov–Arnold representation.

\medskip
\noindent
Because the Kolmogorov Superposition Theorem prescribes only a 
two-layer structure (universal inner functions \(\phi_{q,p}\) followed by
outer maps \(\Phi_q\)), Liu et al.\ generalize the idea by redefining the
basic computational block.  
A \emph{KAN layer} is introduced as a matrix of
univariate functions,
\[
\boldsymbol{\Phi}_\ell
=
\{\phi_{\ell,j,i}\},
\qquad
i=1,\ldots,n_\ell,\;\;
j=1,\ldots,n_{\ell+1},
\]
where \(n_\ell\) and \(n_{\ell+1}\) denote the widths of layers \(\ell\) and \(\ell+1\),
respectively, and where each edge \((i\!\to\! j)\) carries a trainable
B\hbox{-}spline activation \(\phi_{\ell,j,i}: \mathbb{R}\to\mathbb{R}\).
Within a layer, the forward computation is purely
Kolmogorov-like aggregation:
\begin{equation}
    x_{\ell+1,j}
    =
    \sum_{i=1}^{n_\ell}
        \phi_{\ell,j,i}\!\bigl(x_{\ell,i}\bigr),
    \label{eq:KAN_forward_minimal}
\end{equation}
so each layer performs only “inner-type’’ operations: univariate
nonlinearities followed by summation.

\medskip
\noindent
In the deep setting, the outputs of one collection of inner-type
functions become the inputs to the next.
No separate outer function \(\Phi_q\) is introduced: the final layer plays 
the role of an outer map only in the trivial sense that it produces the scalar
output.  
Thus a deep KAN is simply the composition of its inner-type layers,
\begin{equation}
    \mathrm{KAN}(\mathbf{x})
    =
    \bigl(
        \boldsymbol{\Phi}_{L-1}
        \circ
        \cdots
        \circ
        \boldsymbol{\Phi}_{0}
    \bigr)(\mathbf{x}),
    \label{eq:KAN_deep_minimal}
\end{equation}
with “outer-like’’ effects distributed across all layers rather than
concentrated in a single KST-style stage.

\medskip
\noindent
\textbf{Why KANs should not be viewed as exact realizations of the Kolmogorov Superposition Theorem.}  
Although Liu et al.\ describe KANs as \emph{inspired} by the Kolmogorov superposition theorem, it is helpful to distinguish KST and KAN as follows:
\begin{itemize}

    \item \emph{Inner functions are learned, not universal.}  
    In KST the inner maps are universal and fixed for all target functions~\cite{Kolmogorov57,Sprecher65}, whereas in KANs every edge function is trained from data.  
    This points to a structural analogy rather than an exact correspondence.

    \item \emph{Regularity properties differ.}  
    Classical KST inner functions are only Hölder or Lipschitz continuous~\cite{Vitushkin54_nonSmooth,Henkin64_nonSmooth,Fridman67}, whereas KANs use smooth B--splines.  
    This indicates that KANs operate within an approximation framework rather than reproducing the exact KST construction.

    \item \emph{Separation constraints are not enforced.}  
    Constructive proofs of KST rely on fixed shifts and integrally independent weights to separate subcube images~\cite{Sprecher65,Lorentz62}.  
    KANs do not impose these geometric constraints, so they do not replicate the mechanism underlying the theorem.

    \item \emph{Width and depth do not follow KST.}  
    Sternfeld’s optimal width \(2n{+}1\) applies to the shallow superposition setting~\cite{Sternfeld79}.  
    KANs allow arbitrary widths and depths, reflecting modern deep-learning practice rather than KST optimality.

    \item \emph{No final one-dimensional outer function.}  
    KST reduces approximation to a single univariate outer map~\cite{Lorentz62,Sprecher65}.  
    Deep KANs distribute nonlinearity across multiple layers and do not preserve this one-dimensional reduction.

\end{itemize}

\section{Bridging KANs and Kernel Methods}
\label{KANtoKernel}

\noindent
Kolmogorov--Arnold networks constructed from explicit basis functions
admit a direct and instructive comparison with classical kernel methods.
This section establishes a concrete bridge between the two paradigms by
focusing on how basis functions are combined, how coefficients are
parameterized, and how dimensionality is handled.
The comparison shows that a shallow, one-dimensional KAN is mathematically
equivalent to a kernel regression model, while deeper or multi-dimensional
KANs depart from kernel structure in systematic and consequential ways.

\medskip
\noindent
\textbf{Kernel method perspective.}
A classical kernel or basis-expansion method approximates a target function
$f$ by
\[
f(x) \;\approx\; \sum_{k} c_k\,\psi_k(x),
\]
where $\{\psi_k\}$ are fixed basis functions (polynomials, splines, Gaussian
RBFs, Fourier modes, etc.) and $c_k \in \mathbb{R}$ are scalar expansion
coefficients.
Given data $\{(x_n,f_n)\}_{n=1}^N$, one forms the design matrix
\[
\Psi_{n,k} = \psi_k(x_n),
\]
and determines the coefficients $c_k$ by solving the linear system
\[
\Psi\,\mathbf{c} \;\approx\; \mathbf{f},
\]
where $\mathbf{c}$ and $\mathbf{f}$ denote the vectors of expansion
coefficients and sampled function values, respectively.
Once the basis is fixed, the problem is linear, convex, and solved directly
by linear algebra.

\medskip
\noindent
We now compare this structure with KANs in three progressively more complex
settings.

\medskip
\noindent
\textbf{Case 1: Output-only KANs for one-dimensional problems coincide with kernel methods.}

\noindent
Consider a KAN applied to a one-dimensional problem, with a single scalar input,
no hidden layers, and one output neuron.
The network produces an approximation $g$ of the target function $f$, which can
be written as
\[
g(x) = \sum_{k} c_k\,\phi_k(x),
\]
where $\{\phi_k\}$ denotes the chosen univariate basis functions implemented
by the KAN and $\{c_k\}$ are trainable scalar coefficients.
This representation coincides exactly with the standard form of a kernel or
basis-expansion model.

\medskip
\noindent
The interpretation depends only on the selected basis:
\begin{itemize}
\item for spline-based KANs, $\phi_k$ are spline basis functions and $c_k$
      are spline coefficients;
\item for polynomial-based KANs (e.g.\ Chebyshev), $\phi_k$ are orthogonal
      polynomials and $c_k$ are polynomial coefficients;
\item for Gaussian-based KANs, $\phi_k$ are fixed radial basis functions
      and $c_k$ are their amplitudes.
\end{itemize}

\medskip
\noindent
In this output-only setting, the KAN is \emph{linear in its parameters} and spans
the same function class as a classical kernel or basis-expansion model.
The coefficient vector $\mathbf{c}$, however, is obtained differently:
KANs learn $\mathbf{c}$ via gradient-based optimization of the empirical loss,
whereas classical kernel methods typically rely on direct linear-algebraic
procedures such as normal equations, matrix inversion, or orthogonal
projections.
This difference reflects the training strategy rather than the representable
function space.
As shown in the spectral analysis of shallow KANs in \cite{Wang25}, the resulting
loss remains quadratic with a Gram-matrix Hessian, enabling conditioning and
convergence analysis through linear algebra.
Neural-network–specific expressive effects arise only once hidden layers or
multi-dimensional compositions are introduced.

\medskip
\noindent
Therefore, in the absence of hidden layers, KANs applied to
one-dimensional problems are functionally equivalent to classical kernel or
basis-expansion models, with differences arising only from the training
procedure rather than expressivity.

\medskip
\noindent
\textbf{Case 2: Divergence in multiple dimensions.}

\noindent
The equivalence between KANs and kernel methods no longer holds once the input
dimension exceeds one.
Classical kernel and basis-expansion methods construct multivariate features by
explicitly encoding cross-dimensional interactions.
For separable bases, this is achieved through tensor products,
\[
\psi_{k_1,\dots,k_d}(\mathbf{x})
=
\prod_{\ell=1}^{d} \psi_{k_\ell}(x_\ell),
\]
where $(k_1,\dots,k_d)$ is a multi-index and each $\psi_{k_\ell}$ is a fixed
one-dimensional basis function.
For radial basis functions, the dependence is instead on full
$d$-dimensional distances,
\[
\psi_m(\mathbf{x}) = \varphi\!\big(\|\mathbf{x}-\mathbf{x}_m\|\big),
\qquad
\mathbf{x}_m \in \mathbb{R}^d,
\]
where $m$ indexes the kernel centers.
In both cases, multivariate coupling is embedded directly in the basis
functions.

\medskip
\noindent
These constructions are expressive but suffer from the curse of dimensionality.
For polynomial or spline bases of order $p$, the number of tensor-product basis
functions grows as $(p+1)^d$, while for radial kernels the number of required
centers increases rapidly with $d$ in order to resolve multivariate structure
\cite{Schaback05, Wendland04}.

\medskip
\noindent
KANs adopt a different architectural principle.
Given an input $\mathbf{x}=(x_1,\dots,x_d)$, a standard KAN layer computes the
output of neuron $j$ as
\[
z_j
=
\sum_{\ell=1}^{d} \phi_{j,\ell}(x_\ell),
\qquad
\phi_{j,\ell} \in \mathcal{F}_{1\mathrm{D}},
\]
where each $\phi_{j,\ell}$ is an independently parameterized univariate function.
No tensor-product basis functions
$\prod_{\ell} \psi_{k_\ell}(x_\ell)$ and no radial functions
$\varphi(\|\mathbf{x}-\mathbf{x}_m\|)$ are constructed at the level of a single
layer.

\medskip
\noindent
Multivariate interactions arise \emph{only} through the presence of
\emph{hidden layers}.
When a second KAN layer is applied to the outputs of the first, the resulting
composition produces functions that depend jointly on multiple input
coordinates.
With increasing depth, such compositions can approximate interaction terms
resembling tensor-product or radial features, but these interactions are
generated through layer-wise composition rather than encoded explicitly in a
single multivariate basis function.

\medskip
\noindent
This distinction is structural.
As analyzed in \cite{Wang25}, the expressive power of KANs in multiple dimensions
is governed by depth and compositional structure, whereas classical kernel
methods rely on explicit multivariate basis construction.
KANs therefore avoid explicit tensorization and achieve expressivity through
stacked univariate transformations.

\medskip
\noindent
\emph{Interpretation.}
For multivariate problems, KANs can be viewed as structured, low-complexity
alternatives to tensor-product or radial kernel spaces.
They trade explicit enumeration of all cross-dimensional interactions for a
compositional representation that scales more favorably with dimension.
This represents a structurally distinct approach to multivariate function
approximation rather than a direct kernel approximation.

\medskip
\noindent
\textbf{Case 3: Deep KANs and nonlinear coefficient coupling.}

\noindent
As established in Case~2, hidden layers are required for KANs to represent
general multivariate functions.
When such hidden layers are introduced, KANs depart most fundamentally from
classical kernel and basis-expansion methods.
In this deep setting, the coefficients of the resulting function are no longer
independent parameters.

\noindent
This nonlinear coupling arises from function composition across layers and has
several consequences:

\begin{itemize}
\item \textbf{Redundancy.}
      Multiple parameter configurations can represent the same final function,
      leading to flat directions and equivalent minima.
\item \textbf{Loss of linear solvability.}
      Unlike kernel methods, the coefficients cannot be recovered by solving a
      linear system; training requires nonlinear, iterative optimization.
\item \textbf{Entangled representation.}
      Intermediate features generated by hidden neurons are combined by
      subsequent layers, obscuring a direct correspondence between individual
      parameters and the final functional form.
\end{itemize}

\medskip
\noindent
\textbf{Illustrative example.}
The effects described above already appear in the simplest one-dimensional deep
setting and can be understood through two closely related aspects: growth of the
effective polynomial degree and nonlinear coupling of coefficients.

\begin{itemize}
\item \emph{Degree growth.}
Consider a KAN with two Chebyshev layers of degree $P=2$, as in the ChebyKAN
construction~\cite{ChebyKAN}\footnote{\url{https://github.com/SynodicMonth/ChebyKAN}},
but without any regularization.
The first (hidden) layer consists of two neurons, and the second layer produces
a scalar output.
Although each layer employs only quadratic basis functions, their composition
across two layers yields an output polynomial of degree $4$.
More generally, for a one-dimensional KAN with $L$ stacked polynomial layers of
degree $P$, the effective polynomial degree satisfies
\[
\deg(g) = P^{\,L},
\]
where $L$ denotes the number of layers.
This growth is a direct consequence of function composition and does not arise
in shallow models.
In this setting, high-degree terms are generated automatically, and training
must adjust the parameters so that the corresponding coefficients remain close
to zero when the target function does not require such a high polynomial degree.

However, as illustrated in the item below, the same parameters that control
high-degree terms also contribute to lower-degree coefficients, making it
nontrivial for optimization to suppress unwanted high-degree components
without affecting the lower-degree structure of the solution.

\item \emph{Nonlinear coefficient coupling.}
The same compositional mechanism that increases polynomial degree also changes
how the polynomial coefficients depend on the underlying parameters.
Let $A^{(1)}\in\mathbb{R}^{3\times 2}$ and $A^{(2)}\in\mathbb{R}^{6\times 1}$
denote the parameter matrices of the first and second layers, respectively,
with entries
\[
A^{(1)}=\big(a_{ij}\big), \qquad A^{(2)}=\big(b_{ij}\big).
\]

For this two-layer Chebyshev--KAN with per-layer polynomial degree $P=2$,
collecting powers of $x$ yields the equivalent power-form
\[
g(x)=C_0 + C_1 x + C_2 x^2 + C_3 x^3 + C_4 x^4,
\]
where the coefficients are
\[
\begin{aligned}
C_4 &= 8\textcolor{red}{a_{12}}^2 b_{12} + 8a_{22}^2 b_{15},\\
C_3 &= 8a_{11}\textcolor{red}{a_{12}}\, b_{12} + 8a_{21}a_{22} b_{15},\\
C_2 &= 2\textcolor{red}{a_{12}}\,b_{11}
      +\big(8a_{10}\textcolor{red}{a_{12}}+2a_{11}^2-8\textcolor{red}{a_{12}}^2\big)b_{12}
      +2a_{22}b_{14}
      +\big(8a_{20}a_{22}+2a_{21}^2-8a_{22}^2\big)b_{15},\\
C_1 &= a_{11}b_{11} + \big(4a_{10}a_{11}-4a_{11}\textcolor{red}{a_{12}}\big)b_{12}
      +a_{21}b_{14} + \big(4a_{20}a_{21}-4a_{21}a_{22}\big)b_{15},\\
C_0 &= b_{10}+b_{13}-b_{12}-b_{15}
      +(a_{10}-\textcolor{red}{a_{12}})b_{11}
      +2(a_{10}-\textcolor{red}{a_{12}})^2 b_{12}
      +(a_{20}-a_{22})b_{14}
      +2(a_{20}-a_{22})^2 b_{15}.
\end{aligned}
\]

For comparison, a classical Chebyshev approximation truncated at degree~$4$
has coefficients $\mathbf{c}=(c_0,\dots,c_4)^\top$ determined by a linear system
of the form $\Psi\,\mathbf{c}=\mathbf{f}$ (or its least-squares variant),
and admits the power-form
\[
f(x)\;\approx\;
(c_0-c_2+c_4)
+(c_1-3c_3)x
+(2c_2-8c_4)x^2
+(4c_3)x^3
+(8c_4)x^4,
\]
where each polynomial coefficient depends linearly on a single entry of
$\mathbf{c}$.

The distinction is therefore explicit.
In the classical case, the mapping from data to coefficients is linear and
fully described by the design (kernel) matrix $\Psi$.
In the deep KAN, each coefficient $C_k$ is a nonlinear function of parameters
from multiple layers.
Terms such as $\textcolor{red}{a_{12}}$ arise from composition and appear in
several coefficients simultaneously, meaning that a single parameter influences
multiple polynomial orders.
Consequently, increasing depth not only raises the effective polynomial degree,
but also couples the coefficients nonlinearly, making their identification a
fundamentally different problem from solving a linear kernel system, even in
one-dimensional settings.
\end{itemize}

\medskip
\noindent
\textbf{Summary.}
Case~1 establishes that shallow KANs applied to one-dimensional problems coincide
with classical kernel or basis-expansion models, differing only in training
procedure.
Case~2 shows that for multivariate inputs, KANs abandon explicit tensor-product
or radial constructions in favor of factorized, additive representations, and
instead rely on hidden layers and composition to approximate multivariate
interactions.
Case~3 demonstrates that once hidden layers are introduced, depth induces both
multiplicative growth of polynomial degree and nonlinear coupling of
coefficients, fundamentally departing from classical kernel structure.
Together, these mechanisms explain the expressive power of deep KANs as well as
the associated challenges in optimization, conditioning, and parameter
identifiability, motivating kernel-inspired regularization and stability-aware
training strategies.


\section{Bridging KANs and MLPs}
\label{KANtoMLP}
\noindent
MLPs are feedforward neural networks composed of stacked affine layers followed by element-wise nonlinear activations. Given an input vector \( \mathbf{x} \in [0,1]^n \), an MLP with \( L \) layers produces an output \( \hat{f}(\mathbf{x}; \boldsymbol{\theta}) \in \mathbb{R} \) via
\begin{align}
    \mathbf{z}^{(0)} &= \mathbf{x}, \\
    \mathbf{z}^{(\ell)} &= \sigma\!\left( \mathbf{W}^{(\ell)} \mathbf{z}^{(\ell-1)} + \mathbf{b}^{(\ell)} \right), \quad \ell = 1, \dots, L-1, \\
    \hat{f}(\mathbf{x}; \boldsymbol{\theta}) &= \mathbf{W}^{(L)} \mathbf{z}^{(L-1)} + \mathbf{b}^{(L)},
\end{align}
where
\begin{itemize}
    \item \( \sigma(\cdot) \) is a nonlinear activation function (e.g., ReLU, \(\tanh\)),
    \item \( \mathbf{W}^{(\ell)} \in \mathbb{R}^{n_\ell \times n_{\ell-1}} \) and \( \mathbf{b}^{(\ell)} \in \mathbb{R}^{n_\ell} \) are the weights and biases at layer \( \ell \),
    \item \( \mathbf{z}^{(\ell)} \in \mathbb{R}^{n_\ell} \) is the hidden representation at layer \( \ell \),
    \item \( n_\ell \) is the width (number of neurons) of layer \( \ell \),
    \item \( \boldsymbol{\theta} = \{ \mathbf{W}^{(\ell)}, \mathbf{b}^{(\ell)} \}_{\ell=1}^{L} \) collects all trainable parameters.
\end{itemize}

\noindent
KANs and MLPs are both hierarchical function approximators, yet they differ fundamentally in \emph{where} and \emph{how} nonlinearities are applied.  
In a conventional MLP, inputs are first \emph{linearly mixed} and then passed through a fixed activation function (mix \(\rightarrow\) activate).  
In contrast, a KAN applies a (typically trainable) univariate transformation to each input coordinate before aggregation (activate \(\rightarrow\) mix).  
This reversal of operations produces more localized, interpretable, and adaptable mappings.

\medskip
\noindent
This structure closely parallels \emph{Sprecher’s} constructive version of the Kolmogorov superposition theorem~\eqref{eq:KST_modern}, where inner univariate functions act independently on shifted coordinates:
\[
f(x_1,\ldots,x_n)
=\sum_{q}
\Phi\!\biggl(
\smash{\underbrace{\sum_{p} \lambda_p\, \phi(x_p+\eta_q)}_{\substack{\lambda_p:\text{ weights}\\ \eta_q:\text{ shifts}}}}
\;+\;
\smash{\underbrace{c_q}_{\text{offset}}}
\biggr),
\]

\medskip

\bigskip
\noindent
with positive weights \(\lambda_p>0\), shifts \(\eta_q\), and offsets \(c_q\).  
The KAN formulation operationalizes this constructive principle within a neural architecture, turning theoretical decomposition into a learnable process.

\medskip
\noindent
\textbf{Formal Equivalence.}
Wang et al.~\cite{Wang25} establish a precise bidirectional correspondence between
multilayer perceptrons (MLPs) and Kolmogorov--Arnold Networks (KANs), showing that
the two architectures are formally equivalent under mild structural assumptions.

\begin{theorem}[MLP $\Rightarrow$ KAN, {\cite{Wang25}}]
\label{thm:mlp_to_kan}
Let \(\Omega\subset\mathbb{R}^d\) be a bounded domain, and suppose that a function
\(f:\mathbb{R}^d\to\mathbb{R}\) can be represented by an MLP of width \(W\), depth
\(L\), and activation function
\[
\sigma_k(x)=\max(0,x)^k,\qquad k\ge 1.
\]
Then there exists a KAN of width \(W\), depth at most \(2L\), grid size \(G=2\), and
\(k\)-th order B\hbox{-}spline basis functions such that
\[
f(\mathbf{x}) = g(\mathbf{x}), \qquad \forall \mathbf{x}\in\Omega.
\]
\end{theorem}

Concretely, an MLP representation of the form
\[
f(\mathbf{x})
=\sum_{i=1}^{W}
\alpha_i\,
\sigma_k\!\Bigl(
\sum_{j=1}^{d} w_{ij} x_j + b_i
\Bigr)
\]
admits an equivalent KAN parameterization
\[
f(\mathbf{x})
=\sum_{i=1}^{W}
\Phi_i\!\biggl(
\sum_{j=1}^{d} \varphi_{ij}(x_j)
\biggr),
\]
where each univariate map \(\varphi_{ij}:\mathbb{R}\to\mathbb{R}\) is a learned
B\hbox{-}spline that approximates the composite nonlinearity
\(\sigma_k(w_{ij}x_j+b_i)\).
From this perspective, KANs replace fixed pointwise activations in MLPs by
learnable one-dimensional functions along each edge, without increasing the
architectural width.

\medskip
\noindent
The reverse direction holds under mild restrictions on the univariate components.
Non-polynomial nonlinearities (such as SiLU) cannot be captured exactly by
\(\mathrm{ReLU}^k\) networks, but if a KAN employs only polynomial-type univariate
functions—such as B\hbox{-}splines of order \(k\)—then an equivalent MLP
representation exists.

\begin{theorem}[KAN $\Rightarrow$ MLP, {\cite{Wang25}}]
\label{thm:kan_to_mlp}
Suppose that a function \(f:[0,1]^d\to\mathbb{R}\) is represented by a KAN of width
\(W\), depth \(L\), grid size \(G\), and \(k\)-th order B\hbox{-}spline univariate
functions, and assume that no non-polynomial activations are present.
Then there exists an MLP with activation \(\sigma_k(x)=\max(0,x)^k\), depth at most
\(2L\), and width \((G+2k+1)W^2\) that represents \(f\).
\end{theorem}

\noindent
The induced MLP has parameter complexity
\(\mathcal{O}(G^2W^4L)\), whereas the original KAN requires only
\(\mathcal{O}(GW^2L)\) parameters.
This gap highlights the superior parameter efficiency of KANs, particularly for
fine spline grids or higher-order bases.
As a consequence, approximation and convergence results developed for
\(\mathrm{ReLU}^k\) MLPs can be transferred directly to KANs through this formal
equivalence (see Section~\ref{convergence}).

\medskip
\noindent
\textbf{Special Case: Piecewise-Linear Functions.}  
Schoots et al.~\cite{Schoots25} demonstrate that KANs using piecewise-linear univariate functions are functionally identical to ReLU-based MLPs.  
Any piecewise-linear \(\varphi:\mathbb{R}\to\mathbb{R}\) with \(k\) breakpoints admits the expansion
\[
\varphi(x)
=
\underbrace{a_0}_{\text{bias}}
+
\underbrace{a_1 x}_{\text{linear term}}
+
\sum_{j=1}^{k}
\underbrace{\alpha_j}_{\text{coeff.}}\,
\mathrm{ReLU}\!\bigl(x-b_j\bigr),
\]
where \(a_0,a_1,\alpha_j,b_j\in\mathbb{R}\).  
Thus, each KAN univariate map can be embedded into a compact ReLU subnetwork, reinforcing the view of KANs as structured and interpretable MLPs.

\medskip
\noindent
Actor et al.~\cite{Actor25} and Gao et al.~\cite{Gao25} reach similar conclusions:  
KANs can emulate MLP behavior while introducing task-adaptive nonlinearities—such as spline, Fourier, or Chebyshev bases—that improve inductive bias and generalization for structured data.

\medskip
\noindent
\textbf{Summary.}  
KANs and MLPs are expressively equivalent but structurally distinct.  
KANs achieve localized, interpretable representations by applying learnable univariate transformations before aggregation, whereas MLPs rely on fixed activations after linear mixing.  
This reversal leads to substantial parameter savings and smoother function representations, while maintaining theoretical consistency with the Kolmogorov–Arnold superposition principle.  
In practice, KANs can thus be viewed as structured, efficient, and interpretable extensions of MLPs.

\subsection{How KANs Extend the MLP Landscape}
\label{PINNtoPIKAN}

\noindent
KAN layers are increasingly adopted as \emph{drop-in replacements} for traditional MLP blocks across convolutional, transformer, graph, and physics-informed models.  
They enhance expressivity, interpretability, and parameter efficiency—often with minimal architectural or training changes.

\smallskip
\noindent
\textbf{Vision and Representation Learning.}  
In computer vision, \emph{Convolutional KANs} embed spline-based activations directly within convolutional kernels, producing more expressive yet lightweight CNNs~\cite{Bodner24}\footnote{\url{https://github.com/AntonioTepsich/Convolutional-KANs}} and  \cite{Han2026}.  
Residual KAN modules integrate into ResNet backbones to improve gradient flow and generalization~\cite{Yu24_res}\footnote{\url{https://github.com/withray/residualKAN}}, while \emph{U-KAN} extends U-Nets for image segmentation and diffusion models~\cite{Li24b}\footnote{\url{https://github.com/CUHK-AIM-Group/U-KAN}}.  
Similarly, the \emph{KAN-Mixer} adapts MLP-Mixer architectures for image classification~\cite{Cheon24a}\footnote{\url{https://github.com/engichang1467/KAN-Mixer}}.  
Additional applications include remote sensing~\cite{Cheon24}, hyperspectral imaging via \emph{Wav-KAN}~\cite{Seydi24b}, medical image classification~\cite{Chen24}, and autoencoding tasks~\cite{AutoEncoder}\footnote{\url{https://github.com/SekiroRong/KAN-AutoEncoder}}.

\smallskip
\noindent
\textbf{Sequential and Temporal Modeling.}  
For time-series and sequential tasks, \emph{KAN-AD} employs Fourier expansions for efficient anomaly detection~\cite{timeSeries_Zhou24}\footnote{\url{https://github.com/issaccv/KAN-AD}},  
while \emph{T-KAN} and \emph{MT-KAN} enhance forecasting under concept drift and multivariate dependencies~\cite{Xu24}.  
Other extensions include satellite-traffic forecasting~\cite{timeSeries_Rubio24}, recurrent temporal KANs (\emph{TKANs})~\cite{Genet24a}, and the transformer-based \emph{TKAT} model~\cite{Genet24b}\footnote{\url{https://github.com/remigenet/TKAT}}, which integrates learnable univariate mappings within self-attention layers.

\smallskip
\noindent
\textbf{Graph and Structured Data.}  
KAN layers have also been incorporated into graph learning frameworks, where replacing MLPs in GNN message-passing blocks improves numerical stability and feature smoothness.  
Implementations such as \emph{GraphKAN}~\cite{GraphKAN_WillHua}\footnote{\url{https://github.com/WillHua127/GraphKAN-Graph-Kolmogorov-Arnold-Networks}}  
and \emph{KAN4Graph}~\cite{GraphKAN_LiuYue}\footnote{\url{https://github.com/yueliu1999/KAN4Graph}} achieve consistent accuracy gains.  
\emph{GKAN} integrates spline-based kernels directly into graph convolutions~\cite{Kiamari24},  
while general-purpose variants like \emph{S-KAN} and \emph{S-ConvKAN} enable task-dependent activation selection across architectures~\cite{Yang24_Comp}.

\smallskip
\noindent
\textbf{Physics-Informed and Operator Learning.}  
In scientific computing, \emph{Physics-Informed KANs} preserve the original PINN objective but replace fixed activations with learnable basis functions, improving locality and interpretability.  
This modularity enables direct architectural transfers—e.g., \emph{DeepONet} $\rightarrow$ \emph{DeepOKAN}, separable \emph{PINNs} $\rightarrow$ separable \emph{PIKANs}, and even \emph{NTK} analyses in PINNs $\rightarrow$ corresponding analyses in PIKANs.  
Moreover, existing advances such as variational losses, residual reweighting, and adaptive sampling naturally carry over,  
bridging traditional PINN frameworks with modern kernel-based representations.  
Representative examples are summarized in Table~\ref{good_pinns}.

\begin{table}[htbp]
\centering
\scriptsize
\caption{Representative advanced PINN baselines~\cite{Toscano24g,Cuomo22} and corresponding KAN/PIKAN counterparts.}
\label{good_pinns}
\begin{tabular}{@{}l l@{}}
\toprule
\textbf{PINN contribution} & \textbf{KAN/PIKAN counterpart} \\
\textbf{(2017-2025)} & \textbf{(2024-2025)} \\
\midrule
2017—\cite{Raissi19} Foundational PINN for data-driven PDE solutions & \cite{Liu24} \\
2018—\cite{zhang2019quantifying} Uncertainty quantification for forward/inverse PINNs & \cite{pde_bayesian_Giroux24} \\
2019—\cite{Chen19_ode} Neural ordinary differential equations (Neural ODEs) & \cite{pde_Koeing24} \\
2019—\cite{lu2019deeponet} Neural operators for PDE solution maps (DeepONet) & \cite{Abueidda25} \\
2020—\cite{Udrescu20} PINN for symbolic regression via recursive decomposition & \cite{Buhler25_regression} \\
2020—\cite{Meng20} Multi-fidelity PINNs (low/high-fidelity correlation) & \cite{pde_Howard24} \\
2020—\cite{Wang2020_Fourier_nets} Fourier-feature embeddings for multiscale structure & \cite{Zhang25} \\
2020—\cite{Li20_FNO} Fourier Neural Operator (FNO) & \cite{Lee25_operator} \\
2021—\cite{lu2021deepxde} DeepXDE library (resampling strategies and benchmarks) & \cite{Liu24} \\
2022—\cite{Wang22} NTK-based analysis explaining PINN training pathologies & \cite{Gao25,Mostajeran25} \\
2023—\cite{Hou23} Adaptive PINN (moving collocation points) & \cite{pde_Rigas24} \\
2023—\cite{Moseley23} Finite-basis PINNs with overlapping subdomains & \cite{pde_fbkan_Howard24} \\
2023—\cite{cho2023sep} Separable PINN architectures & \cite{pde_jacob24} \\
2023—\cite{Sun23_opt} Surrogate + PDE-constrained optimization & \cite{Yang25} \\
2023—\cite{Eghbalian23} Surrogate modeling in elasto-plasticity & \cite{Mostajeran24} \\
2025—\cite{Song25} Multigrid and multi-resolution training strategy & \cite{Yang25_multiScale}\\
\bottomrule
\end{tabular}
\end{table}



\section{Basis Functions}\label{Basis}
\medskip
\noindent
This section reviews commonly used bases in KANs: B\hbox{-}splines, Chebyshev and Jacobi polynomials, ReLU compositions, Fourier series, Gaussian kernels, wavelets, finite-basis partitions, and Sinc functions. 
We begin with an overview (Table~\ref{AllBases}) to orient the reader.
Subsequent subsections expand on each family, beginning with B\hbox{-}splines.

\begin{table}[hbt!]
\centering
\caption{Comparison of Basis Functions in KANs}\label{AllBases}
\resizebox{\textwidth}{!}{
\begin{tabular}{|c|c|c|c|c|c|}
\hline
\textbf{Name} & \textbf{Support} & \textbf{Equation Form}  & \textbf{Grid} & \textbf{Basis/Activation} & \textbf{Ref.} \\
&  &   & \textbf{Required} & \textbf{Type} &  \\
\hline
B-spline & Local & \( \sum_n c_n B_n(x) \)  & Yes & B-spline & \cite{Liu24} \\
\hline
Chebyshev & Global & \( \sum_k c_k T_k(\tanh x) \)  & No & Chebyshev + Tanh & \cite{SS24} \\
\hline
Stabilized Chebyshev & Global & \( \tanh\!\Big(\sum_{k} c_k\,T_k(\tanh x)\Big) \) & No & Chebyshev + linear head & \cite{Daryakenari25} \\
\hline
Chebyshev (grid) & Global & \( \sum_k c_k T_k\!\Big(\tfrac{1}{m}\sum_i \tanh(w_i x+b_i)\Big) \) & Yes & Chebyshev + Tanh & \cite{Toscano24_kkan} \\
\hline
ReLU-KAN & Local & \( \sum_i w_i R_i(x) \)  & Yes & Squared ReLU & \cite{Qiu24} \\
\hline
HRKAN & Local & \( \sum_i w_i [\mathrm{ReLU}(x)]^m \)  & Yes & Polynomial ReLU & \cite{KAN_pde_So24} \\
\hline
Adaptive ReLU-KAN & Local & \( \sum_i w_i v_i(x) \) & Yes & Adaptive ReLU & \cite{pde_Rigas24} \\
\hline
fKAN (Jacobi) & Global & \( \sum_n c_n P_n(x) \) & No & Jacobi & \cite{Aghaei24_fkan} \\
\hline
rKAN (Padé/Jacobi) & Global & \( \tfrac{\sum_i a_i P_i(x)}{\sum_j b_j P_j(x)} \) & No & Rational + Jacobi & \cite{Aghaei24_rkan} \\
\hline
Jacobi-KAN & Global & \( \sum_i c_i P_i(\tanh x) \) & No & Jacobi + Tanh & \cite{Kashefi25} \\
\hline
FourierKAN & Global & \( \sum_k a_k\cos(kx)+b_k\sin(kx) \) & No & Fourier & \cite{Xu25_fourier} \\
\hline
KAF  & Global & \( \alpha\,\mathrm{GELU}(x)+\sum_j \beta_j \psi_j(x) \) & No & RFF + GELU & \cite{Zhang25} \\
\hline
Gaussian + residual & Local &
$ \sum_i w_i \exp\!\Big(-{\big(\tfrac{x-g_i}{\varepsilon}\big)}^2\Big) + w_b \rho(x) $
& Yes & Gaussian RBF with SiLU & \cite{Li24}\\
\hline
Gaussian & Local &
$ \sum_i w_i \exp\!\Big(-{\big(\tfrac{x-g_i}{\varepsilon}\big)}^2\Big) $
& Yes & Pure Gaussian RBF & \cite{Amir_GKAN, Amir_PUGKAN}\\
\hline
RSWAF-KAN & Local & \( \sum_i w_i \left(s_i - \tanh^2\!\big(\tfrac{x-c_i}{h_i}\big)\right) \) & Yes & Reflectional Switch & \cite{Athanasios2024} \\
\hline
CVKAN & Local & \( \sum_{u,v} w_{uv} \exp\!\big(-|z-g_{uv}|^2\big) \)  & Yes & Complex Gaussian & \cite{Wolff25} \\
\hline
BSRBF-KAN & Local & \( \sum_i a_i B_i(x)+\sum_j b_j \exp\!\big(-\tfrac{(x-g_j)^2}{\varepsilon^2}\big) \)  & Yes & B-spline + Gaussian & \cite{Ta24} \\
\hline
Wav-KAN & Local & \( \sum_{j,k} c_{j,k}\,\psi\!\big(\tfrac{x-u_{j,k}}{s_j}\big) \) & No & Wavelet & \cite{Bozorgasl24} \\
\hline
FBKAN & Local & \( \sum_j \omega_j(x) K_j(x) \)  & Yes & PU + B-spline & \cite{pde_fbkan_Howard24} \\
\hline
SincKAN & Global & \( \sum_i c_i \,\mathrm{Sinc}\!\big(\tfrac{\pi}{h}(x-ih)\big) \) & Yes & Sinc & \cite{Yu24} \\
\hline
Poly-KAN & Global & \( \sum_i w_i P_i(x) \)  & No & Polynomial & \cite{Seydi24a} \\
\hline
\end{tabular}
}
\end{table}

\subsection{B-spline}\label{spline}

\noindent
B\hbox{-}spline bases are among the most widely adopted in KANs due to their compact support, smoothness, local control, and piecewise–polynomial structure~\cite{Actor25, Basina24, Coffman25, Guo25, Kalesh25, Gao25, KAN_pde_Zeng24, Khedr25, Lei25, Li25_DEKAN, Lin25_geo, Pal25, pde_Howard24, pde_jacob24, Aghaei24_kantorol, pde_Patra24, pde_Ranasinghe24, pde_Rigas24, pde_shuai24, pde_Wang24, pde_Zhang24, Raffel25, Schoots25, Wang25, Wang25old, Xu25, Shen25, Yang25, pde_fbkan_Howard24, DeepKAN, Gong25, Guo25_equation, Lee25_operator, Mallick25_battery, Sen25_time}.  
Their locality and flexibility make them expressive, numerically stable, and particularly well suited for interpretable representations. The original KAN formulation~\cite{Liu24} and its accompanying software package further contributed to their early adoption and widespread use.

\medskip
\noindent
Each univariate KAN map \( \varphi:\mathbb{R}\to\mathbb{R} \) is represented as a linear combination of fixed–knot B\hbox{-}spline basis functions:
\begin{equation}
\varphi(x) \;=\; \sum_{n=0}^{N-1} c_n \, B_n^{(k)}(x),
\label{eq_spline}
\end{equation}
where \(N\) is the number of basis functions, \(B_n^{(k)}(x)\) denotes the \(n\)-th B\hbox{-}spline of polynomial degree \(k\) defined on a knot vector \( \mathbf{t}=(t_0,\ldots,t_{N+k}) \), and \( c_n \in \mathbb{R} \) are trainable coefficients. The Cox–de Boor recursion yields \(C^{k-1}\) continuity, and open–uniform knot choices give well–behaved boundary conditions.

\medskip
\noindent
Variants of this formulation also employ fractional B\hbox{-}splines. 
Aghaei and Zaky~\cite{Alireza2026} introduce a physics-informed fractional KAN in which fractional B\hbox{-}spline bases are combined with a tensorized operational-matrix formulation to efficiently solve distributed-order fractional differential equations. 
These variants extend the classical spline basis to fractional operators but retain the same underlying KAN framework.

\begin{figure}[!ht]
\centering%
\subfigure[Non–extended cubic B\hbox{-}spline basis]{ \label{SplineBasis}
\includegraphics[width=2.35in]{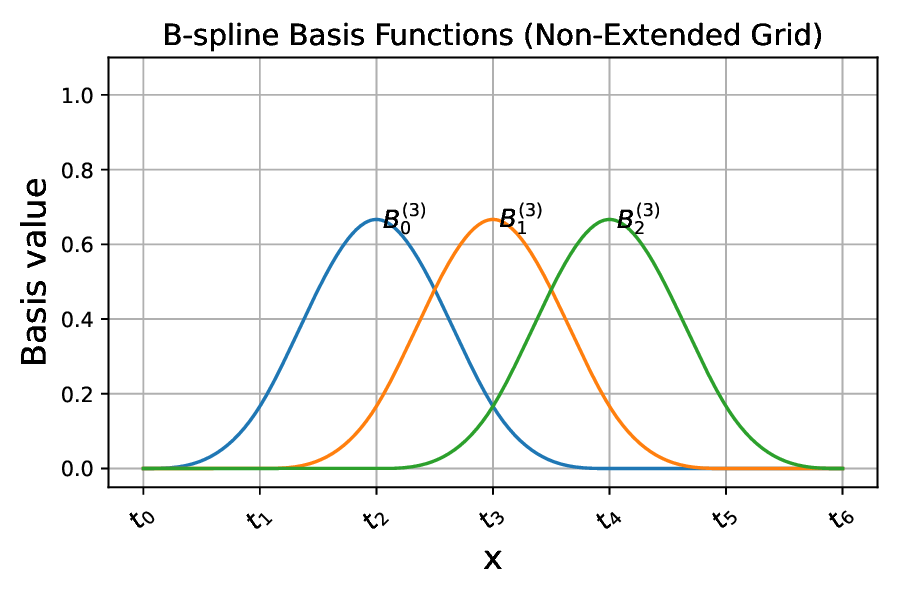}}
\subfigure[Extended cubic B\hbox{-}spline basis]{ \label{SplineBasisExt}
\includegraphics[width=2.35in]{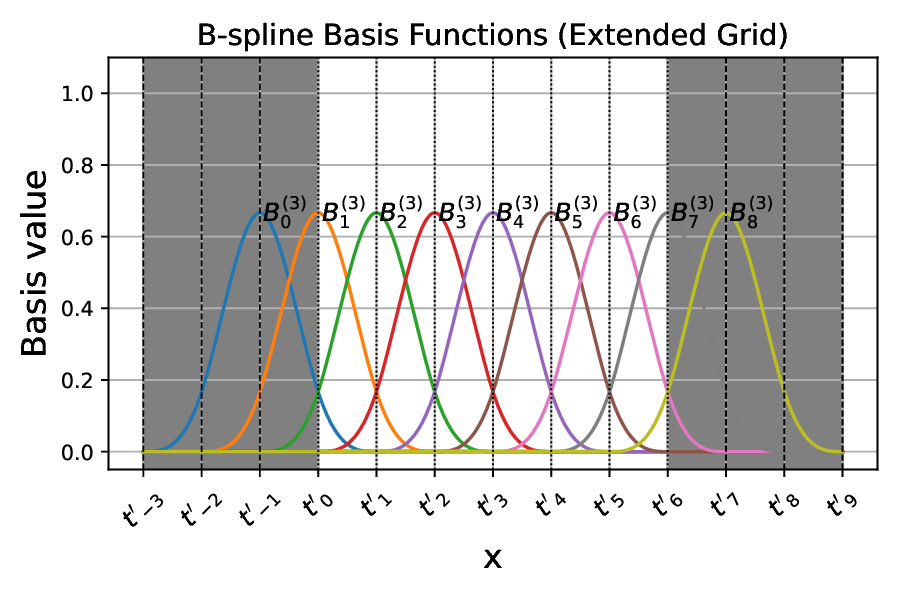}}
\subfigure[Learned map \(\varphi(x)\) on non–extended grid]{ \label{SplineLearned}
\includegraphics[width=2.35in]{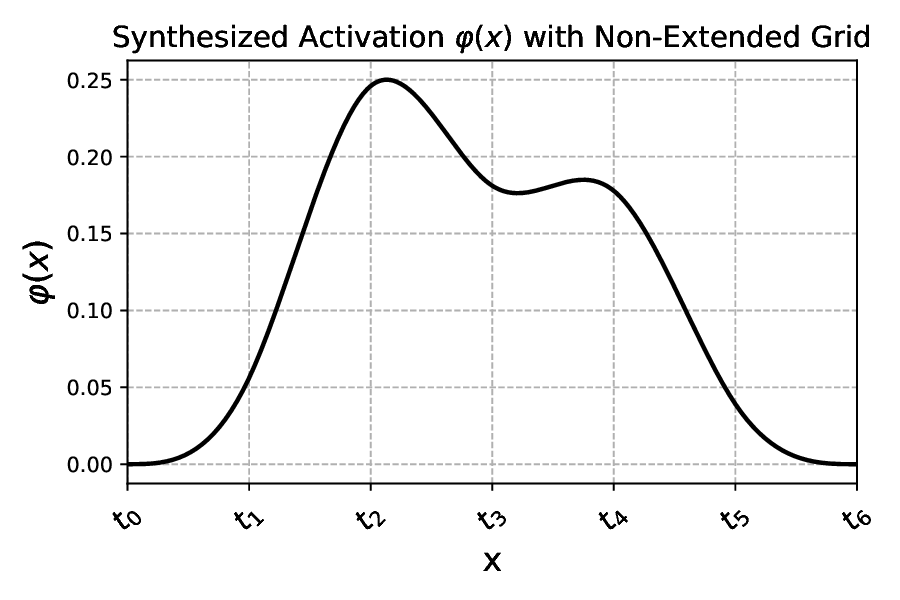}}
\subfigure[Learned map \(\varphi(x)\) on extended grid]{ \label{SplineLearnedExt}
\includegraphics[width=2.35in]{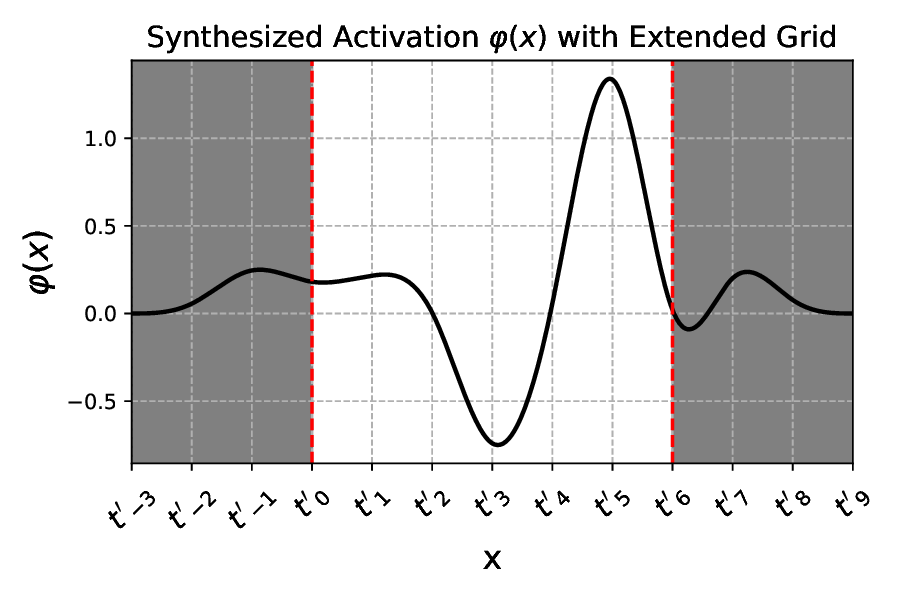}}
\caption{Comparison of B\hbox{-}spline bases and synthesized univariate maps with/without grid extension.
(a,b) Cubic (\(k{=}3\)) bases on non–extended vs.\ extended grids. (c,d) Learned maps \(\varphi(x)\) using identical random coefficients \(c_n\). Gray regions mark padded boundary intervals.}
\label{Spline_Basis}
\end{figure}

\medskip
\noindent
For the standard spline construction used in most KAN implementations, the knot vector is often \emph{extended} by \(k\) points on each side. This padding
allows the model to progressively capture finer details \cite{Wang25}. As illustrated in Figure~\ref{Spline_Basis}, the number of active
basis functions grows from \(G-k\) to \(G+k\), where \(G\) is the number of internal intervals in the original, non–extended grid.

\medskip
\noindent
Subfigures~\ref{SplineBasis} and~\ref{SplineBasisExt} visualize cubic (\(k{=}3\)) B\hbox{-}spline bases on non–extended and extended grids, respectively. Without extension, boundary bases are truncated and
 lose symmetry; with extension, added knots outside the domain restore full polynomial support (gray bands). To compare
  expressivity, subfigures~\ref{SplineLearned}
 and~\ref{SplineLearnedExt} synthesize univariate maps
\[
\varphi(x) \;=\; \sum_{n=0}^{N-1} c_n B_n^{(k)}(x),
\]
using the \emph{same} coefficient vector across the two grids (truncated in the non–extended case). Differences in \( \varphi \)
 thus arise solely from the basis itself: the extended grid yields
 richer boundary behavior and overall smoother maps, while the 
 non–extended grid flattens near edges. This highlights the practical benefit of grid extension for boundary–sensitive tasks (e.g., PDEs).

\medskip
\noindent
In the original KAN~\cite{Liu24} and KAN~2~\cite{Liu24b}\footnote{\url{https://github.com/KindXiaoming/pykan}}, each
univariate activation uses uniformly spaced cubic B\hbox{-}splines
(\(k=3\)) together with a smooth residual term to aid gradients and
enhance expressivity in flat regions:
\[
\tilde{\varphi}(x) \;=\; \sum_{n=0}^{N-1} c_n\, B_n^{(k)}(x) \;+\; \frac{x}{1+e^{-x}}.
\]
A global \texttt{tanh} is applied after each hidden layer (but not the output) to keep activations within the spline domain. 
KAN~2 further supports post--training grid refinement, which increases knot resolution without reinitializing parameters, thereby improving fidelity with minimal overhead. 

\medskip
\noindent\textbf{Convergence implications.}
Recent theory gives spline-KANs some of the strongest guarantees in the KAN family.  
Wang~\cite{Wang25} proves accelerated Sobolev approximation rates; Kratsios et~al.~\cite{Kratsios25} show optimal Besov performance on Lipschitz and even fractal domains; and Liu et~al.~\cite{Liu25_convergence} establish minimax-optimal regression rates together with optimal spline–knot scaling.  
These results place B\hbox{-}splines on the firmest theoretical footing among KAN bases.  
(See Sec.~\ref{convergence}.)

\medskip
\noindent\textbf{Spectral bias and NTK conditioning.}
Spline grids control frequency reach: finer grids reduce spectral bias but may worsen conditioning~\cite{Wang25,Farea25_BasisComp}.  
Initialization also matters—Rigas et~al.~\cite{Rigas25_init}\footnote{\url{https://github.com/srigas/KAN_Initialization_Schemes}} show that Glorot-style and power-law initializations yield more stable NTKs and faster early convergence.  
(See Secs.~\ref{bias} and~\ref{ntk}.)

\subsection{Chebyshev Polynomials}
\label{Cheby}

\noindent
Chebyshev polynomials offer a clean and theoretically grounded alternative to B\hbox{-}splines for KAN univariate mappings—especially in PIKANs—due to their global orthogonality, excellent spectral approximation properties, and simple recursive structure suited for smooth or oscillatory targets.

\medskip
\noindent
\textbf{Definition and Recurrence.}  
Let \(T_k(x)\) denote the Chebyshev polynomials of the first kind, defined recursively as
\begin{equation}\label{chebyEq}
T_0(x)=1,\qquad T_1(x)=x,\qquad T_k(x)=2x\,T_{k-1}(x)-T_{k-2}(x)\quad (k\ge2),
\end{equation}
which form an orthogonal basis on \([-1,1]\) with weight \(w(x)=(1-x^2)^{-1/2}\).

\medskip
\noindent
\textbf{ChebyKAN Formulation.}  
The \emph{ChebyKAN} model~\cite{ChebyKAN}\footnote{\url{https://github.com/SynodicMonth/ChebyKAN}} represents each univariate map as
\[
\varphi^{(\ell)}_{q,p}(x)\;=\;\sum_{k=0}^{K} c^{(\ell)}_{q,p,k}\,T_k\!\big(\tilde{x}\big),
\qquad
\tilde{x}=\tanh(x)\ \ \text{(per-layer input normalization)},
\]
with trainable coefficients \(c^{(\ell)}_{q,p,k}\in\mathbb{R}\).  
Each layer thus evaluates \(T_k\) on normalized inputs \(\tilde{x}\in[-1,1]\) for numerical stability, producing the output
\[
x^{(\ell+1)}_q \;=\; \sum_{p=1}^{P}\varphi^{(\ell)}_{q,p}\!\big(x^{(\ell)}_p\big),
\]
consistent with the generic KAN layer formulation.

\medskip
\noindent
\textbf{Efficient Evaluation.}  
In practice, most implementations (e.g.,~\cite{Yang25_multiScale}) compute \(T_k\) via
\[
T_k(z) = \cos\!\left(k\,\arccos z\right),
\]
which vectorizes efficiently on GPUs. Alternatively, the recurrence relation~\eqref{chebyEq} or Clenshaw’s algorithm provides equivalent, numerically stable computation—particularly advantageous for high-degree expansions where repeated \(\arccos\) calls become costly.  
Mahmoud et al.~\cite{Mahmoud25} further employ \emph{shifted Chebyshev polynomials} on \([0,1]\), defined as
\[
T_n(x)=\cos\!\big(n\,\arccos(2x-1)\big),
\]
to align the domain with the standard Kolmogorov–Arnold representation on non-symmetric intervals.

\begin{figure}[!ht]
\centering
\subfigure[Chebyshev basis without \(\tanh\) normalization]{%
  \label{cheby_basis_NoTanh}
  \includegraphics[width=2.65in]{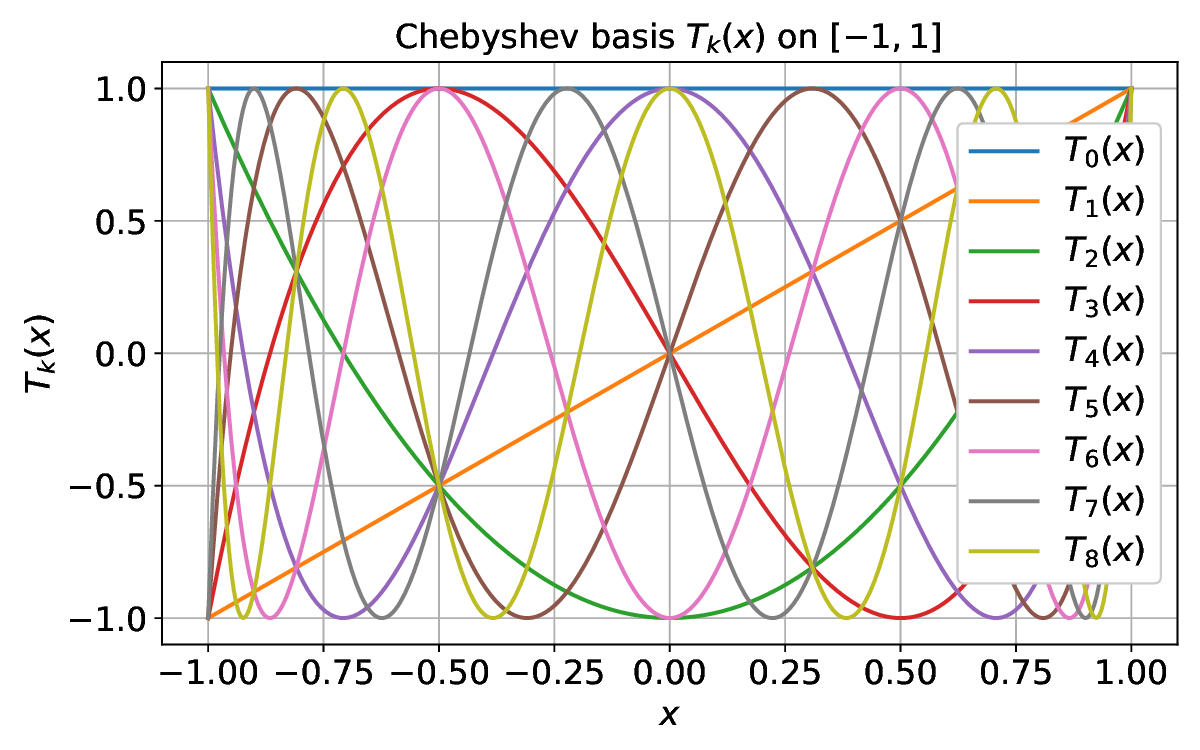}}
\subfigure[Chebyshev basis with \(\tanh\) normalization]{%
  \label{cheby_basis_Tanh}
  \includegraphics[width=2.65in]{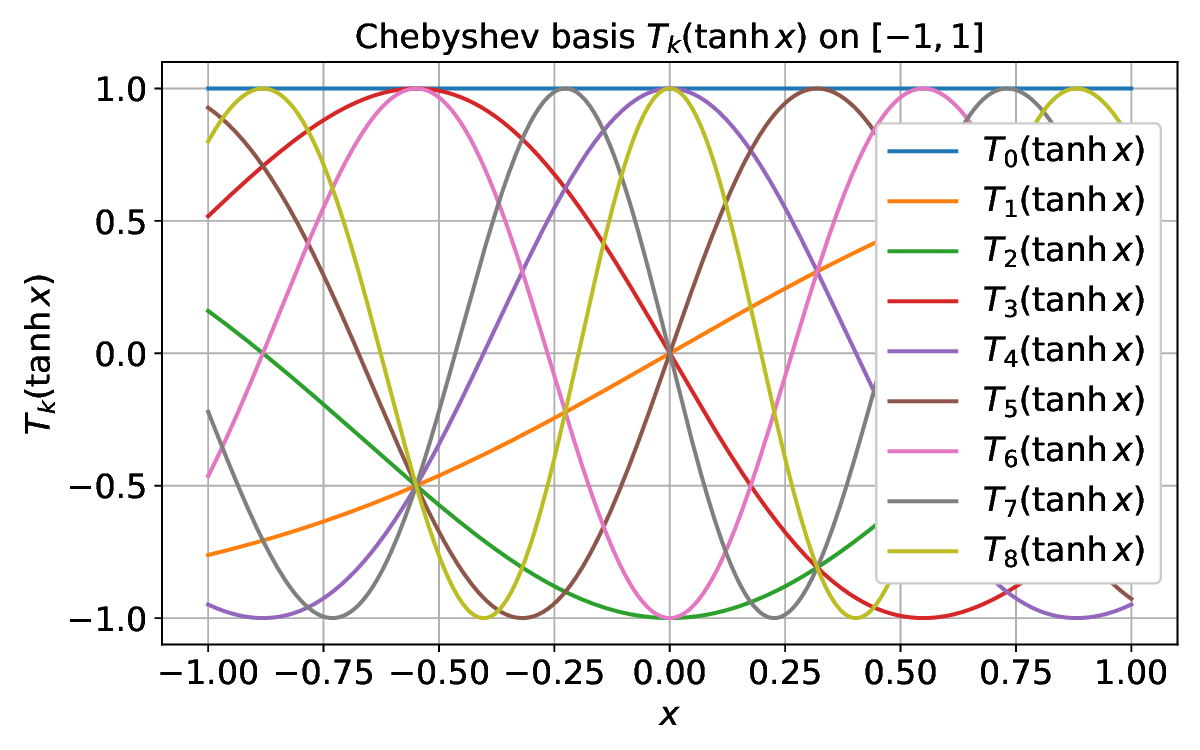}} \\[1em]
\subfigure[Deep Chebyshev KAN map: with vs.\ without \(\tanh\) normalization]{%
  \label{cheby_tanh_vs_plain}
  \includegraphics[width=2.55in]{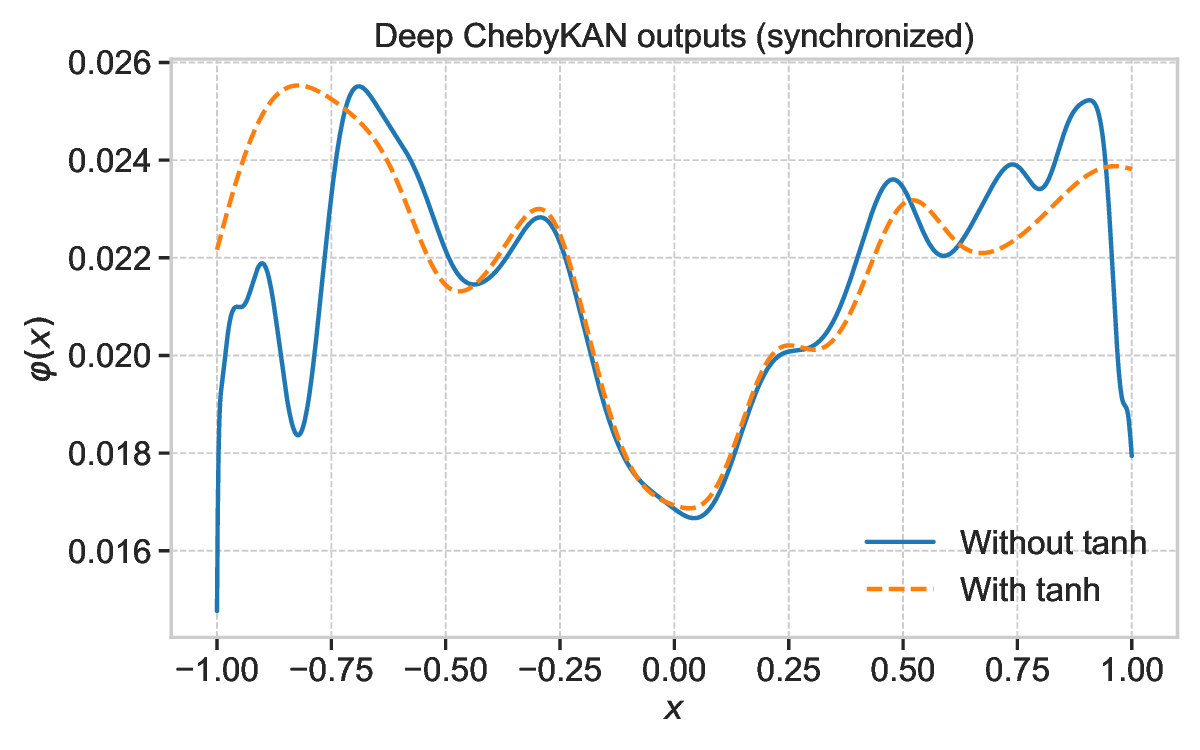}}
\caption{(a) Standard Chebyshev basis functions without per-layer \(\tanh\) normalization.  
(b) Same basis with \(\tanh\) normalization, showing compressed input range and moderated edge slopes.  
(c) Deep Chebyshev KAN map (\(K{=}8\)) comparing \(\sum c_k T_k(\tanh x)\) (blue) vs.\ \(\sum c_k T_k(x)\) (orange); the normalized version exhibits smoother behavior and smaller endpoint gradients, indicating improved conditioning.}
\label{Cheby_Basis}
\end{figure}

\medskip
\noindent
\textbf{Normalization and Stability.}  
Figure~\ref{Cheby_Basis} illustrates how per-layer \(\tanh\) normalization stabilizes Chebyshev activations.  
Panel~(a) shows the standard basis \(T_k(x)\) on \([-1,1]\) for degree \(K{=}8\), where extrema occur at \(T_k(\pm1)=\pm1\), forcing steep slopes near the boundaries.  
Panel~(b) instead evaluates \(T_k(\tanh x)\), compressing the effective range to \(\tanh(\pm1)\approx\pm0.762\), which moderates endpoint slopes while preserving interior structure.  
This compression prevents saturation at \(\pm1\) and keeps features in a well-conditioned range, critical for deep stacks of Chebyshev layers.  

\medskip
\noindent
The difference becomes clearer in Panel~(c), comparing the deep composite maps
\[
\sum_{k=0}^{K} c_k\,T_k(\tanh x)
\quad\text{and}\quad
\sum_{k=0}^{K} c_k\,T_k(x)
\]
with identical coefficients \(c_k\).  
Both are bounded, but by the chain rule,
\[
\frac{d}{dx}\!\left[\sum_{k=0}^{K} c_k\,T_k(\tanh x)\right]
= \bigl(1-\tanh^2 x\bigr)\,\sum_{k=0}^{K} c_k\,T_k'\!\big(\tanh x\big),
\]
the \(\tanh\)-normalized version suppresses slope growth near \(|x|\approx1\), removing the synchronized oscillations visible in the unnormalized case.  
Even without training, this normalization yields smoother and better-conditioned responses.

\medskip
\noindent
\textbf{Grid-Averaged Variant.}  
Following Toscano et al.~\cite{Toscano24_kkan}, a practical enhancement introduces grid before evaluating the Chebyshev expansion.  
For a univariate input \(x\),
\[
g(x)=\frac{1}{m}\sum_{i=1}^{m}\tanh\!\big(w_i x + b_i\big),
\qquad
\varphi^{(\ell)}_{q,p}(x)=\sum_{k=0}^{K} c^{(\ell)}_{q,p,k}\,T_k\!\big(g(x)\big),
\]
where \(m\) denotes the number of centers, \(b_i\) are uniformly spaced biases within \([\beta_{\min},\beta_{\max}]\) (e.g., \(-0.1\) to \(0.1\)), and \(w_i\) control local slope.  
The coefficients \(c^{(\ell)}_{q,p,k}\) can follow the stable initialization proposed in~\cite{SS24}.  
This averaged-\(\tanh\) grid constrains \(g(x)\in(-1,1)\), introduces mild spatial warping, improves numerical conditioning, and as reported in~\cite{Toscano24_kkan}, supports stable training even with higher polynomial degrees.

\medskip
\noindent
\textbf{Spectral Properties and Efficiency.}  
The spectral parameterization of Chebyshev-based KANs was first formalized in Sidharth et al.~\cite{SS24}, while Guo et al~\cite{Guo24} demonstrated superior parameter efficiency and generalization in data-scarce regimes.  
From a theoretical perspective, Faroughi and Mostajeran~\cite{Faroughi25} showed that Chebyshev PIKANs (cPIKANs) yield better-conditioned NTKs with slower spectral decay, accelerating convergence for PDEs such as diffusion and Helmholtz equations.

\medskip
\noindent
\textbf{Stabilization and Hybrid Designs.}  
Yu et al.~\cite{Yu24} confirmed these advantages for function approximation and PDE learning but also highlighted failure modes of raw polynomial stacks at high depth, motivating stabilizers such as domain normalization, nested nonlinearities, and contractive mappings.  
Building on these insights, Daryakenari et al.~\cite{Daryakenari25} proposed a stabilized Chebyshev stack by inserting an additional \(\tanh\) between layers and replacing the Chebyshev head with a linear readout:
\begin{equation}
x^{(\ell+1)}_q \;=\; \tanh\!\Bigg(\sum_{p=1}^{P}\sum_{k=0}^{K} c^{(\ell)}_{q,p,k}\,T_k\!\big(\tanh(x^{(\ell)}_p)\big)\Bigg),
\qquad \ell=0,\dots,L-1,
\label{eq:cheby_tanh_between}
\end{equation}
with the network output
\begin{equation}
\hat{f}(\mathbf{x}) \;=\; \mathbf{W}^{\text{out}}\,\mathbf{x}^{(L)} + \mathbf{b}^{\text{out}},
\label{eq:linear_head}
\end{equation}
where \(\mathbf{x}^{(L)}=[x^{(L)}_1,\dots,x^{(L)}_{H_L}]^\top\).  
The inter-layer \(\tanh\) in~\eqref{eq:cheby_tanh_between} acts as a contraction, curbing gradient growth~\cite{Yu24}, while the linear head~\eqref{eq:linear_head} isolates the final mapping from additional polynomial expansions, improving stability in inverse and PDE learning tasks~\cite{Daryakenari25}.

\medskip
\noindent
\textbf{Depth Stabilization.}
Deep Chebyshev KANs often become unstable.  
Rigas et al.~\cite{Rigas25_deep}\footnote{\url{https://github.com/srigas/RGA-KANs}} mitigate this with a \emph{basis-agnostic Glorot-style initialization} that preserves activation variance, and a \emph{Residual–Gated Adaptive (RGA) KAN} block that stabilizes deep cPIKANs on PDEs.  
\emph{One-liner: variance-preserving init + residual gating enable deep Chebyshev KANs to train without divergence.}

\medskip
\noindent
\textbf{Summary.}  
Chebyshev-based KANs combine spectral approximation theory with practical stability mechanisms.  
Their global orthogonality and efficient recursion yield compact and interpretable representations, while nested nonlinearities and grid-averaged normalization ensure stable deep training.  
These features make them particularly well suited for scientific computing, operator learning, and inverse problems—complementing and often surpassing spline-based KANs within the broader KAN framework.

\subsection{ReLU}
\label{Relu}

\noindent
ReLU–based KANs (ReLU\hbox{-}KANs) were introduced by
 Qiu et al.~\cite{Qiu24}\footnote{\url{https://github.com/quiqi/relu_kan}} as a hardware–efficient alternative to
  B\hbox{-}spline KANs. The key idea is to replace 
  spline activations with compactly supported, bell–shaped functions constructed from ReLU compositions. This preserves 
  the localized, compositional spirit of 
  Kolmogorov–Arnold models while enabling fast, GPU–friendly primitives.

\begin{figure}[!ht]
\centering%
\subfigure[Step-by-step construction of ReLU basis]{%
  \label{Relu5Figs}
  \includegraphics[width=6.55in]{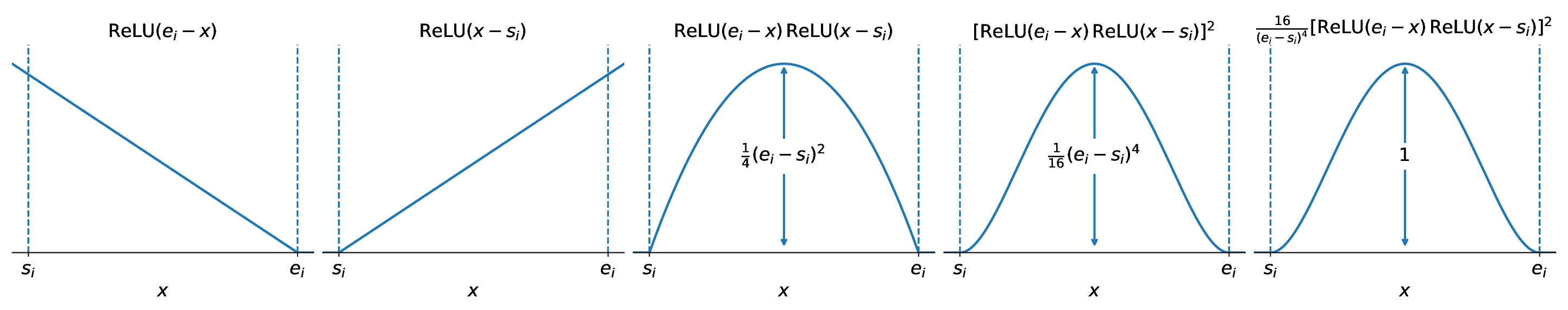}}
\subfigure[Complete set of normalized ReLU–KAN basis functions]{%
  \label{Relu4Basis}
  \includegraphics[width=4.55in]{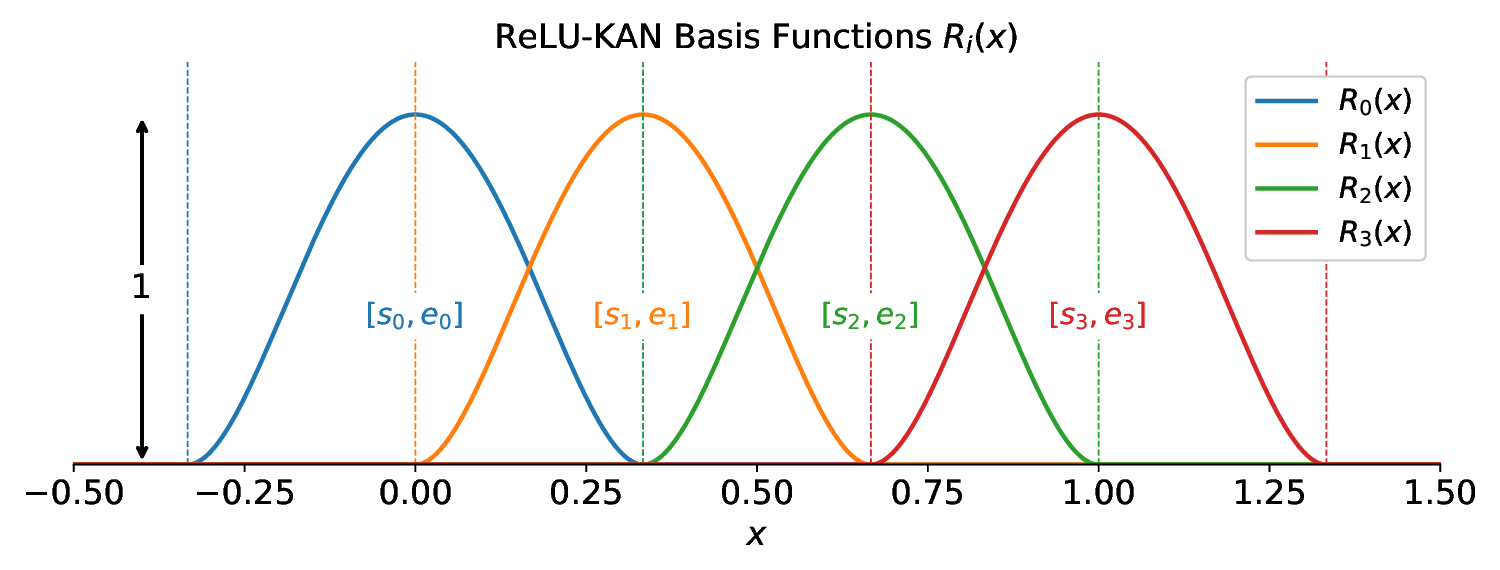}}
\caption{(a) Step-by-step construction of the normalized ReLU-based local basis function from its constituent $\operatorname{ReLU}(e_i - x)$ and $\operatorname{ReLU}(x - s_i)$ terms. 
(b) Complete set of normalized ReLU–KAN basis functions $R_i(x)$ with supports $[s_i, e_i]$~\cite{Qiu24}.}
\label{Spline_Basis_Relu}
\end{figure}

\medskip
\noindent
\textbf{Local ReLU bases and univariate maps.}
In ReLU\hbox{-}KANs, each univariate map \( \varphi:\mathbb{R}\to\mathbb{R} \) is a weighted sum of compact local bases:
\[
\varphi(x) \;=\; \sum_{i=0}^{G+k-1} w_i\, R_i(x),
\]
where \(w_i\in\mathbb{R}\) are trainable weights, \(G\) is the number of grid intervals, and \(k\) controls overlap among neighboring bases. Each \(R_i\) is supported on \([s_i,e_i]\) with uniformly spaced endpoints
\[
s_i \;=\; \frac{i-k}{G}, 
\qquad
e_i \;=\; s_i + \frac{k+1}{G},
\quad i=0,\ldots,G+k-1.
\]
Inside its support, \(R_i\) has a smooth bell shape (Figure~\ref{Spline_Basis_Relu}):
\[
R_i(x)=
\begin{cases}
0, & x < s_i,\\[2pt]
\big((x-s_i)(e_i-x)\big)^2\,\dfrac{16}{(e_i-s_i)^4}, & s_i \le x \le e_i,\\[6pt]
0, & x > e_i,
\end{cases}
\]
which can be written equivalently using squared ReLUs:
\[
R_i(x)
\;=\;
\Big[\operatorname{ReLU}(e_i-x)\,\operatorname{ReLU}(x-s_i)\Big]^2
\cdot \frac{16}{(e_i-s_i)^4},
\qquad 
\operatorname{ReLU}(x)=\max(0,x).
\]

\medskip
\noindent
\textbf{Speed and a smoothness caveat.}
ReLU\hbox{-}KANs often train \(5\text{--}20\times\) faster than spline variants in practice~\cite{Qiu24}, but the squared–ReLU construction has limited smoothness, which can hinder PDE tasks requiring higher–order derivatives.

\medskip
\noindent
\textbf{Higher–order ReLU KAN (HRKAN).}
So et al.~\cite{KAN_pde_So24} generalize the squared–ReLU basis by introducing \emph{local} powers of ReLU. 
For a finite interval \([s_i,e_i]\), the basis function of order \(m\) is
\[
v_{m,i}(x)
\;=\;
\Big[\operatorname{ReLU}(x-s_i)\,\operatorname{ReLU}(e_i-x)\Big]^{\,m}
\cdot
\left(\frac{2}{\,e_i - s_i\,}\right)^{\! 2m},
\]
where \(\operatorname{ReLU}(x) = \max(0,x)\) and \(m\in\mathbb{Z}_{+}\) controls the interior smoothness: \(v_{m,i}\in C^{m-1}\). 
Larger \(m\) produces lobes that are more peaked in the interior and decay more smoothly to zero at the boundaries, 
leading to higher continuity of derivatives and potentially better performance for PDEs that require smooth high-order derivatives.

\begin{figure}[!ht]
\centering
\subfigure[Square–of–ReLU basis]{\label{Relu_SqureRelu}\includegraphics[width=2.55in]{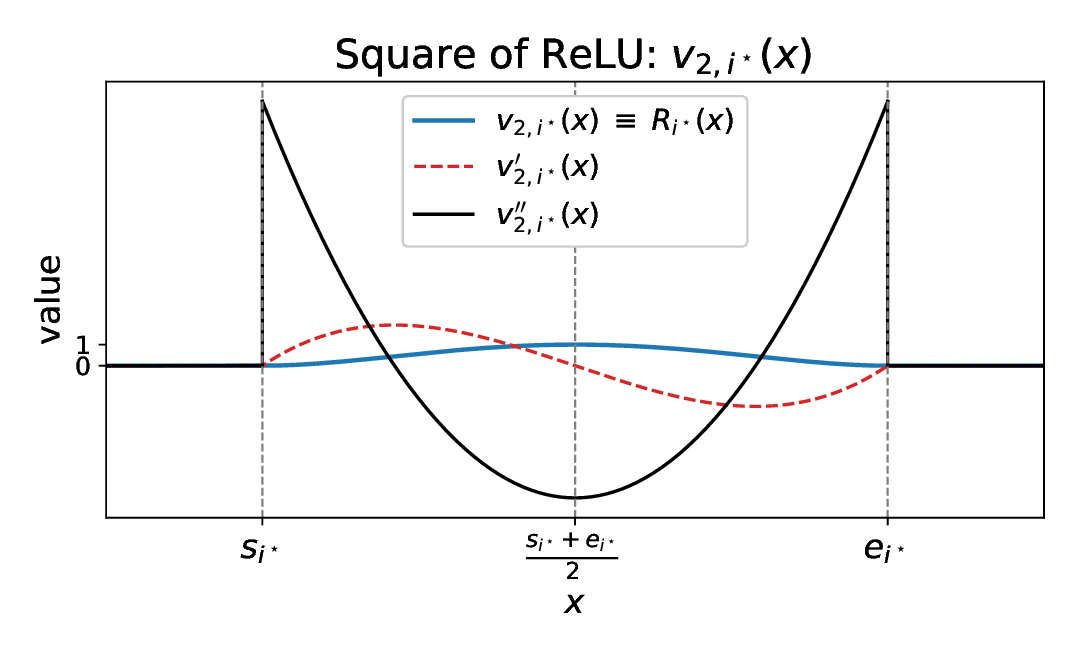}}
\subfigure[Higher–order ReLU basis ]{\label{Relu_HRELU}\includegraphics[width=2.55in]{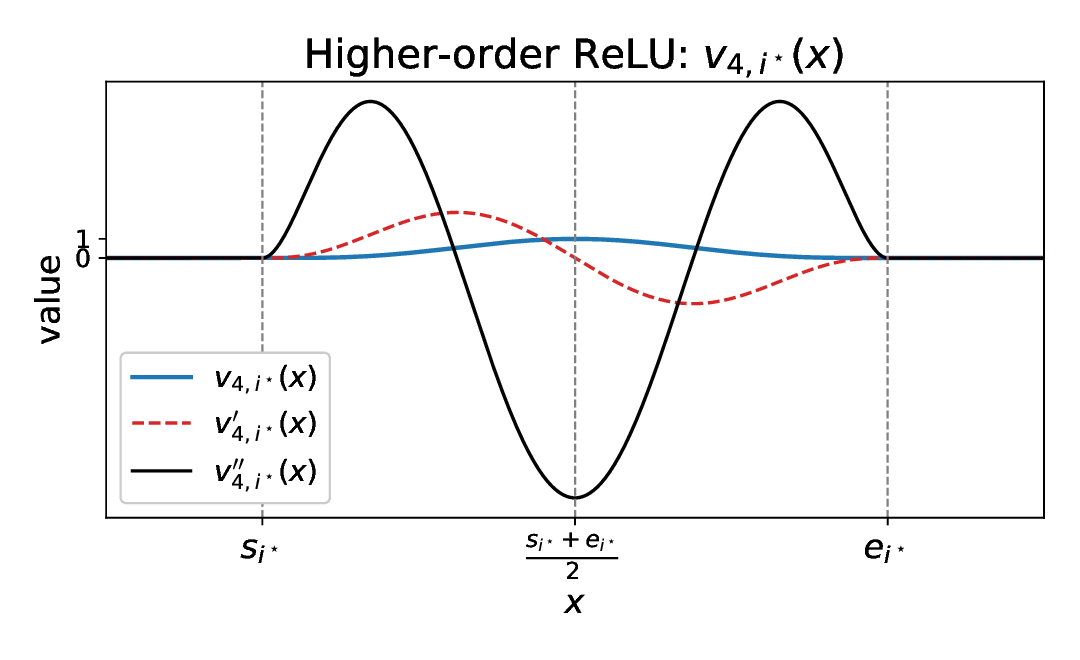}}
\caption{Symbolic basis on $[s_{i^\star},e_{i^\star}]$: (a) $v_{2,i^\star}(x)$ with first and second derivatives; (b) $v_{4,i^\star}(x)$ with first and second derivatives. Observation: $v_{m,i}$ is globally $C^{m-1}$ (with the $m$-th derivative discontinuous at $s_{i^\star},e_{i^\star}$); hence $v_{4,i^\star}$ ($C^3$) offers smoother higher-order derivatives than $v_{2,i^\star}$ ($C^1$)~\cite{KAN_pde_So24}.}
\label{ReluSquared}
\end{figure}

\medskip
\noindent
Figure~\ref{ReluSquared} compares individual basis functions \(v_{m,i^\star}\) over \([s_{i^\star},e_{i^\star}]\). 
Panel~(a) shows the squared–ReLU case (\(m=2\)), which is \(C^1\) but exhibits visible jumps in the second derivative at the boundaries. 
Panel~(b) shows a higher–order ReLU with \(m=4\), which is \(C^3\) and decays smoothly to zero, eliminating derivative discontinuities. 
This single-basis view makes clear how increasing \(m\) improves both interior smoothness and boundary regularity.

\begin{figure}[hbt!]
\centering
\includegraphics[width=0.55\textwidth]{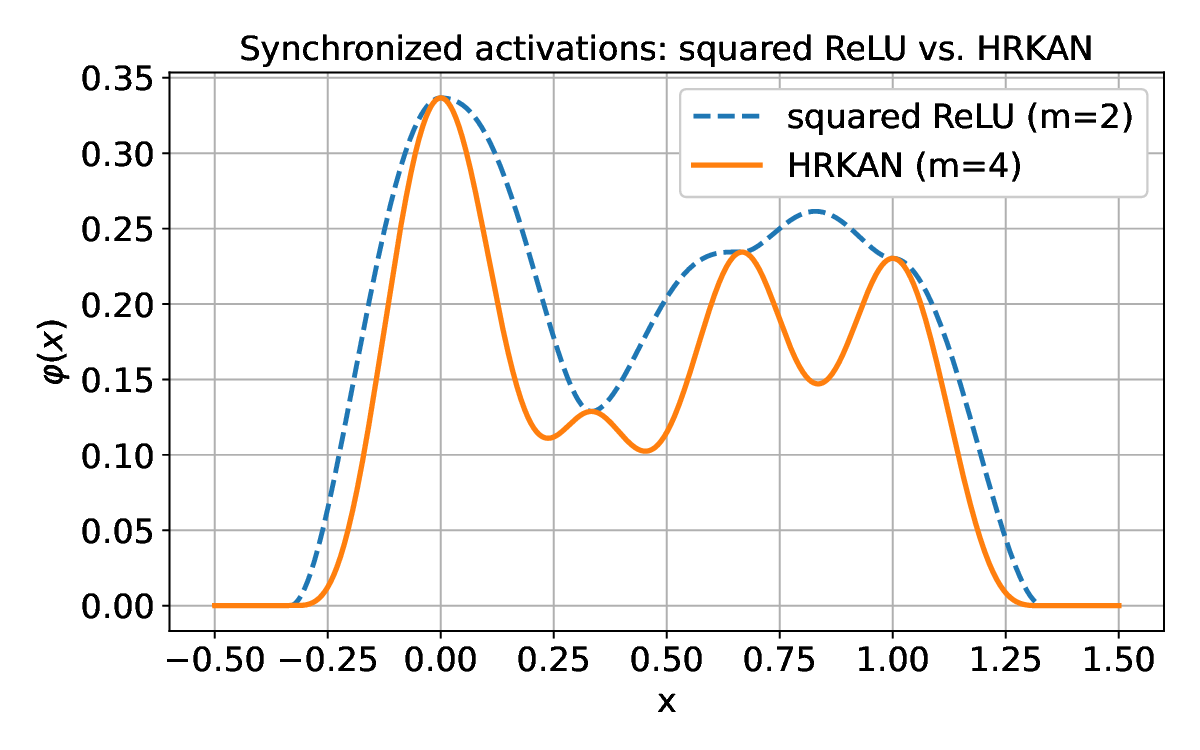}
\caption{Synthesized activation \( \varphi(x) \) from ReLU\hbox{-}KAN bases using shared random coefficients. 
Dashed: squared–ReLU basis (\(m=2\)), Solid: higher–order ReLU basis (\(m=4\)). 
Higher order yields sharper, smoother lobes while preserving the same coefficient structure.}
\label{relu_kan_basis_Learned}
\end{figure}

\medskip
\noindent
Figure~\ref{relu_kan_basis_Learned} compares synthesized activations
\[
\varphi(x) \;=\; \sum_{i} c_i\,v_{m,i}(x),
\]
constructed from the two basis types using \emph{identical random coefficients} (\(G=3\), \(k=1\)) over support of \([s_{i^\star},e_{i^\star}]=[-0.6,1]\). 
Because the coefficients are shared, differences between the dashed (\(m=2\)) and solid (\(m=4\)) curves arise solely from the change in \(m\). 
The higher-order case produces narrower, more sharply peaked bumps with smoother interior derivatives, 
while the squared–ReLU case yields broader lobes with lower differentiability. 
Together with Figure~\ref{ReluSquared}, this demonstrates both the \emph{local} and \emph{global} effects of the order parameter \(m\) in HRKAN.

\medskip
\noindent
Both ReLU\hbox{-}KAN and HRKAN originally used fixed, uniformly spaced grids. To add adaptivity, Rigas et al.~\cite{pde_Rigas24}\footnote{\url{https://github.com/srigas/jaxKAN}} define bases over a nonuniform, data–dependent grid \( \mathcal{G}=\{x_0,\ldots,x_G\} \). With neighborhood parameter \(p\),
\[
s_i \;=\; \mathcal{G}[i] - \tfrac12\big(\mathcal{G}[i+p]-\mathcal{G}[i-p]\big),
\qquad
e_i \;=\; 2\mathcal{G}[i] - s_i.
\]
Adaptive widths and centers improve resolution in regions with singularities, steep gradients, or boundary layers; their \texttt{jaxKAN} implementation also uses resampling and loss reweighting for heterogeneous PDEs.

\subsection{Jacobi and General Polynomials}
\label{Jacobi}

\noindent
Polynomial families constitute one of the most general and historically established classes of basis functions in KANs.  
Beyond splines and Chebyshev polynomials, a comprehensive benchmark by Seydi~\cite{Seydi24a}\footnote{\url{https://github.com/seydi1370/Basis_Functions}} systematically evaluated \emph{eighteen} distinct polynomial families as KAN activation functions, offering a unified comparison across orthogonal, recurrence-based, and rational constructions.  
Among all tested variants, the \emph{Gottlieb} polynomial achieved the highest accuracy and stability metrics on the MNIST benchmark.  
The surveyed families can be grouped according to their mathematical origin:

\begin{itemize}
    \item \textbf{Classical and General Orthogonal Polynomials:}  
    Charlier and Gottlieb (discrete orthogonal), Boas--Buck and Boubaker (generalized continuous families encompassing Hermite, Laguerre, and related forms).

    \item \textbf{Advanced Orthogonal Polynomials (Askey Scheme \& Related):}  
    Askey--Wilson and Al--Salam--Carlitz (\(q\)-orthogonal series), Bannai--Ito (Racah generalization).

    \item \textbf{Recurrence-Based and Number-Theoretic Polynomials:}  
    Tribonacci, Tetranacci, Pentanacci, Hexanacci, Heptanacci, and Octanacci (generalized Fibonacci-type recurrences); Fermat, Vieta--Pell, and Narayana (number-theoretic families).

    \item \textbf{Rational Constructions:}  
    Padé approximants~\cite{Aghaei24_rkan}, explicitly adopted in the rational Jacobi network (rKAN).
    
\item \textbf{Holomorphic Monomial Polynomials:}  
Simple powers $(1, z, \dots, z^P)$, where $z = x + i\,y$ denotes the complex input, as used in PIHKAN~\cite{Clafa25}.

\end{itemize}

\noindent
Overall, Seydi’s benchmark revealed that KAN layers can flexibly host a wide variety of polynomial structures beyond splines, broadening the design space for architectures targeting spectral, combinatorial, or rational approximation behaviors.  
Among these, orthogonal and number-theoretic polynomials—particularly Gottlieb—exhibited the best numerical conditioning and convergence stability.  
These insights motivate the deeper exploration of Jacobi-type formulations discussed below, since Jacobi, Legendre, and Chebyshev families together form the classical orthogonal polynomial hierarchy within the Askey scheme.

\begin{figure}[!ht]
\centering
\includegraphics[width=0.65\textwidth]{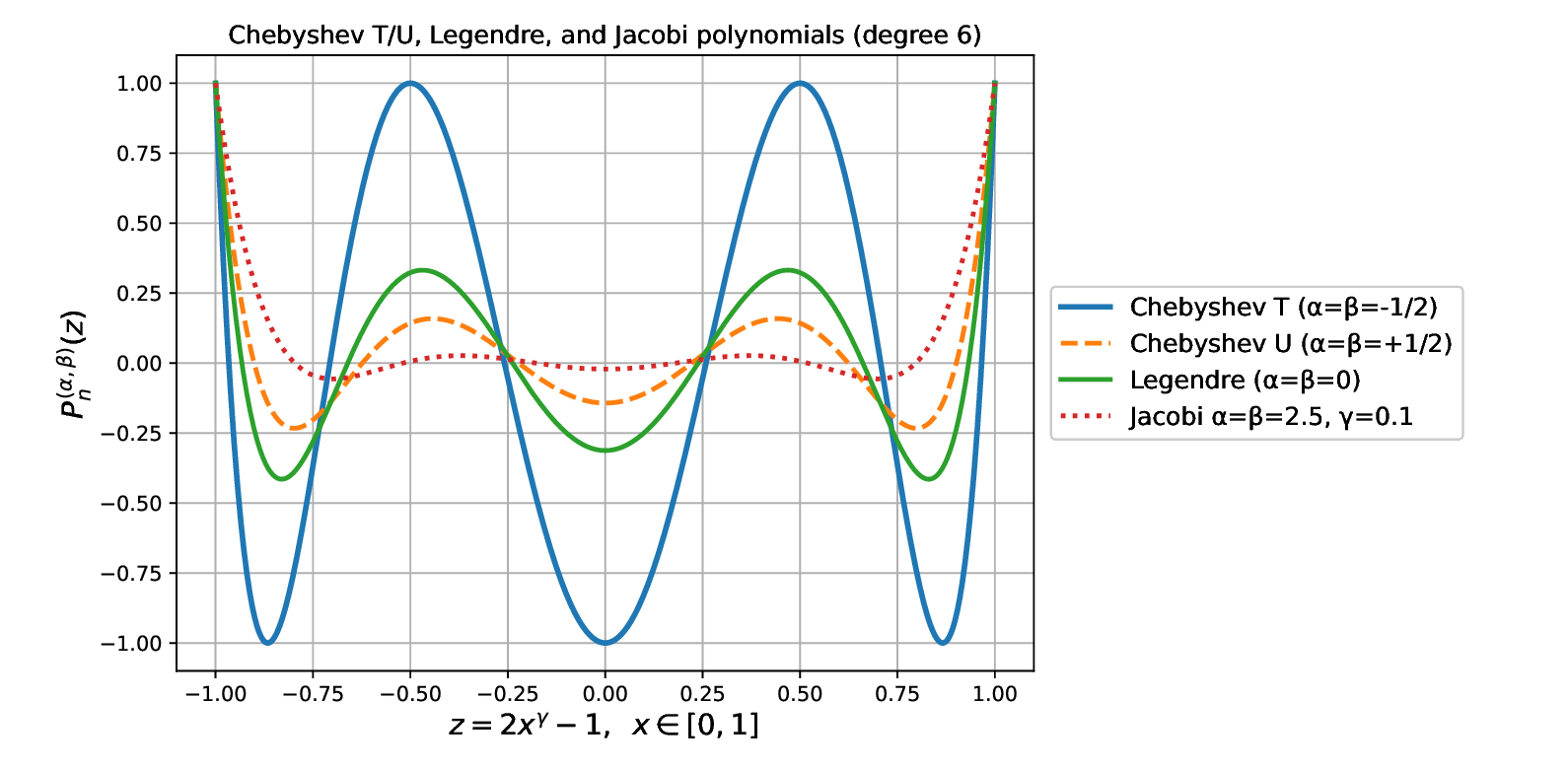}
\caption{Degree-\(n=6\) polynomials on \(z\in[-1,1]\):  
Chebyshev \(T_6(z)\), Chebyshev \(U_6(z)\), Legendre \(P_6(z)\), and Jacobi \(P_6^{(2.5,2.5)}(z)\).  
The Chebyshev--\(T\) curve coincides with \(T_6(z)\) in Fig.~\ref{cheby_basis_NoTanh}.}
\label{JacobiPolynomials}
\end{figure}

\medskip
\noindent
\textbf{Fractional Jacobi Basis (fKAN).}  
To enhance smoothness, domain flexibility, and adaptivity, Aghaei introduced the \emph{Fractional KAN (fKAN)}~\cite{Aghaei24_fkan}\footnote{\url{https://github.com/alirezaafzalaghaei/fKAN}}, employing fractional-order Jacobi polynomials as trainable univariate maps:
\[
P_n^{(\alpha,\beta)}(z_\gamma),
\qquad 
z_\gamma = \phi_\gamma(x) = 2\,x^{\gamma}-1,\quad x\in[0,1],
\]
where \(\alpha,\beta>-1\) are Jacobi exponents and \(\gamma>0\) is a fractional warp parameter controlling the domain stretch.  
The basis remains orthogonal on the canonical interval \([-1,1]\), and different \((\alpha,\beta)\) recover classical cases: Legendre for \((0,0)\), Chebyshev of the first kind for \((-\tfrac12,-\tfrac12)\), and Chebyshev of the second kind for \((\tfrac12,\tfrac12)\), as illustrated in Fig.~\ref{JacobiPolynomials}.

\noindent
In fKAN, the univariate activation is expressed as
\[
\varphi(x)\;=\;P_n^{(\alpha,\beta)}\!\big(\,\phi_\gamma(\,x(r)\,)\,\big),
\]
where \(r\) is the raw input, \(x(r)\in[0,1]\) denotes a normalization map (linear, sigmoid, or \(\tfrac12(1+\tanh r)\)), and \(\phi_\gamma(x)=2x^\gamma-1\) applies fractional warping.  
The parameters \(\alpha,\beta,\gamma\) are trainable, and positivity of \(\alpha,\beta\) is enforced via ELU or sigmoid reparameterizations.  
This yields a globally smooth and tunable basis that performs effectively on regression and PDE benchmarks.  
Kashefi~\cite{Kashefi25} implements fKAN using the \(\tanh\)-based normalization \(x(r)=\tfrac12(1+\tanh r)\) for stable input scaling.

\begin{figure}[!ht]
\centering
\includegraphics[width=0.55\textwidth]{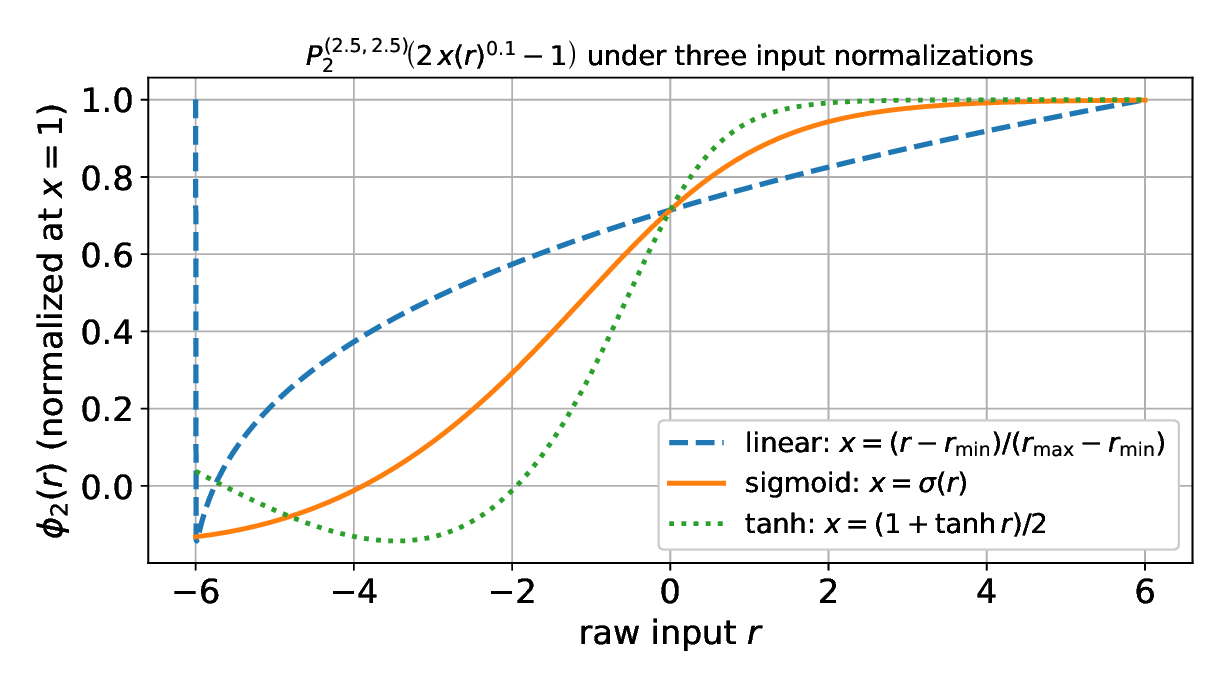}
\caption{Degree-\(n=2\) Jacobi polynomial with \((\alpha,\beta)=(2.5,2.5)\), \(\gamma=0.1\):  
\(\varphi(r)=P^{(\alpha,\beta)}_2\!\big(2\,x(r)^{\gamma}-1\big)\) under three normalizations \(x(r)\in[0,1]\):  
linear \(x=(r-r_{\min})/(r_{\max}-r_{\min})\) (dashed), sigmoid \(x=\sigma(r)\), and \(\tanh\)-based \(x=\tfrac12(1+\tanh r)\).  
All curves are rescaled by \(\varphi(r{=}1)\) for comparability.}
\label{fig:Jacobi_f2_input_norms}
\end{figure}

\medskip
\noindent
\textbf{Effect of Input Normalization.}  
Implementations typically evaluate \(P_n^{(\alpha,\beta)}(z_\gamma)\) with \(z_\gamma=2\,x(r)^\gamma-1\).  
Figure~\ref{fig:Jacobi_f2_input_norms} illustrates how the normalization choice shapes the activation.  
Fixing \((\alpha,\beta)=(2.5,2.5)\), \(n=2\), and \(\gamma=0.1\), a linear mapping preserves dynamic range but produces steep edge slopes, whereas sigmoid and \(\tfrac12(1+\tanh r)\) compress the tails, improving conditioning.  
A smaller \(\gamma<1\) increases the warp toward \(+1\), further sharpening features near that end of the domain.

\medskip
\noindent
\textbf{Rational Extensions (rKAN).}  
To further expand expressivity, Aghaei proposed the \emph{Rational KAN (rKAN)}~\cite{Aghaei24_rkan}\footnote{\url{https://github.com/alirezaafzalaghaei/rKAN}}, which generalizes fKAN by incorporating rational Jacobi compositions.  
Two representative variants are:

\begin{itemize}
\item \textbf{Padé--rKAN:} a rational quotient of Jacobi expansions,
\[
\varphi(x)
=
\frac{\sum_{i=0}^{p} w_i^{(P)} P_i^{(\alpha_p,\beta_p)}\!\big(\phi(\sigma(x))\big)}
{\sum_{j=0}^{q} w_j^{(Q)} P_j^{(\alpha_q,\beta_q)}\!\big(\phi(\sigma(x))\big)},
\]
where \(w_i^{(P)}\) and \(w_j^{(Q)}\) are trainable weights, and \(\sigma(x)\) is an activation function.

\item \textbf{Jacobi--rKAN:} a rational domain warping approach,
\[
\varphi(x)
=
P_n^{(\alpha,\beta)}\!\big(\phi(x;\iota)\big),
\]
with \(\phi(x;\iota)\) representing a rational transform parameterized by \(\iota>0\) (often implemented using SoftPlus for positivity).
\end{itemize}

\noindent
Both fKAN and rKAN exploit Jacobi polynomials with adaptive shape and domain parameters:  
fKAN emphasizes fractional-order smoothness and controlled warping, while rKAN introduces rational expressivity and sharper nonlinear transitions.

\medskip
\noindent
\textbf{Classical Fixed-Degree Jacobi KANs.}  
Several works retain fixed polynomial degrees (\(\gamma=1\)) and use only \(\tanh\) input compression into \([-1,1]\).  
Each univariate activation is then represented as
\[
\varphi^{(l)}_{ij}(x)
=\sum_{k=0}^{K} c^{(l)}_{k,ij}\,P_k^{(\alpha,\beta)}\!\big(\tanh (x)\big),
\]
with trainable coefficients \(c^{(l)}_{k,ij}\) and often fixed \((\alpha,\beta)\).  
Kashefi~\cite{Kashefi25}\footnote{\url{https://github.com/Ali-Stanford/KAN_PointNet_CFD}} reported that low-degree expansions (\(K=2\)) with Chebyshev-like parameters \(\alpha=\beta=-0.5\) achieve an optimal balance between accuracy and numerical stability, suppressing Runge oscillations in inverse PDEs with sparse boundary data.  
Shukla~et~al.~\cite{KAN_pde_Shukla24} applied the same formulation within a PIKAN framework for high–Reynolds cavity flow, using \((\alpha,\beta)=(1,1)\) and degrees \(K=3\)–\(8\) depending on the regime.  
While higher degrees increase computational cost, their Jacobi–based PIKANs matched or exceeded conventional PINNs in accuracy.

\medskip
\noindent
Building upon this foundation, Xiong~et~al.~\cite{Xiong25} proposed the \emph{Jacobian Orthogonal Polynomial–based KAN (J–PIKAN)}, enforcing explicit orthogonality among polynomial components to ensure stable layerwise synthesis and interpretable coefficient updates.  
Leveraging the three-term recurrence of Jacobi polynomials, J–PIKAN efficiently evaluates higher-degree expansions and maintains numerical stability.  
Using fluid-dynamics benchmarks, their study compared B-spline, Fourier, Hermite, Legendre, Chebyshev, and Jacobi bases within a unified physics-informed framework, identifying Jacobi bases with moderate parameters (\(\alpha=\beta\approx2\)) as offering the best compromise between accuracy, conditioning, and convergence rate.

\medskip
\noindent
In parallel, Zhang~et~al.~\cite{Zhang26_legendre}\footnote{\url{https://github.com/zhang-zhuo001/Legend-KINN}} examined the Legendre specialization \((\alpha,\beta)=(0,0)\) within Jacobi-based KANs, demonstrating favorable conditioning and rapid convergence for physics-informed PDE solvers.  
A complementary direction is offered by the Adaptive PolyKAN~\cite{Attouri25}, which treats the polynomial degree as a trainable parameter, enabling the model to adjust its functional complexity dynamically over the course of training.

\medskip
\noindent
\textbf{Summary.}  
Jacobi-based KANs form a versatile and mathematically rich subclass within polynomial KANs.  
Classical (fixed-degree) variants emphasize numerical stability and geometric fidelity, while fractional (fKAN) and rational (rKAN) extensions introduce adaptive domain warping and enhanced expressivity.  
Together, they provide a cohesive Jacobi toolkit—bridging spectral approximation, rational modeling, and physics-informed learning within the broader KAN framework.

\subsection{Gaussian (RBF)}
\label{gaussiankan}

\noindent
As researchers sought to improve the efficiency and performance of the original B–spline–based KANs, the \emph{Gaussian radial basis function (RBF)} emerged as a powerful and computationally elegant alternative.  
Its smooth, localized, and infinitely differentiable nature makes it an excellent choice for representing univariate activation functions along the edges of a KAN.  

Across recent studies~\cite{Li24, Abueidda25, Koenig25, Athanasios2024, Buhler25_regression, RBFKAN, Liu2026, Chao2026, Chiu2026}, Gaussian–based activations have appeared in several formulations—ranging from fixed–grid hybrids to fully learnable and reflectional variants—each striking a different balance between stability, adaptivity, and efficiency.

\bigskip
\noindent
\textbf{Foundational Gaussian Layer.}  
The fundamental univariate Gaussian basis function is defined as
\begin{equation}
\varphi(x; c, \varepsilon) = \exp\!\left[-\Big(\tfrac{x-c}{\varepsilon}\Big)^2\right],
\label{eq:gaussian_basic}
\end{equation}
where \(c\) is the center and \(\varepsilon>0\) is the width (shape parameter).  
A general univariate activation is then constructed as a linear combination
\begin{equation}
\phi(x) = \sum_{i=0}^{G-1} w_i\,\varphi(x; g_i, \varepsilon),
\label{eq:gaussian_phi}
\end{equation}
with trainable coefficients \(w_i\) and equispaced grid centers \(g_i\in[a,b]\).  
This formulation replaces the piecewise polynomial behavior of splines with globally smooth Gaussian features while retaining locality and analytical gradients.

\bigskip
\noindent
\textbf{Hybrid Gaussian--Residual Formulation (Fixed Grid).}  

\begin{figure}[H]
\centering
\includegraphics[width=0.55\textwidth]{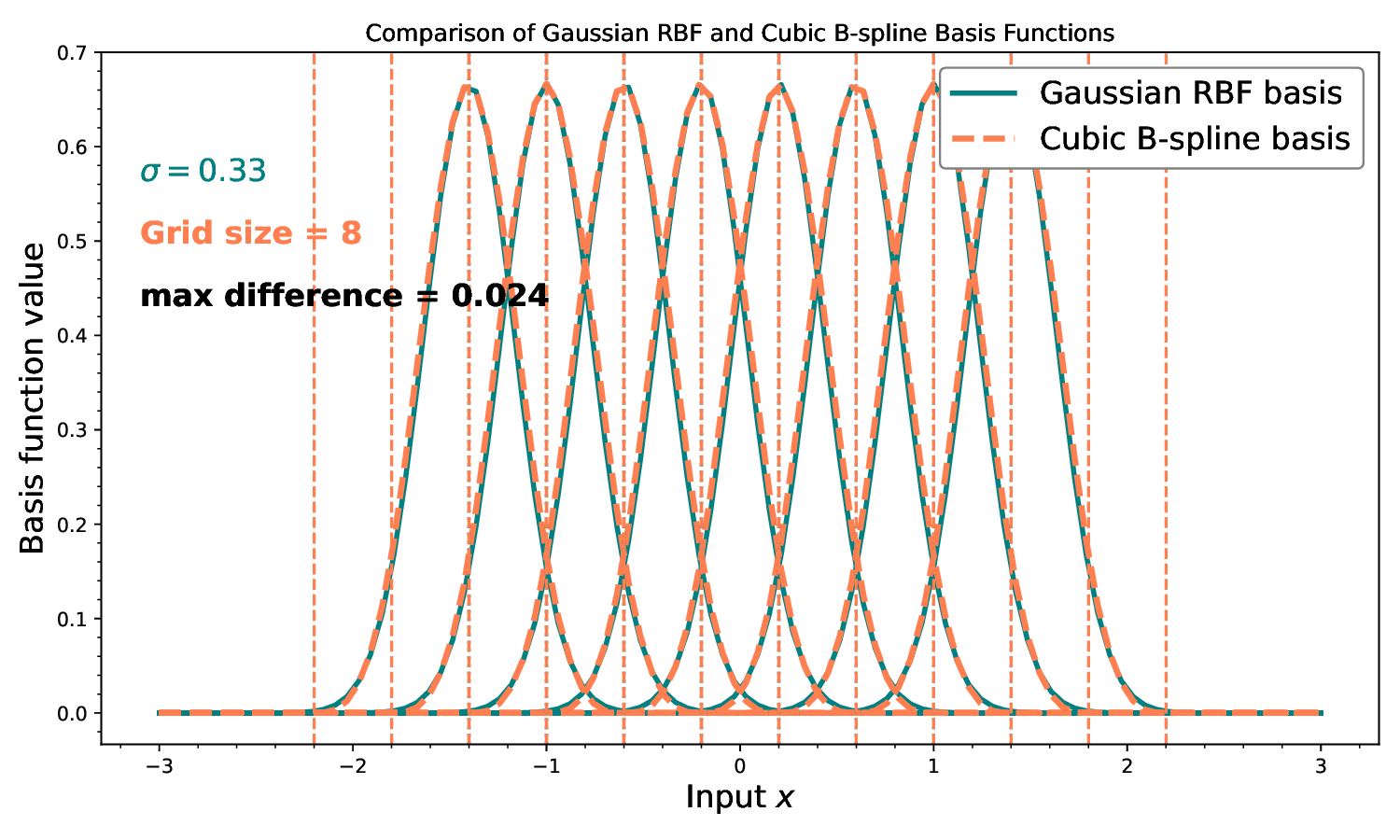}
\caption{Comparison of cubic B–spline basis functions \(\mathcal{B}_3(x)\) (non-extended) and Gaussian RBFs 
\(\exp[-((x-c_i)/\sigma)^2]\) with matched grid spacing and width.}
\label{compare_basis}
\end{figure}

\noindent
Gaussian RBFs are a fast, smooth alternative to B–splines for KAN activations.  
The key observation is that cubic B–splines can be closely approximated by scaled Gaussians.  
Li et~al.~\cite{Li24}\footnote{\url{https://github.com/ZiyaoLi/fast-kan}} proposed \emph{FastKAN}, replacing the cubic B–spline \(\mathcal{B}_3(x-c_i)\) with
\begin{equation}
\mathcal{B}_3(x - c_i) \;\approx\; \lambda\,\exp\!\left[-\Big(\tfrac{x-c_i}{\sigma}\Big)^2\right],
\end{equation}
where \(\lambda\in\mathbb{R}\) is a scaling constant and \(\sigma \approx h\) matches the grid spacing.  
As shown in Figure~\ref{compare_basis}, this preserves spline-like locality and smoothness while reducing computational cost.

\begin{figure}[H]
\centering
\includegraphics[width=0.55\textwidth]{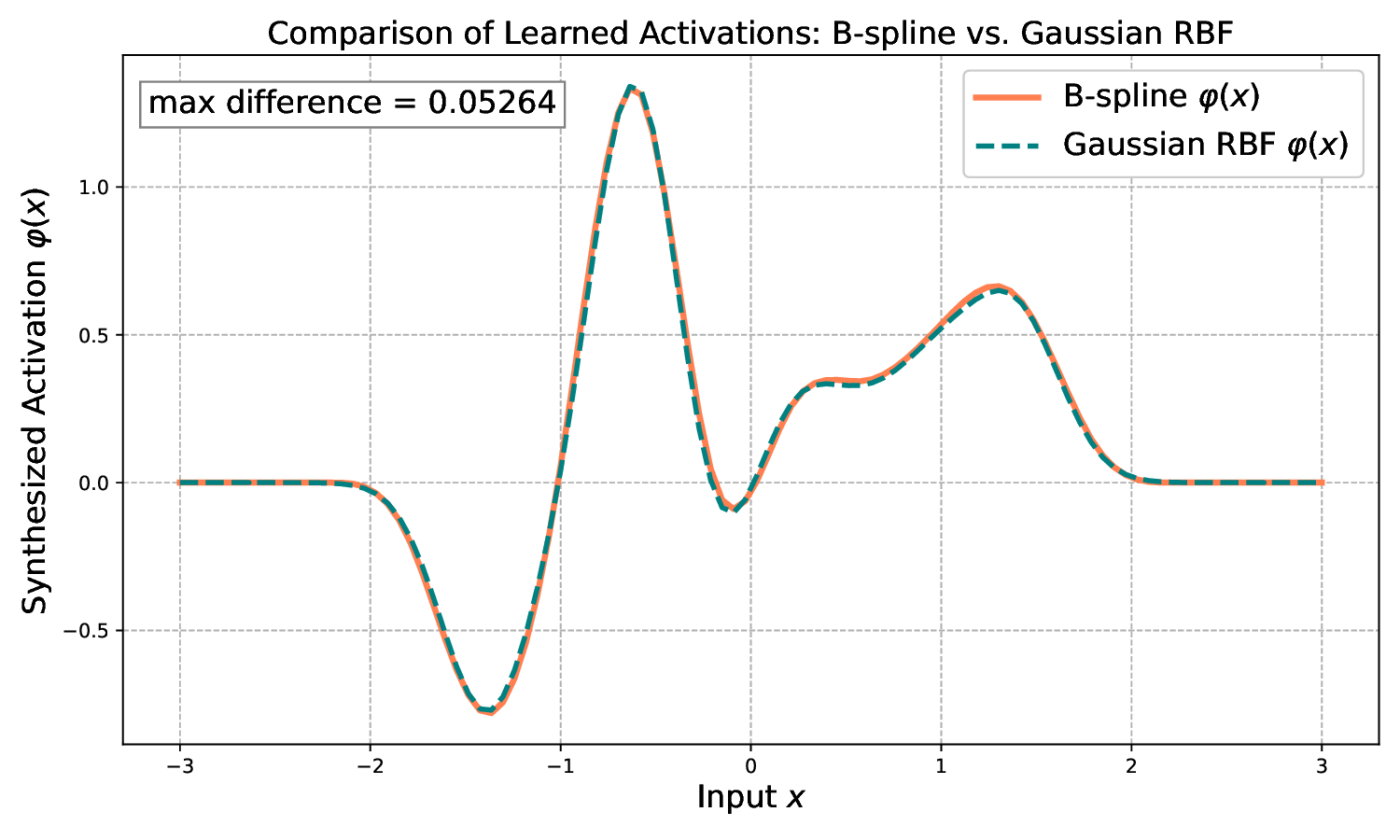}
\caption{Learned univariate activation \( \varphi(x) \) synthesized from Gaussian RBF and cubic B–spline bases (Figure~\ref{compare_basis}) using identical random coefficients.}
\label{GaussianSplineLearned}
\end{figure}

\noindent
\noindent
Figure~\ref{GaussianSplineLearned} demonstrates that both bases yield nearly identical activation profiles, with only minor differences near the peaks.  
Each basis decays smoothly near the boundaries, producing flat tails where the support vanishes. This confirms that Gaussian RBFs can serve as faithful and efficient surrogates for B-splines in KAN layers.

This behavior is achieved by tuning the scale parameter $\varepsilon$ in \eqref{eq:gaussian_basic}. 
The scale parameter plays a crucial role in both the accuracy and stability of Gaussian KANs. 
Since finding an ``optimal'' value of $\varepsilon$ is difficult and may depend on the data distribution, target function, domain, and network architecture, Noorizadegan and Wang~\cite{Amir_GKAN} proposed the practical range
\begin{equation}
\varepsilon \in \left[\frac{1}{G-1}, \frac{2}{G-1}\right],
\end{equation}
as a useful search interval for selecting a good value of $\varepsilon$ in function approximation and PDE-solving problems. 
Here, $G$ denotes the number of grid points. 
They also discussed several practical strategies for choosing a suitable scale parameter.

A related improvement was introduced in the partition-of-unity Gaussian formulation proposed in~\cite{Amir_PUGKAN}. 
In this approach, the standard Gaussian expansion in \eqref{eq:gaussian_phi} is replaced by a Shepard-type normalized activation,
\begin{equation}
\phi(x)
=
\frac{\sum_{i=0}^{G-1} w_i\,\varphi(x; g_i, \varepsilon)}
{\sum_{i=0}^{G-1} \varphi(x; g_i, \varepsilon)},
\end{equation}
which improves robustness and reduces sensitivity to the scale parameter $\varepsilon$. 
Thus, while the standard Gaussian KAN depends directly on the choice of scale, the normalized formulation provides a more accurate pure Gaussian alternative.

Beyond pure Gaussian KAN methods, Li, in the FastKAN GitHub repository,\footnote{\url{https://github.com/ZiyaoLi/fast-kan}} showed that Gaussian basis functions can also be combined with a residual activation, such as SiLU/Swish, to form a hybrid composite activation,
\begin{equation}
\Phi(x) 
= 
\underbrace{
\sum_{i=0}^{G-1} 
w_i\,
\exp\!\left[-\left(\frac{x-g_i}{\varepsilon}\right)^2\right]
}_{\text{RBF component}}
+
w_b\,\rho(x),
\label{eq:gaussian_hybrid}
\end{equation}
where $\varepsilon$ is the Gaussian scale parameter, $\rho(x)$ denotes the base activation, such as SiLU, and $w_b\in\mathbb{R}$ controls its contribution.
The Gaussian component captures local variations, while the residual activation stabilizes optimization and represents smoother global trends.
Here, both the grid $\{g_i\}$ and the scale parameter $\varepsilon$ are fixed, ensuring efficient and stable evaluation.  
This hybrid configuration, later adopted in LeanKAN~\cite{Koenig25} and RBFKAN~\cite{RBFKAN}, provides a strong and stable baseline for Gaussian-based KANs.

\noindent
\textbf{Pure RBF Layers with Learnable Parameters.}
A more flexible RBF formulation appears in the DeepOKAN architecture of Abueidda et al.~\cite{Abueidda25}\footnote{\url{https://github.com/DiabAbu/DeepOKAN}}, where Gaussian bases are used in \emph{both branch and trunk} networks.  
Each activation takes the form
\begin{equation}
\varphi(x) = \exp\!\left[-\Big(\tfrac{x-g_j}{\beta}\Big)^2\right],
\label{eq:abueidda_rbf}
\end{equation}
with trainable centers \(g_j\) and width \(\beta>0\).
The layer update is
\begin{equation}
x_{l+1} = W_l\,\varphi(x_l; G_l, \beta),
\end{equation}
allowing the RBFs to migrate toward regions of high variation.  
Abueidda et al.\ show that RBF--KANs resolve \emph{high-frequency} sinusoidal fields with markedly higher accuracy than MLPs in mechanics problems.  
A similar observation appears in the two-step DeepOKAN for bubble dynamics~\cite{Zhange25_bubble}, where RBFs remain effective in the branch network while spline bases outperform them in the trunk for extremely high-frequency oscillations.

\medskip
\noindent
Beyond operator-learning settings, the \emph{Free-RBF-KAN} model of Chiu et al.~\cite{Chiu2026} introduces adaptive radial basis functions whose centers and widths evolve during training, improving efficiency and stability in general function-learning tasks.

\bigskip
\noindent
\textbf{Reflectional Gaussian Approximations (RSWAF).}  
To further reduce computational cost, several works have replaced the exponential kernel with an algebraically similar \emph{Reflectional Switch Activation Function (RSWAF)} proposed by Delis~\cite{Athanasios2024}\footnote{\url{https://github.com/AthanasiosDelis/faster-kan}} and extended in Bühler et~al.~\cite{Buhler25_regression}.  
This activation approximates the Gaussian bell using the hyperbolic–secant squared function:
\begin{equation}
\varphi_{\text{RSWAF}}(x; c, h) = 1 - \tanh^2\!\Big(\tfrac{x-c}{h}\Big)
= \mathrm{sech}^2\!\Big(\tfrac{x-c}{h}\Big),
\label{eq:rswaf_basis}
\end{equation}
which preserves reflectional symmetry and smooth decay.  
Both the centers \(c_i\) and widths \(h_i\) may be fixed or trainable, controlled by configuration flags (e.g., \texttt{train\_grid}, \texttt{train\_inv\_denominator}) in implementations such as \emph{FasterKAN}~\cite{Athanasios2024}.  
The full activation is often expressed as
\begin{equation}
\phi(x) = \sum_i w_i\,s_i\,\bigg[1-\tanh^2\!\Big(\tfrac{x-c_i}{h_i}\Big)\bigg],
\label{eq:rswaf_phi}
\end{equation}
where \(w_i\) are learnable amplitudes, \(s_i\) are reference scaling factors, and both \(c_i\) and \(h_i\) evolve during training.  
This formulation yields a localized, computationally efficient surrogate that mimics Gaussian curvature while offering explicit control over both shape and location.

\bigskip
\noindent
\textbf{Hybrid and Unified Gaussian Extensions.}  
Koenig et~al.~\cite{Koenig25}\footnote{\url{https://github.com/DENG-MIT/LeanKAN}} proposed a hybrid activation that mixes RBFs with a base nonlinearity:
\begin{equation}
  \phi(x) \;=\; \sum_{i=1}^{N} w_i\,\varphi(x-c_i) \;+\; w_b\, b(x),
\end{equation}
where $\varphi(\cdot)$ is a Gaussian RBF, $b(x)$ is a base activation (e.g., Swish or linear), and $w_b$ modulates its contribution.  
Related implementations such as \emph{RBFKAN}~\cite{RBFKAN}\footnote{\url{https://github.com/sidhu2690/RBF-KAN}} also apply simple min–max normalization,
\begin{equation}
  x_{\text{norm}} \;=\; \frac{x - x_{\min}}{x_{\max} - x_{\min}},
\end{equation}
to stabilize input scaling.  

\noindent
The \emph{BSRBF–KAN} model~\cite{Ta24}\footnote{\url{https://github.com/hoangthangta/BSRBF_KAN}} unifies B–splines and Gaussian RBFs within a single layer:
\begin{equation}
\mathrm{BSRBF}(x) \;=\; w_b\, b(x) \;+\; w_s\big(\mathrm{BS}(x)+\mathrm{RBF}(x)\big),
\end{equation}
where \(\mathrm{BS}(x)\) and \(\mathrm{RBF}(x)\) are evaluated at identical grid locations, and \(w_b,w_s\in\mathbb{R}\).  
This configuration inherits the local compactness of B–splines and the global smoothness of Gaussian functions.

\bigskip
\noindent
\textbf{Complex–Valued Gaussian Extensions.}  
In the complex domain, Wolff et~al.~\cite{Wolff25}\footnote{\url{https://github.com/M-Wolff/CVKAN}} introduced the \emph{Complex–Valued KAN (CVKAN)}, extending FastKAN with concepts from complex–valued neural networks (CVNNs)~\cite{Lee22_complex}.  
The residual pathway uses a complex SiLU activation:
\begin{equation}
b_{\mathbb{C}}(z) = \mathrm{CSiLU}(z) = \mathrm{SiLU}(\mathrm{Re}(z)) + i\,\mathrm{SiLU}(\mathrm{Im}(z)),
\end{equation}
and the Gaussian RBF is generalized to complex inputs~\cite{Chen94}:
\begin{equation}
\mathrm{RBFC}(z) = \sum_{i=1}^N w_i\,\varphi(|z-c_i|), \qquad c_i\in\mathbb{C}, \; w_i\in\mathbb{C}.
\end{equation}
The resulting activation combines both terms with a bias term:
\begin{equation}
\phi(z) = w_s\,\mathrm{RBFC}(z) + w_b\,b_{\mathbb{C}}(z) + \beta, \qquad \beta\in\mathbb{C}.
\end{equation}
More recently, Che et~al.~\cite{Che25_complexKAN} extended this formulation by introducing a \emph{ModELU–based CVKAN}, replacing the split–type CSiLU with a modulus–preserving residual and learnable RBF shape parameters, supported by a formal complex–valued Kolmogorov–Arnold theorem.

\noindent
\textbf{Practical caveat: the shape parameter.}
Gaussian RBF performance is highly sensitive to the width parameter (\(\varepsilon\)), which governs both smoothness and conditioning~\cite{Amir22, Amir23, Larsson24, Tizian1, Tizian2, Tizian3, Ling06, Ling20}. 
Even small changes can strongly impact accuracy and numerical stability~\cite{Amir_s24, Cavoretto21AD, Fasshauer07}. 
From a kernel–theoretic perspective, this sensitivity reflects an \emph{uncertainty principle}: larger shape (width) parameters produce flatter, globally correlated kernels that improve smoothness and accuracy but worsen conditioning, whereas smaller parameters yield well–conditioned, highly localized bases at the expense of approximation power~\cite{Schaback95, Schaback23}. 
While heuristic tuning and learnable widths (as in FastKAN and DeepONet–RBF variants) offer partial remedies, robust and theoretically guided selection of this parameter remains an active research problem at the intersection of kernel methods and neural approximation.

\bigskip
\noindent
\textbf{Summary.} Gaussian–based KANs form a versatile class of architectures bridging spline–based and kernel–theoretic perspectives.  
Starting from fixed equispaced Gaussian grids~\cite{Li24, Koenig25}, through adaptive RBF layers with learnable centers~\cite{Abueidda25}, to reflectional Gaussian and complex–valued extensions~\cite{Athanasios2024, Buhler25_regression, Wolff25, Che25_complexKAN}, these models collectively demonstrate how Gaussian basis functions can serve as both efficient computational surrogates and expressive functional primitives within the KAN framework.

\subsection{Fourier}\label{fourierkan}

Fourier features provide global, smooth, periodic expressivity that complements localized bases (e.g., B--splines, RBFs). 
Within the KAN family, Fourier–based KANs replace the 
univariate edge functions with harmonic expansions, retaining the KAN wiring while improving spectral resolution on tasks 
with strong periodic structure \cite{Xu25_fourier,Yuan2026,Shamim2026}. 
By contrast, Fourier networks such as FAN are \emph{not} KANs: they embed cosine–sine transforms directly into each layer 
as a drop-in MLP replacement, targeting efficient
 periodicity modeling (often with reduced parameters/FLOPs)
  and stronger in-/out-of-domain generalization \cite{Dong25_FAN}\footnote{\url{https://github.com/YihongDong/FAN}}. 
  We mention FAN here as a related spectral approach; 
  our focus remains on Fourier features used \emph{inside} KANs.

\noindent
\textbf{Standard FourierKAN.}
In the baseline FourierKAN architecture~\cite{FourierKAN}\footnote{\url{https://github.com/GistNoesis/FourierKAN}}, each univariate mapping is expressed as a truncated Fourier series,
\begin{equation}
\varphi(x) = \sum_{k=1}^{K} \big( a_k \cos(kx) + b_k \sin(kx) \big),
\label{eq:fourier_phi}
\end{equation}
with learnable coefficients $a_k,b_k\in\mathbb{R}$ and cutoff frequency $K\in\mathbb{N}$. Cosines capture even–symmetric modes; sines capture odd–symmetric modes. The resulting $\varphi(x)$ is a flexible global harmonic mixture, well–suited to periodic or high–frequency structure.
Guo~et~al.~\cite{Guo25_Fourier} adopt this classical truncated Fourier formulation, using fixed integer harmonics with trainable sine–cosine coefficients. 

\noindent
\textbf{Random Fourier features.}
A limitation of \eqref{eq:fourier_phi} is the fixed 
set of frequencies $k=1,\ldots,K$. Zhang et al.~\cite{Zhang25}\footnote{\url{https://github.com/kolmogorovArnoldFourierNetwork/KAF}} propose the \emph{Kolmogorov–Arnold–Fourier Network} 
(KAF), replacing fixed bases with a learnable spectral embedding via Random Fourier Features (RFF):
\begin{equation}
\psi_{\mathrm{RFF}}(\mathbf{x})
=
\sqrt{\tfrac{2}{m}}\,
\big[\,
\cos(\mathbf{x}W+\mathbf{b}),\;
\sin(\mathbf{x}W+\mathbf{b})
\,\big]
\in\mathbb{R}^{2m},
\label{eq:rff_embedding}
\end{equation}
where $W\in\mathbb{R}^{d_{\mathrm{in}}\times m}$ is a trainable frequency matrix and $\mathbf{b}\in\mathbb{R}^{m}$ is a learnable phase vector. KAF then blends this adaptive spectral block with a smooth baseline:
\begin{equation}
\varphi(\mathbf{x})
=
\boldsymbol{\alpha}\cdot \mathrm{GELU}(\mathbf{x})
\;+\;
\boldsymbol{\beta}\cdot V\,\psi_{\mathrm{RFF}}(\mathbf{x}),
\label{eq:kaf_hybrid}
\end{equation}
with $\boldsymbol{\alpha},\boldsymbol{\beta}\in\mathbb{R}^{d_{\mathrm{in}}}$ and $V\in\mathbb{R}^{d_{\mathrm{in}}\times 2m}$. The GELU term captures low–frequency structure, while the RFF term adaptively models higher–frequency content throughout training. This continuous frequency control improves regression/classification where global and local patterns coexist~\cite{Zhang25}.

\begin{figure}[H]
\centering
\includegraphics[width=0.75\textwidth]{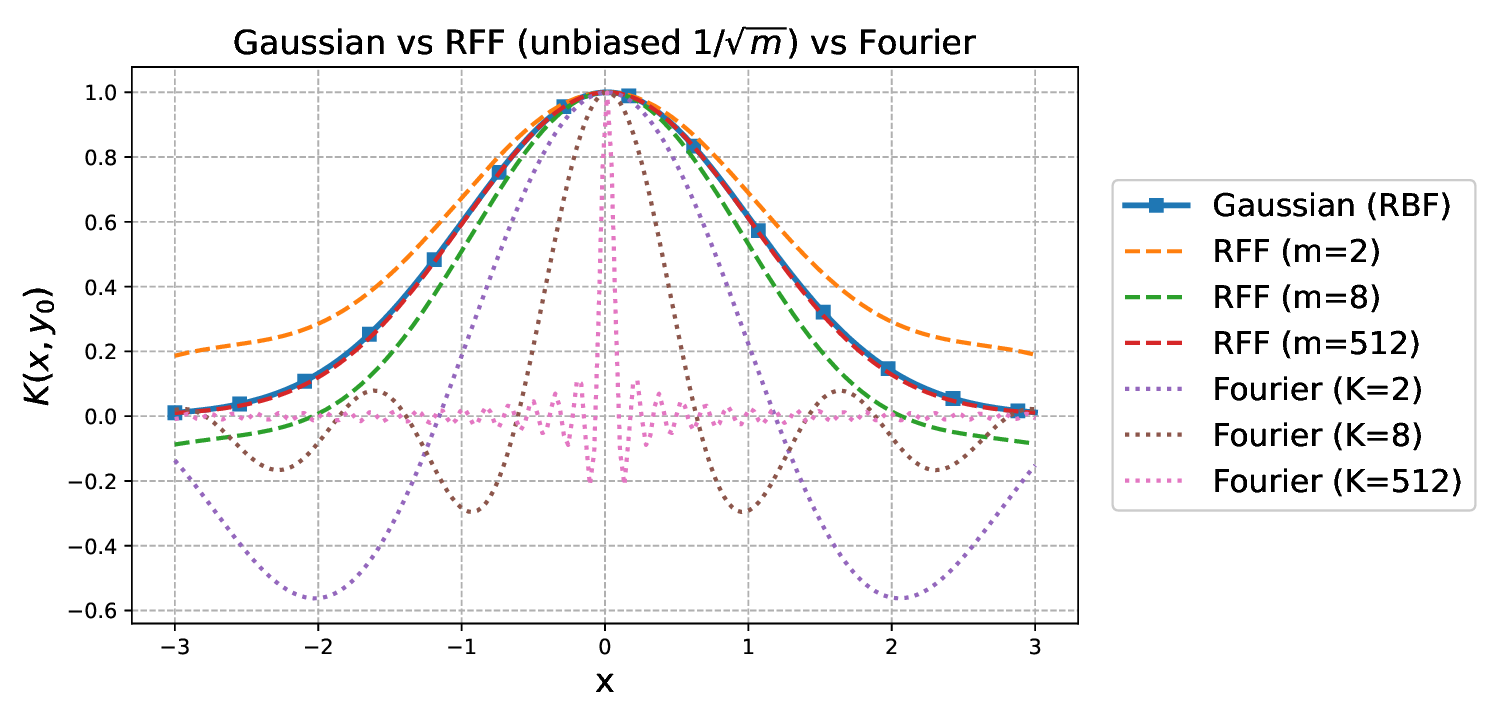}
\caption{Gaussian (RBF) kernel slice compared with Random Fourier Features (RFF, 
~\eqref{eq:rff_embedding}) and a truncated Fourier expansion \eqref{eq:fourier_phi}.  
RFF uses the unbiased scaling $1/\!\sqrt{m}$ with both $\cos$ and $\sin$ terms, 
so that the diagonal kernel value is close to~1.  
As $m$ increases, the RFF slice aligns with the Gaussian, while the Fourier 
expansion (fixed integer harmonics) retains its periodic structure.}
\label{FourierVsGauss}
\end{figure}

\noindent
Figure~\ref{FourierVsGauss} compares a Gaussian (RBF) 
kernel slice with both random Fourier features  \eqref{eq:rff_embedding} and a truncated Fourier expansion
 using the full cosine–sine basis. Settings match the code: one-dimensional grid $x\in[-3\varepsilon,3\varepsilon]$ with
  $\varepsilon=1.0$. The RFF map uses the unbiased scaling $1/\!\sqrt{m}$. With this setting, increasing $m$ makes the
   RFF curve visually coincide with the Gaussian slice—clearly showing how the two are connected in practice. The Fourier
    expansion \eqref{eq:fourier_phi}, constructed from fixed integer harmonics $\{\cos(k\gamma x),\sin(k\gamma x)\}_{k=1}^K$,
     remains strictly periodic, so its kernel slice differs structurally from the Gaussian, in contrast to RFF where 
     frequencies are sampled from a Gaussian distribution.

\noindent
In addition to classical Fourier-based designs, Jiang et al.~\cite{Jiang25_quantum}\footnote{\url{https://github.com/Jim137/qkan}} propose the quantum-inspired Kolmogorov--Arnold Network (QKAN), where each univariate map is realized by a data re-uploading variational quantum circuit. These circuits naturally generate Fourier-like expansions with exponentially scalable frequency modes, offering parameter-efficient alternatives to grid-based Fourier KANs while retaining the structured interpretability of classical KANs.

\noindent
\textbf{Summary.}
Fourier–based KANs—via deterministic truncations or 
randomized spectral embeddings—inject global harmonic
 structure into KAN layers. Compared with local polynomial bases such as B–splines, these global bases are particularly 
 effective for smooth periodicities or problems with
  broad spectral support.
  
\subsection{Wavelet}
\label{Wavelet}

\noindent
Wavelet-based KANs (Wav-KANs) extend the KAN framework 
by using \emph{localized} wavelet bases, enabling multiscale representation and 
spatial adaptivity—especially useful for heterogeneous or hierarchical data. The architecture was introduced by
 Bozorgasl et al.~\cite{Bozorgasl24}\footnote{\url{https://github.com/zavareh1/Wav-KAN}} and draws on both the 
 Continuous Wavelet Transform (CWT) and Discrete Wavelet 
 Transform (DWT).

\noindent
In wavelet analysis, a signal $g(t)\in L^2(\mathbb{R})$ is decomposed via scaled/translated copies of a mother wavelet $\psi(t)$. The CWT is defined~\cite{Grossmann84,Calderon64} as
\begin{equation}
C(s,\tau) \;=\; \int_{-\infty}^{\infty} g(t)\,\frac{1}{\sqrt{s}}\,\psi\!\left(\frac{t-\tau}{s}\right)\,dt,
\end{equation}
where $s>0$ is the scale and $\tau\in\mathbb{R}$ is the translation. The discrete analog used in signal processing is the DWT:
\begin{equation}
a_j(k) \;=\; \sum_{n} g(n)\,\phi_{j,k}(n), 
\qquad
d_j(k) \;=\; \sum_{n} g(n)\,\psi_{j,k}(n),
\end{equation}
with $\phi_{j,k}$ and $\psi_{j,k}$ the scaling and wavelet functions at resolution level $j$ and position $k$.

\noindent
\textbf{Wavelet activations in KAN layers.}
Wav-KAN embeds analytic wavelet functions directly into univariate edge functions. Each activation is parameterized by scale $s$ and translation $u$ \cite{Bozorgasl24,LiC2026,Li2026}:
\[
\varphi(x) \;=\; \psi\!\left(\frac{x-u}{s}\right),
\]
where $\psi(t)$ is chosen from canonical families:
\begin{align}
\psi_{\text{mex}}(t) &= \tfrac{2}{\sqrt{3}\,\pi^{1/4}}(1-t^2)\,e^{-t^2/2}, \\[4pt]
\psi_{\text{mor}}(t) &= \cos(\omega_0 t)\,e^{-t^2/2}, \quad \omega_0=5, \\[4pt]
\psi_{\text{dog}}(t) &= -t\,e^{-t^2/2}.
\end{align}

\begin{figure}[hbt!]
\centering
\includegraphics[width=0.55\textwidth]{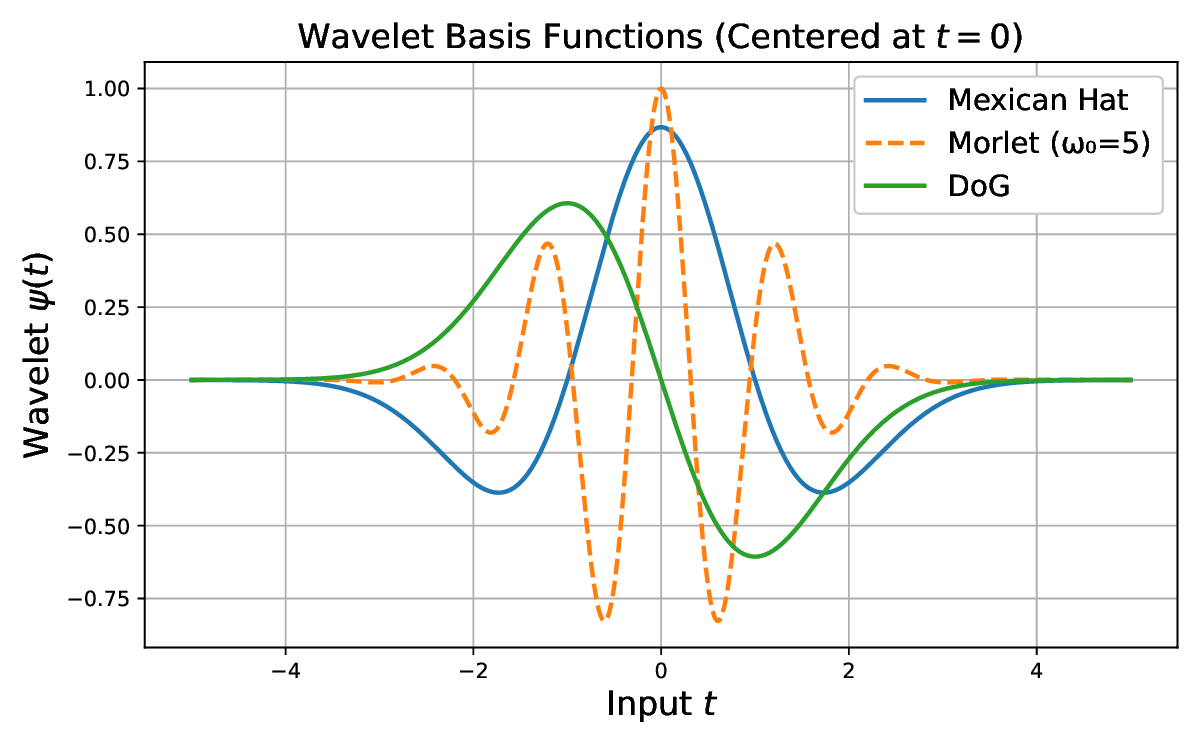}
\caption{Three wavelet bases centered at $t=0$: Mexican Hat, Morlet (with $\omega_0=5$), and Derivative of Gaussian (DoG). Each has localized support and distinct oscillatory behavior.}
\label{WaveletBasis}
\end{figure}

\begin{figure}[hbt!]
\centering
\includegraphics[width=0.55\textwidth]{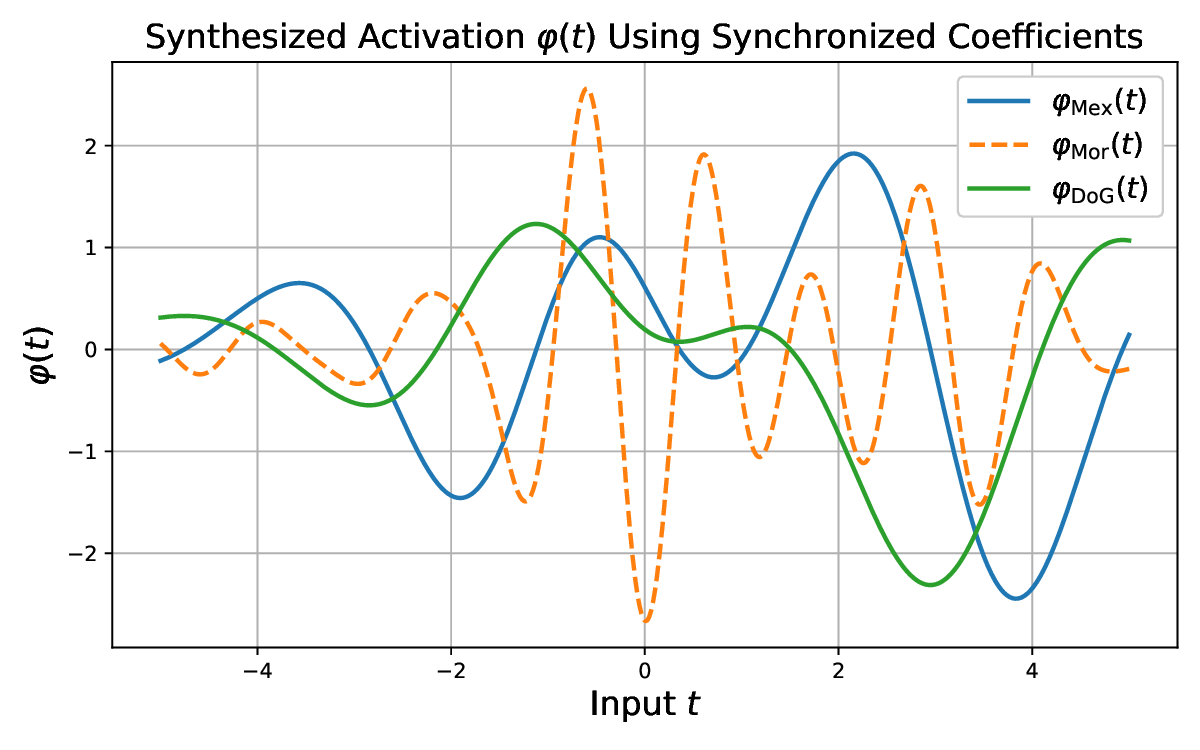}
\caption{Synthesized activations $\varphi(t)=\sum_{j=1}^M c_j \psi(t-\mu_j)$ using synchronized coefficients and centers across wavelet families. Differences in frequency and amplitude follow from each wavelet’s intrinsic shape.}
\label{waveletLearned}
\end{figure}

\noindent
To compare expressivity across wavelet families, we form synchronized linear combinations
\[
\varphi(t) \;=\; \sum_{j=1}^{M} c_j\,\psi(t-\mu_j),
\]
using the same coefficients $\{c_j\}$ and centers
 $\{\mu_j\}$ for each $\psi$. Figure~\ref{WaveletBasis} shows three canonical wavelets (Mexican Hat, Morlet, DoG) 
 centered at $t=0$. Using these bases, Figure~\ref{waveletLearned} presents the synthesized activations $\varphi(t)$ 
 built with identical $\{c_j\}$ and $\{\mu_j\}$. The Morlet activation exhibits higher-frequency oscillations,
  whereas Mexican Hat and DoG yield smoother bell-shaped 
profiles. Since coefficients are shared, differences arise solely from the intrinsic wavelet shapes.

\noindent
Learning $s$ and $u$ per neuron allows Wav-KAN to adapt receptive fields and
frequency resolution, aiding generalization in settings with spatially varying
feature scales, such as image classification~\cite{Bozorgasl24}.

\noindent
Patra et al.~\cite{pde_Patra24} apply wavelet activations—using a sine--Gaussian
basis—to physics-informed training, tuning frequency response to obtain faster
data-free convergence on coupled nonlinear PDEs.  
Meshir et al.~\cite{Meshir25_NTK} complement this with an NTK analysis,
showing that the Morlet mother-wavelet frequency (or hidden-layer width)
directly controls NTK eigenvalue decay, providing explicit spectral-bias
mitigation and strong gains in physics-informed settings
(Poisson/Heat/Wave/Helmholtz).

\noindent
Applications further highlight the versatility of wavelet bases: Pratyush
et al.~\cite{Pratyush24} employ a DoG wavelet in CaLMPhosKAN for enhanced
phosphorylation-site detection, while Seydi et al.~\cite{Seydi24b} combine CWT
and DWT for hyperspectral imaging, outperforming spline-KANs and MLPs. 
Wavelet–KAN blocks have also been integrated with attention mechanisms for
dense prediction tasks such as medical image segmentation~\cite{Liang2026}.

\medskip
\noindent
\textbf{Summary.}
Wavelet-based KANs provide a flexible multiscale basis whose localized
frequency control improves convergence and robustness across PDEs,
bioinformatics, and high-dimensional imaging.

\subsection{Finite-Basis (FBKANs)}
\label{Finite}

\noindent
Finite-Basis KANs, introduced by Howard et al.~\cite{pde_fbkan_Howard24}\footnote{\url{https://github.com/pnnl/neuromancer/tree/feature/fbkans/examples/KANs}},\footnote{\url{https://github.com/pnnl/neuromancer/blob/feature/fbkans/src/neuromancer/modules}},
extend KANs with a domain-decomposition strategy 
inspired by finite-basis PINNs (FBPINNs)~\cite{Moseley23, Dolean24, Dolean24b, Heinlein24}. The key idea is to represent 
a global mapping as a sum of \emph{local} subnetworks, each modulated by a smooth \emph{partition-of-unity} (PoU) function. 
This yields scalability, robustness, and improved 
generalization for multiscale and physics-informed problems. 

\medskip
\noindent
Let $\Omega \subset \mathbb{R}$ be covered by $L$ overlapping subdomains $\{\Omega_j\}_{j=1}^{L}$. Assign a smooth PoU weight $\omega_j(x)$ to each $\Omega_j$ such that~\cite{Roberto1,Roberto2,Roberto3}
\[
\mathrm{supp}\big(\omega_j\big)=\Omega_j, 
\qquad 
\sum_{j=1}^{L}\omega_j(x)\equiv 1 \ \ \forall x\in\Omega.
\]
Cosine windows are used to build unnormalized weights $\hat{\omega}_j$, which are then normalized:
\begin{equation}
\omega_j(x) \;=\; \frac{\hat{\omega}_j(x)}{\sum_{k=1}^{L}\hat{\omega}_k(x)},
\label{eq:fbkan_omega}
\end{equation}
with
\begin{equation}
\hat{\omega}_j(x) \;=\;
\begin{cases}
1, & L=1,\\[4pt]
\Big[1+\cos\!\big(\pi \tfrac{x-\mu_j}{\sigma_j}\big)\Big]^2, & L>1,
\end{cases}
\end{equation}
and
\[
\mu_j \;=\; \frac{l(j-1)}{L-1}, 
\qquad 
\sigma_j \;=\; \frac{\delta\,l}{2(L-1)}, 
\qquad 
l \;=\; \max(x)-\min(x),
\]
where $\delta>1$ is the overlap ratio controlling the window width.

\medskip
\noindent
Each subdomain $\Omega_j$ hosts a local KAN, $K_j(x;\boldsymbol{\theta}_j)$, trained only on its region. The global predictor is a smooth PoU-weighted sum of local outputs:
\begin{equation}
f(x) \;\approx\; \sum_{j=1}^{L} \omega_j(x)\,K_j(x;\boldsymbol{\theta}_j),
\label{eq:fbkan}
\end{equation}
where $\boldsymbol{\theta}_j$ are the local parameters. Within each local KAN, a univariate edge function $\varphi_j$ is expanded in cubic B-splines:
\[
\varphi_j(x) \;=\; \sum_{n=0}^{N_j-1} c_{j,n}\,B_n^{(3)}(x),
\]
with trainable coefficients $c_{j,n}\in\mathbb{R}$. The local spline grid is initialized from the effective support of $\omega_j$:
\[
a_j \;=\; \min\{x \mid \omega_j(x)>\varepsilon\}, 
\quad 
b_j \;=\; \max\{x \mid \omega_j(x)>\varepsilon\}, 
\quad 
\varepsilon \;=\; 10^{-4}.
\]

\begin{figure}[hbt!]
\centering
\includegraphics[width=0.85\textwidth]{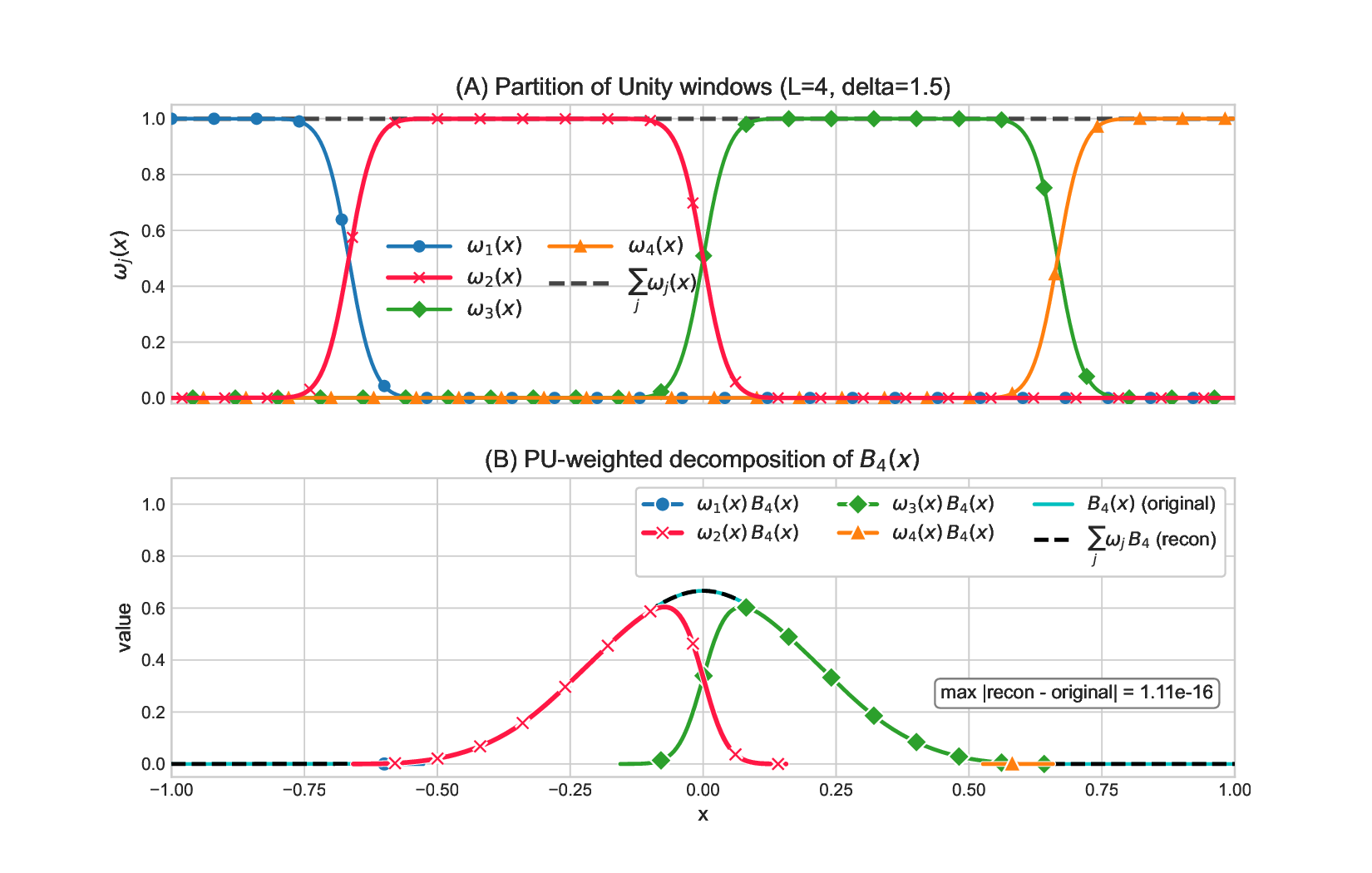}
\caption{Two-panel illustration of PoU–weighted 
B-spline decomposition on $[-1,1]$ ($L=4$, overlap $\delta=1.5$). 
(A) Cosine-based PoU windows $\{\omega_j\}$ summing to 
one. (B) Decomposition of a cubic B-spline basis
 $B_i^{(3)}(x)$ into localized pieces 
 $\omega_j(x)B_i^{(3)}(x)$; their sum (black) 
 matches the original basis, confirming 
 $B_i^{(3)}(x)=\sum_{j=1}^{L}\omega_j(x)\,B_i^{(3)}(x)$.}
\label{SplineVsPU}
\end{figure}

\medskip
\noindent
To illustrate functional flexibility, 
Figure~\ref{SplineVsPU} shows, on $[-1,1]$, 
(A) the PoU weights $\{\omega_j\}_{j=1}^{4}$ with $\sum_j \omega_j \equiv 1$, and (B) the PoU–weighted decomposition 
of a single cubic B-spline basis $B_i^{(3)}(x)$. The colored curves are the localized pieces $\omega_j(x)\,B_i^{(3)}(x)$; 
their sum coincides with the original basis up to 
numerical precision. Only windows overlapping the 
spline support contribute nonzero pieces, so the PoU
 localizes the representation without changing the 
 underlying spline span. 

\medskip
\noindent
\textbf{Summary.}
FBKANs are practical because they let you model
 different parts of the domain with specialized 
 sub-KANs, which can be trained independently (and in parallel) to cut memory use and wall time. They work well with
  physics-informed losses, handle noisy data robustly, 
  and maintain smooth transitions between subdomains without requiring hard boundary constraints.  Moreover, because sub-KANs operate on smaller supports, each local kernel matrix is better conditioned—an important advantage for large-scale PDEs.

\subsection{SincKAN}
\label{sinckan}

\noindent
To model singularities, sharp gradients, and 
high-frequency transitions, Yu et al.~\cite{Yu24}\footnote{\url{https://github.com/DUCH714/SincKAN}}
 introduced the \emph{Sinc-based KAN} (SincKAN). 
 The building block is the globally supported Sinc kernel
\[
\mathrm{Sinc}(x)=\frac{\sin x}{x},\qquad \mathrm{Sinc}(0):=1,
\]
a classical tool for bandlimited approximation.

\medskip
\noindent
A basic SincKAN activation expands a univariate 
function using uniformly shifted Sinc atoms:
\begin{equation}
\varphi(x)=\sum_{i=-N}^{N} c_i\,
\mathrm{Sinc}\!\left(\frac{\pi}{h}\,(x-i h)\right),
\label{eq:sinckan_single}
\end{equation}
where $h>0$ is the step size, $\{c_i\}$ are trainable c
oefficients, and $N$ controls the truncation degree 
(total number of atoms $2N+1$).

\medskip
\noindent
\textbf{Practical multi--step-size form.}
To avoid choosing a single ``optimal'' $h$ and to 
improve flexibility on finite domains, SincKAN 
mixes several step sizes and applies a normalized coordinate transform $\gamma^{-1}$:
\begin{equation}
\varphi_{\mathrm{multi}}(x)
=\sum_{j=1}^{M}\sum_{i=-N}^{N} c_{i,j}\,
\mathrm{Sinc}\!\left(\frac{\pi}{h_j}\big(\gamma^{-1}(x)-i h_j\big)\right).
\label{eq:sinckan_multi}
\end{equation}
Here $\{h_j\}_{j=1}^{M}$ are (learned or preset) step sizes, $M$ is the number of scales, $c_{i,j}$ are trainable 
coefficients, and $\gamma^{-1}$ maps the working interval to $\mathbb{R}$ while normalizing scale. Setting 
$M{=}1$ and $\gamma^{-1}(x){=}x$ recovers 
\eqref{eq:sinckan_single}.

\begin{figure}[htbp]
    \centering
    \includegraphics[width=0.7\textwidth]{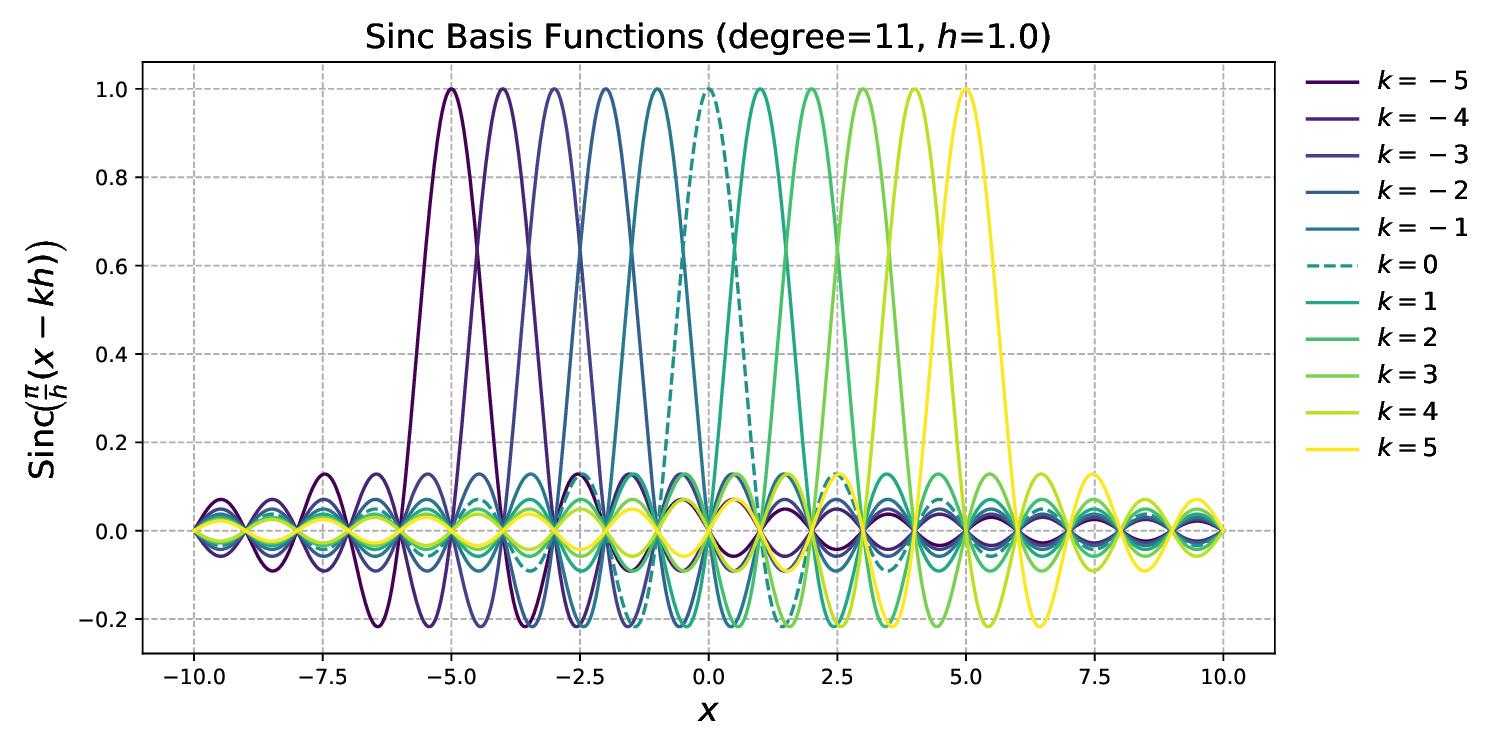}
    \caption{Truncated Sinc basis functions used in SincKAN. Each kernel is centered at $x=k h$ with spacing $h=1.0$; the total number of shifts is $2N+1=11$. This dictionary underlies the learned univariate map $\varphi(x)$.}
    \label{SincBasis}
\end{figure}

\begin{figure}[hbt!]
  \centering
  \includegraphics[width=0.78\textwidth]{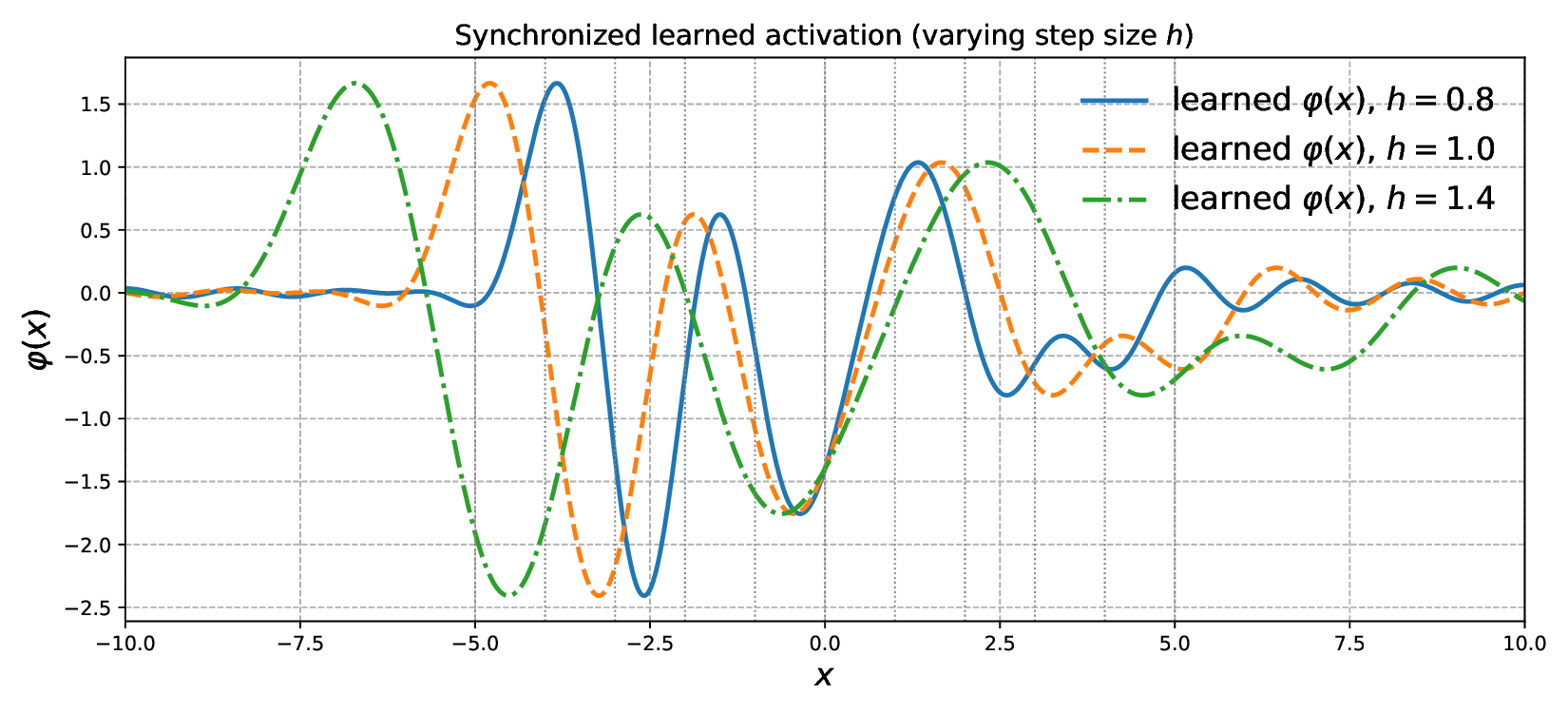}
  \caption{Synchronized learned activations with a pure (non-windowed) Sinc basis (degree $2N+1=11$). The same coefficients $\{c_i\}$ are used for all curves; only the step size $h$ changes ($h\in\{0.8,1.0,1.4\}$). Smaller $h$ yields higher-frequency, more oscillatory structure, whereas larger $h$ produces smoother responses. Vertical dotted lines mark the Sinc centers for $h=1.0$.}
  \label{sinc_learned}
\end{figure}

\medskip
\noindent
Figures~\ref{SincBasis} and \ref{sinc_learned} show the building blocks and a synthesized activation using a fixed truncation degree $(2N+1=11)$ and spacing $h=1.0$ over $x\in[-10,10]$. Figure~\ref{SincBasis} displays the truncated Sinc dictionary
$\{\mathrm{Sinc}(\tfrac{\pi}{h}(x-k h))\}_{k=-N}^{N}$, whose elements are globally supported and oscillatory with decaying sidelobes. These atoms are linearly combined to form the learned univariate map $\varphi(x)$.

\medskip
\noindent
Figure~\ref{sinc_learned} plots \emph{synchronized} learned activations
\[
\varphi(x)=\sum_{i=-5}^{5} c_i\,\mathrm{Sinc}\!\big(\tfrac{\pi}{h}(x-i h)\big)
\]
with the same random coefficients $\{c_i\}$, while varying only the step size $h\in\{0.8,1.0,1.4\}$. The step size $h$ sets the spacing of Sinc centers and therefore the effective bandwidth: smaller $h$ $\Rightarrow$ denser centers, higher-frequency capacity, more oscillations; larger $h$ $\Rightarrow$ smoother, lower-frequency behavior. Because the coefficients are synchronized, differences across curves reflect only the choice of $h$.

\medskip
\noindent
For numerical stability, SincKAN applies a coordinate normalization step that maps inputs to a bounded interval (typically $[-1,1]$) before evaluating the Sinc expansion. This preprocessing, consistent with other KAN variants, improves conditioning and convergence across layers.

\section{Accuracy Improvement}\label{sec:accuracy-improvement}

\noindent We group works by \emph{how} they improve
 accuracy, independent of task (regression, PINN, DeepONet),
  basis choice, or efficiency tricks. In most cases the 
same mechanism transfers across tasks.

\noindent
To orient the reader, we begin with a compact summary of accuracy–improving mechanisms for KANs. The table groups methods by mechanism and lists concise technique keywords with representative references. The remainder of this section expands on each group in turn, detailing when/why they help and practical choices and caveats.

\subsection{Physics Constraints \& Loss Design}\label{physics}
\noindent
Physics-informed learning, introduced by Raissi et al.~\cite{Raissi19}, improves accuracy by embedding governing physical laws—PDE residuals and boundary/initial conditions—directly into the training objective. 

A standard PINN-style loss combines three components:
\begin{equation}\label{PINN_eq}
\mathcal{L} \;=\;
w_u \underbrace{\mathcal{L}_{\mathrm{data}}}_{\substack{\text{fit to observations} \\ \text{improves data accuracy}}}
+ w_f \underbrace{\mathcal{L}_{\mathrm{phys}}}_{\substack{\text{PDE residuals} \\ \text{enforce physical laws}}}
+ w_b \underbrace{\mathcal{L}_{\mathrm{bc}}}_{\substack{\text{boundary/initial} \\ \text{conditions enforcement}}},
\end{equation}
where $\mathcal{L}_{\mathrm{data}}$ is the data-fidelity loss, $\mathcal{L}_{\mathrm{phys}}$ enforces PDE residuals, and $\mathcal{L}_{\mathrm{bc}}$ enforces boundary/initial conditions.  
The scalar weights $w_u, w_f, w_b > 0$ determine the balance between data fidelity and physics consistency.

\medskip
\noindent
Building on this foundation, a wide range of studies have extended the basic PINN loss through specialized 
constraints and tailored formulations in cooperation with KAN architectures~\cite{Liu24,Yu24,Zhang25_comp,Yang25,Guo25,Xu25,
Toscano24_kkan,Mostajeran24,Mostajeran25,Guo24,
pde_Ranasinghe24,pde_shuai24,KAN_pde_So24,
pde_bayesian_Giroux24,Khedr25,Lei25,Kalesh25,Gao25, Chen25_weak}.  
Among these, several works move beyond the conventional physics-based loss in~\eqref{PINN_eq} by embedding 
more advanced physics constraints and loss designs directly into the KAN architecture:

\begin{itemize}

  \item \textbf{Structural \& Conservation Laws.}  
  Enforce hard invariants such as Hamiltonian, quantum, algebraic, or geometric symmetries.  
  Includes \emph{SPEL-KAN}~\cite{Wang25_Hamiltonian}, \emph{KAN-ETS}~\cite{Sen25_time}, \emph{DAE-KAN}~\cite{Luo25_DAE},  
  and the symmetry-preserving \emph{GKSN} architecture of Alesiani et al.~\cite{Alesiani25b}.

  \item \textbf{Weak-Form \& Variational Constraints.}  
  Use integral/variational formulations instead of pointwise PDE residuals.  
  Includes \emph{WCN}~\cite{Chen25_weak} and \emph{KANtrol}~\cite{Aghaei24_kantorol}.

  \item \textbf{Alternative Problem Formulations.}  
  Modify the optimization objective or physical variables.  
  Includes \emph{AL-PKAN}~\cite{pde_Zhang24}, \emph{AIVT}~\cite{Toscano25_aivt},  
  \emph{cPIKAN+RBA+EVM}~\cite{KAN_pde_Shukla24}, \emph{KKAN+ssRBA}~\cite{Toscano24_kkan}, \emph{KAN-MHA}~\cite{Yang25}, and holomorphic–potential \emph{PIHKAN}~\cite{Clafa25}.

  \item \textbf{Physics-Guided Discovery \& Design.}  
  Physics-based losses used for inference, discovery, or design tasks.  
  Includes \emph{PKAN}~\cite{Guo25_equation} and physics-guided sensor design~\cite{Gong25_sensors}.

  \item \textbf{General PDE–Residual Models.}  
  Standard PINN-style residuals without special structural constraints.  
  Includes \emph{KAN-PINN}~\cite{Zhang25_comp}, \emph{Res-KAN}~\cite{Guo25}, \emph{Anant-KAN}~\cite{Menon25}, and \emph{PI-KAN}~\cite{Cui25_flow}.

\end{itemize}

\medskip
\noindent
In all these approaches, \emph{loss engineering}—choosing residual forms, designing constraint terms, and applying adaptive weighting strategies—serves as a powerful inductive bias. By aligning the optimization process with the underlying physics, these designs improve accuracy, stability, and extrapolation capabilities across a broad range of PDE-driven problems. Beyond modifying the loss, several works instead adapt the sampling or grid itself, yielding comparable accuracy gains through geometric rather than variational control.

\subsection{Adaptive Sampling \& Grids}\label{sample}
Adaptive sampling and grid refinement focus computational
effort where the solution is most difficult to capture (steep gradients, shocks, oscillations). 
One simple approach is multi-resolution training, where a model is trained on alternating coarse and fine point clouds to accelerate convergence and lower computational cost~\cite{Yang25_multiScale}. 
More advanced methods, long used in finite elements~\cite{Zienkiewicz89,Nguyen21}, adapt the model's internal grid or sampling based on runtime dynamics for complex geometries~\cite{pde_Wang24,Lei25}. 
In stochastic systems, Chen et al.~\cite{Chen25_weak} introduced \emph{adaptive collocation sampling}, selecting Gaussian test functions directly from observed data distributions.

\medskip
\noindent\textbf{Geometric/grid refinement with KANs.}  
Actor et al.~\cite{Actor25} treat spline \emph{knots} as a learnable grid, which can be refined during training:
\begin{enumerate}
  \item \emph{Multilevel refinement.} Training begins on a coarse knot grid and periodically switches to finer grids. Since spline spaces are nested (coarse $\subset$ fine), the coarse solution transfers exactly to the refined grid. This preserves training progress, adds capacity only where needed, and achieves lower error at the same FLOPs compared with single-level training.  
  \item \emph{Free-knot adaptation.} Knot locations are made trainable on $[a,b]$, with ordering enforced by a cumulative softmax:
  \[
  t_i \;=\;
  \underbrace{a}_{\text{left endpoint}}
  \;+\;
  \underbrace{(b-a)}_{\text{interval length}}
  \sum_{j=1}^{i}
  \underbrace{\mathrm{softmax}(s)_j}_{\text{ordered weights}},
  \qquad s\in\mathbb{R}^{n},
  \]
  ensuring that endpoints remain fixed ($t_0=a$, $t_n=b$).  
This naturally places knots in regions of high variation or non-smoothness.
Related analyses of free-knot spline behavior in KANs have been carried out by
Zheng et al.~\cite{Zheng2026}\footnote{\url{https://github.com/IcurasLW/FR-KAN}},
who derive bounds on the number of spline knots and propose a Free-Knots KAN
that improves training stability while reducing the number of trainable parameters.
\end{enumerate}

\medskip
\noindent On both regression and physics-informed tasks, multilevel schedules and free-knot adaptation consistently reduced errors relative to fixed grids and MLP baselines~\cite{Actor25}.

\begin{figure}[!ht]
\centering
\subfigure[Loss vs.\ epochs]{%
 \label{pde_Rigas24a}
 \includegraphics[width=3.25in]{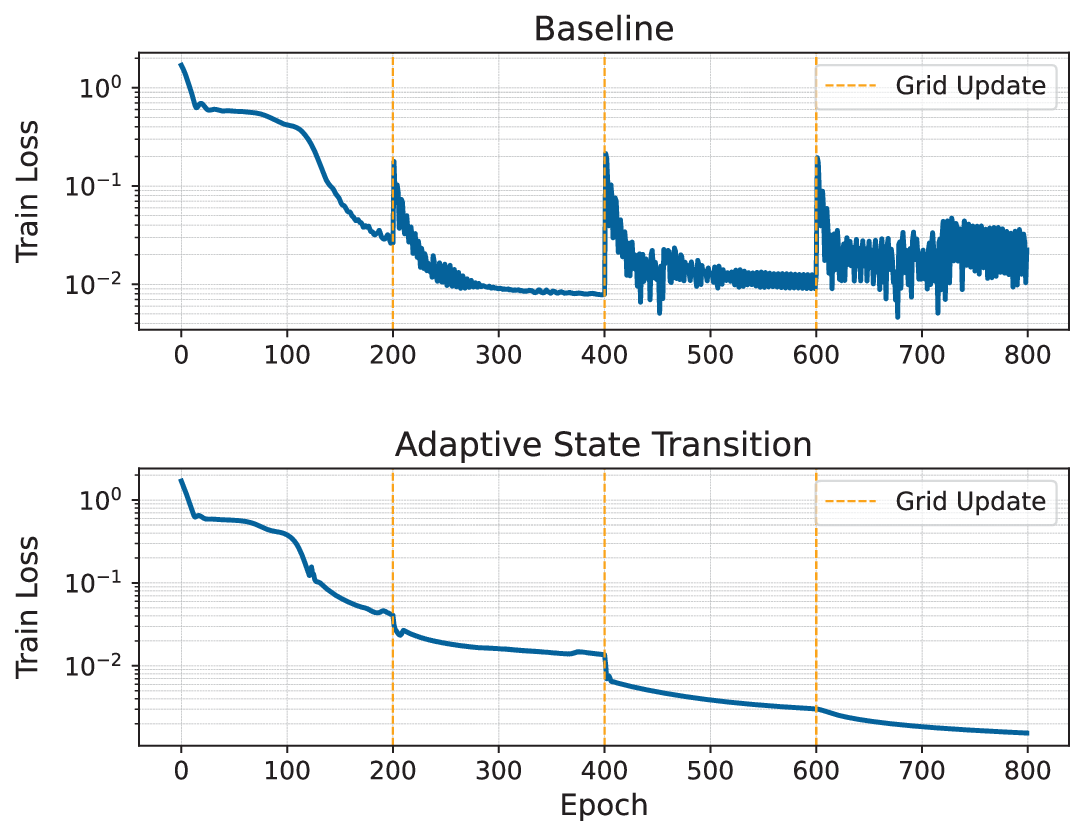}}
\subfigure[Residual-based adaptive sampling]{%
 \label{pde_Rigas24b}
 \includegraphics[width=1.86in]{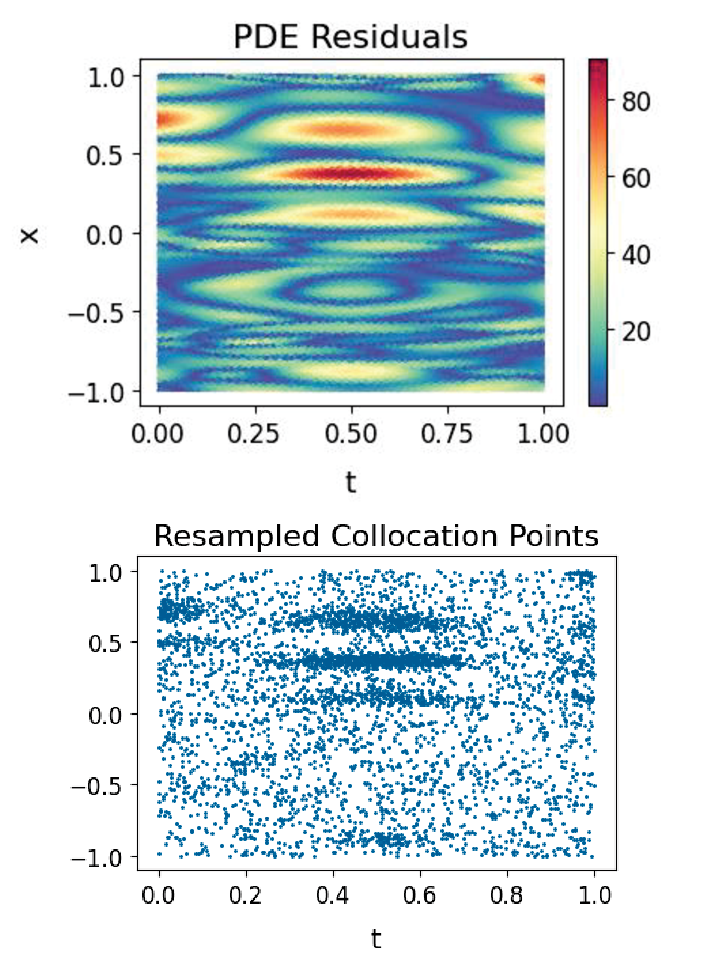}}
\caption{Adaptive PIKAN training after~\cite{pde_Rigas24}. 
(a) Loss vs.\ epochs: adaptive state transition smooths grid updates. 
(b) RAD concentrates collocation points where PDE residuals are largest.}
\label{pde_Rigas24c}
\end{figure}

\medskip
\noindent\textbf{Grid–adaptive PIKANs.}  
Rigas et al.~\cite{pde_Rigas24, pde_Rigas24f} extend this idea by
 adapting not only the spline grid but also the collocation
  points, so resolution follows the PDE residual. 
More recently, a dynamic grid-adaptation framework was proposed that treats knot allocation as a density-estimation problem governed by importance density functions, allowing curvature or training metrics to determine grid resolution during learning~\cite{Rigas2026}.
\begin{enumerate}
  \item \emph{Grid extension with smooth state transition.} Start with few spline intervals ($G$) and periodically 
  increase $G$. At each extension, an adaptive optimizer state transition preserves momentum and avoids loss spikes,
   while the finer grid captures unresolved structure (Fig.~\ref{pde_Rigas24a}).  
  \item \emph{Residual-based adaptive resampling (RAD).} 
  Compute residuals on a dense probe set $S$ and convert 
  them into sampling probabilities: larger residuals
   $\Rightarrow$ higher sampling chance. A new collocation
    set is then drawn accordingly (Fig.~\ref{pde_Rigas24b}).  
\end{enumerate}
Since KAN basis functions are tied to the knot grid, 
jointly adapting the grid and collocation points reduces 
projection error and concentrates model capacity in residual hotspots.

\subsection{Domain decomposition}\label{decomposition}
\begin{figure}[htbp]
\centering
\includegraphics[width=0.9\textwidth]{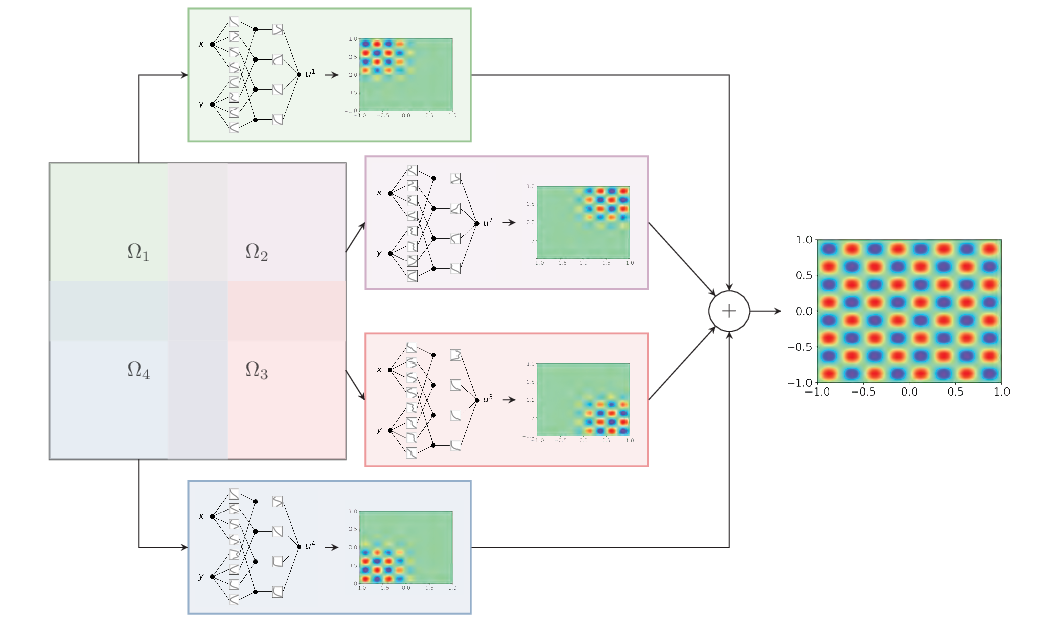}
\caption{Schematic of FBKAN domain decomposition and PoU blending. Overlapping subdomains $\Omega_j$ host lightweight local KANs $K_j$, smoothly combined via PoU weights $\omega_j(x)$. This enables parallel training and per-subdomain knot/hyperparameter choices (adapted  from \cite{pde_fbkan_Howard24}).}
\label{fig:fbkan}
\end{figure}

\noindent Splitting a difficult global map into easier local ones and blending them smoothly often improves accuracy. For example,
 temporal subdomaining can sharpen NTK conditioning and 
 convergence \cite{Faroughi25}. Finite-basis KANs~\cite{pde_fbkan_Howard24}, as discussed in
  Sec.~\ref{Finite}, achieve this by decomposing $\Omega$ (or $\Omega\times[0,T]$) into $L$ overlapping 
  subdomains $\{\Omega_j\}_{j=1}^L$ with a
  smooth partition of unity (PoU) $\{\omega_j\}_{j=1}^L$ such that
\[
\underbrace{\mathrm{supp}(\omega_j)}_{\text{support of local weight}}=\Omega_j,
\qquad 
\underbrace{\sum_{j=1}^L \omega_j(x)\equiv 1}_{\text{partition of unity}}.
\]
A schematic is shown in Fig.~\ref{fig:fbkan}.  

\noindent A convenient 1D choice uses normalized ``cosine bumps.'' With domain length $l$ and overlap ratio $\delta>1$:
\[
\hat{\omega}_j(x)=\bigl[1+\cos\!\big(\pi(x-\mu_j)/\sigma_j\big)\bigr]^2,
\quad
\mu_j=\underbrace{\tfrac{l(j-1)}{L-1}}_{\text{center of bump}},
\quad
\sigma_j=\underbrace{\tfrac{\delta\,l/2}{L-1}}_{\text{width/overlap}},
\]
\[
\omega_j(x)=
\underbrace{\frac{\hat\omega_j(x)}{\sum_{k=1}^L\hat\omega_k(x)}}_{\text{normalized local weight}}.
\]
In $d$ dimensions, the construction extends by taking products 
across coordinates.  

\noindent The global surrogate is then the PoU blend of local KANs:
\[
f_\Theta(x)=\sum_{j=1}^L 
\underbrace{\omega_j(x)}_{\text{local weight}}\,
\underbrace{K_j(x;\boldsymbol{\theta}_j)}_{\text{local KAN}},
\]
where each $K_j$ is trained only on $\Omega_j$, using its \emph{own} knot grid induced by $\omega_j$ (basis/grid choices per Sec.~\ref{Finite}).  

\noindent A basic error estimate illustrates the benefit:
\[
\|u-f_\Theta\| \;\le\; 
\sum_{j=1}^L 
\underbrace{\|\omega_j\|_\infty}_{\text{PoU weight}}\,
\underbrace{\|u-K_j\|_{\Omega_j}}_{\text{local error}},
\]
showing that reducing each local error reduces the global error.  Increasing $L$ or the tunable overlap $\delta$ 
concentrates resolution where it is most needed. Each local KAN operates on a smaller, better-conditioned subproblem, allowing parallel training with improved numerical stability.
In practice, per-subdomain grids adapt to local scales and consistently yield lower relative $L_2$ errors 
than single-domain KANs at the same model capacity \cite{pde_fbkan_Howard24}.

\subsection{Function decomposition}\label{function_decompose}

Howard et al.~\cite{pde_Howard24} introduce a simple but effective fidelity–space split.  
A low-fidelity surrogate $K_L$ is first trained on a large LF dataset $\{(x_i,f_L(x_i))\}_{i=1}^{N_{LF}}$ and then frozen.  
A high-fidelity model $K_H$ is trained on a much smaller HF dataset $\{(x_j,f_H(x_j))\}_{j=1}^{N_{HF}}$, using both $x$ and $K_L(x)$ as inputs.  
The HF predictor blends a linear LF$\!\to\!$HF trend with a nonlinear correction:
\[
K_H(x)
=(1-\alpha)\,K_{\ell}\big(x,K_L(x)\big)
+\alpha\,K_{nl}\big(x,K_L(x)\big),
\]
where $\alpha\in[0,1]$ is trainable, $K_{\ell}$ is the linear head, and $K_{nl}$ the nonlinear correction.

The HF loss balances accuracy, linearity preference, and regularization:
\[
\mathcal{L}_{HF}
=\frac{1}{N_{HF}}\sum_{j=1}^{N_{HF}}
\!\big(K_H(x_j)-f_H(x_j)\big)^2
+\lambda_\alpha\,\alpha^{\,n}
+w\,\|\Phi_{nl}\|_2^2.
\]
Here $\lambda_\alpha>0$ and $n\in\mathbb{N}$ control the preference for the linear block, while $w>0$ regularizes the nonlinear one.

\noindent
The intuition is that the LF model captures the dominant structure, while the HF nonlinear head corrects only the residual.  
This reduces overfitting and stabilizes training when HF data are scarce.  
MFKAN achieved lower relative $L_2$ errors than single-fidelity KANs under the same HF budget~\cite{pde_Howard24}.  

The same fidelity split also benefits PIKANs, where $K_L$ serves as a coarse physics surrogate and the HF head is trained only on residual or boundary losses, making fidelity decomposition complementary to spatial domain decomposition (Sec.~\ref{decomposition}).  
Related coupled PIKAN strategies sequentially train subnetworks representing different subsystems to stabilize optimization~\cite{Liu2026_PI}, while multi-branch physics-informed KANs decompose interacting physical subsystems into parallel KAN branches with physics constraints~\cite{Hu2026}.

The same fidelity split also benefits PIKANs, where $K_L$ serves as a coarse physics surrogate and the HF head is trained only on residual or boundary losses, making fidelity decomposition complementary to spatial domain decomposition (Sec.~\ref{decomposition}).

\noindent
In a related direction, Jacob et al.~\cite{pde_jacob24}\footnote{\url{https://github.com/pnnl/spikans}} propose a \emph{separable} PIKAN (SPIKAN), expressing the solution as a sum of products of 1D KAN factors—one per coordinate—before combining them into a global field.
Bühler et al.~\cite{Buhler25_regression} propose KAN-SR, a symbolic regression framework inspired by Udrescu et al.~\cite{Udrescu20}.  
The method checks for separability or symmetry before modeling.  
If such structure exists, the function is decomposed into smaller subproblems, each fit with a compact KAN, and then recomposed into a closed-form symbolic expression.  
This divide--and--conquer strategy improves interpretability and robustness in noisy or high-dimensional settings.

\begin{figure}[htbp]
\centering
\includegraphics[width=0.7\textwidth]{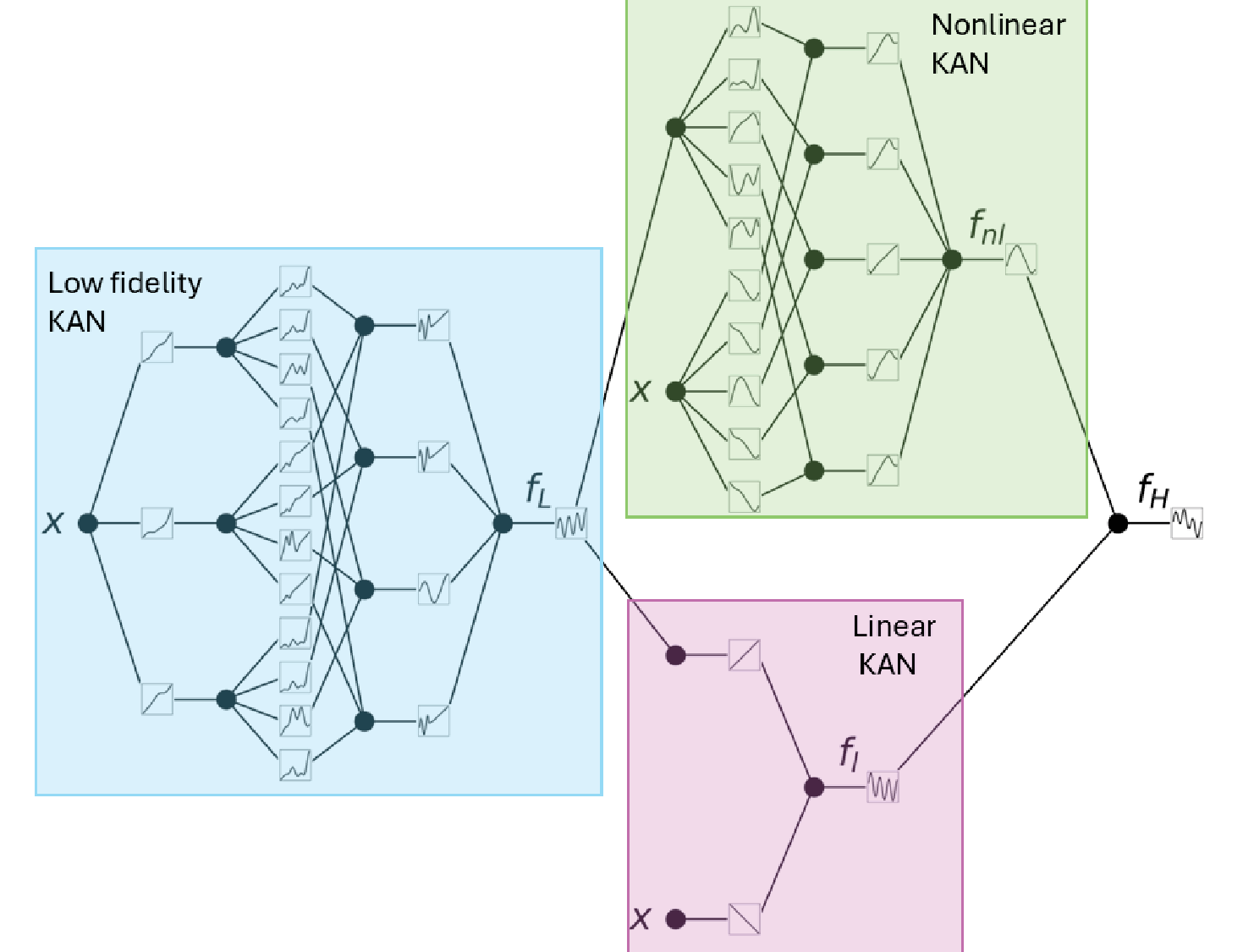}
\caption{Multifidelity KAN (MFKAN) (adapted from \cite{pde_Howard24, Liu24}). 
A low-fidelity KAN $K_L$ is trained and then frozen. Linear and nonlinear heads $K_{\ell}$ and $K_{nl}$ learn LF$\!\to\!$HF correlations and residual corrections, which are blended to form $K_H$.}
\label{pde_Howard24_color}
\end{figure}

\subsection{Hybrid, Ensemble, and Data-Integrated Models}\label{hybrid}

\noindent
A complementary line of work boosts accuracy not by modifying KAN bases, but by
\emph{combining KAN modules with other neural components}—MLPs, attention blocks, or ensemble
mechanisms.  
These hybrids exploit KAN’s structured 1D functional mappings while leveraging the expressive power
of deep networks.

\begin{figure}[htbp]
\centering
\includegraphics[width=0.7\textwidth]{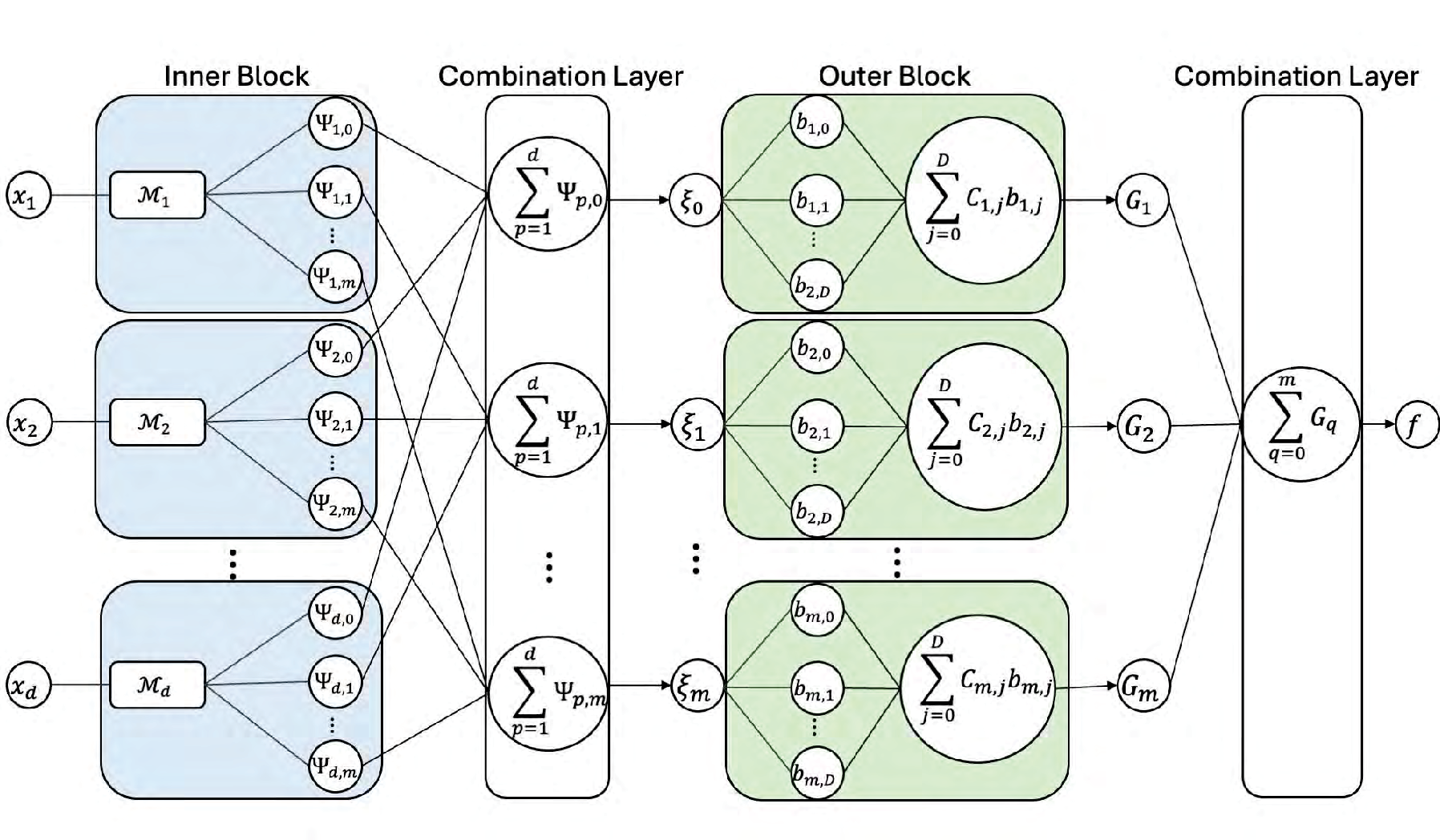}
\caption{KKAN architecture (adapted  from \cite{Toscano24_kkan}). Each input coordinate \(x_p\) 
 These are summed across dimensions to form \(\xi_m\), which are expanded via basis functions in the outer block. A final
 summation yields the output. Bases may include Chebyshev/Legendre polynomials, sine series, or RBFs. This
  “MLP inside + basis outside” layout mirrors 
the Kolmogorov–Arnold recipe.}
\label{Toscano24_kkan_ar}
\end{figure}

\begin{itemize}

  \item \textbf{Function-combination KANs.}  
 FC-KAN combines multiple basis families (e.g., B-splines, wavelets, and radial basis functions)
  through element-wise operations such as sums, products, and concatenations, enabling richer
  functional representations within a single KAN layer \cite{Ta2026}\footnote{\url{https://github.com/hoangthangta/FC_KAN}}.
  
  \item \textbf{MLP--KAN hybrids.}  
  These architectures use MLPs for broad feature extraction and KANs for structured refinement.  
  Examples include the MLP--KAN mixture-of-experts of He et al.~\cite{He24}, the HPKM-PINN
parallel fusion model of Xu et al.~\cite{Xu25}, and two-stage medical analysis pipelines
that couple foundation-model segmentation with KAN-based classification~\cite{Wang2026}.

  \item \textbf{Task-specific KAN replacements.}  
  Kundu et al.~\cite{Kundu24_quantom} (KANQAS)\footnote{\url{https://github.com/Aqasch/KANQAS_code}}
  replace the MLP in a Double Deep Q-Network with a KAN module for quantum architecture search.

  \item \textbf{Structured functional hybrids.}  
  KKAN~\cite{Toscano24_kkan} inserts small per-dimension MLPs to generate learned
  one-dimensional features \(\Psi_{p,q}(x_p)\), sums them across dimensions, and applies a simple
  outer basis expansion (Chebyshev, Legendre, sinusoidal, or RBF).  
  This yields a pipeline
  \[
  \text{inputs} \rightarrow \text{inner MLPs} \rightarrow \text{dimensionwise sum}
  \rightarrow \text{basis expansion} \rightarrow \text{output},
  \]
  combining expressive learned features with a stable, low-bias basis.  
  KKAN reports consistent gains over vanilla KANs and MLPs across regression, operator learning,
  and PDE benchmarks (Fig~\ref{Toscano24_kkan_ar}).
  
  \item  \textbf{Symbolic-regression hybrids.}
  Howard et al.~\cite{Howard26} propose \emph{SINDy-KANs}, which combine KANs with SINDy-style sparse regression to identify parsimonious symbolic representations at the level of the learned activation functions. This hybrid design targets equation discovery and dynamical-system identification, showing that KANs can be coupled not only with neural modules, but also with sparse scientific model-discovery frameworks.

\end{itemize}

\subsection{Sequence and Attention Hybrids}\label{seq-hybrids}

\noindent
KANs are increasingly combined with sequence encoders and attention mechanisms to capture long-range temporal structure, global context, or multi-scale interactions.  
Most architectures fall into three categories:  
(i) attention-enhanced KANs, (ii) sequence-to-field encoder–decoder hybrids, and (iii) generative or data-augmentation models that stabilize training.

\begin{itemize}

  \item \textbf{Attention-enhanced KANs.}  
  These models insert KAN blocks into attention layers to improve multi-scale representation.  
Includes \emph{KAN–MHA}~\cite{Yang25}, \emph{FlashKAT} (Group-Rational KAN in Transformers)~\cite{Raffel25},  
\emph{Attention-KAN-PINN} for battery SOH forecasting~\cite{Wei26_battery}, and wavelet-attention hybrids such as \emph{DWSen}~\cite{Li2026}.

  \item \textbf{Sequence encoders and encoder–decoder hybrids.}  
  These pair temporal encoders with KAN decoders to map sequences into spatial or field outputs.  
  Includes \emph{AL-PKAN} (GRU$\to$KAN)~\cite{pde_Zhang24}, \emph{GINN-KAN}~\cite{pde_Ranasinghe24},  
\emph{KAN-ODE} for continuous-time modeling~\cite{pde_Koeing24}, and physics-informed temporal–spatial hybrids combining TCN/GRU encoders with KAN-based operators for real-time structural response prediction~\cite{Liu2026_TS}, as well as masked self-attention architectures for multimodal fatigue detection~\cite{Xu2026}.

  \item \textbf{Generative / data-augmentation hybrids.}  
  These integrate KANs with GAN or adaptive-activation mechanisms to improve learning from sparse or noisy data.  
  Includes \emph{AAKAN–WGAN}~\cite{Shen25}, which uses a Wasserstein GAN to augment training data.
  
\end{itemize}

\noindent
Across all designs, the auxiliary module supplies memory, context, or synthetic data, while the KAN block provides a compact and interpretable functional mapping at the output stage.

\subsection{Discontinuities and Sharp Gradients}\label{subsec:discsharp}

\noindent
KAN variants approach discontinuities, shocks, and steep gradients through three main mechanisms:  
(i) using bases that naturally represent non-smooth features,  
(ii) attention or residual reweighting that concentrates learning where gradients are large, and  
(iii) physics-driven stabilization for shock-dominated PDEs.

\begin{itemize}

  \item \textbf{Non-smooth or localized bases.}  
  Designed to capture kinks, singularities, or asymptotic regimes.  
  Includes \emph{SincKAN}~\cite{Yu24}, \emph{rKAN}~\cite{Aghaei24_rkan}, and \emph{KINN} for stress singularities~\cite{pde_Wang24}.

  \item \textbf{Attention and adaptive residuals.}  
  Focuses computational effort on regions with steep gradients.  
  Includes \emph{KAN--MHA}~\cite{Yang25}, \emph{KKAN+ssRBA}~\cite{Toscano24_kkan}, and \emph{AIVT}~\cite{Toscano25_aivt}.

  \item \textbf{Physics-driven shock stabilization.}  
Includes \emph{cPIKAN+EVM/RBA}~\cite{KAN_pde_Shukla24}, \emph{Buckley--Leverett KANs}~\cite{Kalesh25},  
\emph{EPi--cKAN} for elasto-plastic transitions~\cite{Mostajeran24},and physics-informed KAN solvers for the Vlasov--Poisson system in cosmological structure formation~\cite{Cerardi2026}.

\end{itemize}

\noindent
Overall, these approaches mitigate non-smoothness by introducing localized bases (Sinc, rational), adaptive residual focusing, or physics-aware stabilization that controls oscillations near shocks.

\medskip
\noindent\textbf{Edge-aware KANs.}
Lei et al.~\cite{Lei25} embed an explicit \emph{edge activation} inside each KAN layer, enabling sharp jumps while keeping the rest of the representation smooth:
\[
\psi(x)
=
w_t\,\tanh\!\big(\alpha(x-\beta)\big)
+
w_s\sum_i c_i\,S_i(x),
\]
where $\alpha,\beta,w_t,w_s,c_i \in\mathbb{R}$ and $S_i(\cdot)$ are B-spline basis functions.

\noindent
The $\tanh$ term places and sharpens the discontinuity, while the spline component fits the surrounding smooth regions. Incorporated into a PIKAN framework, this mechanism reduces spectral bias and yields more stable representations of steep fronts, a persistent difficulty for both MLPs and classical spline-based KANs.

\subsection{Optimization \& Adaptive Training}\label{optmization}

\noindent
Training KANs—especially in physics-informed or data-scarce settings—relies heavily on optimization choices. Two themes dominate: (i) hybrid optimization schedules and (ii) Bayesian/probabilistic methods.

\medskip
\noindent\textbf{Hybrid and staged optimizers.}
A recurring finding is that optimizer choice strongly influences KAN convergence.  
Kalesh~\cite{Kalesh25} showed that Adam$\to$L-BFGS schedules improve sharp-front resolution in two-phase flow PINNs.  
Mostajeran~\cite{Mostajeran25} reported similar gains on large-domain PDEs, while Faroughi and Mostajeran~\cite{Faroughi25} linked these improvements to better NTK conditioning.  
Daryakenari~\cite{Daryakenari25} benchmarked several first/second-order combinations (e.g., RAdam$\to$BFGS) with warm-up and mixed precision, noting improved stability for both PINNs and PIKANs.  
Zeng et al.~\cite{KAN_pde_Zeng24} refined these results by identifying regimes where Adam alone or L-BFGS alone is preferable.  
Overall, staged optimizers—early Adam-type exploration followed by quasi-Newton refinement—consistently yield more reliable KAN training.

\medskip
\noindent\textbf{Bayesian and probabilistic methods.}
Probabilistic formulations are increasingly used to calibrate KANs or quantify prediction uncertainty.  
Lin~\cite{Lin25_geo} employed Bayesian optimization to tune RBF--KAN hyperparameters.  
Giroux and Fanelli~\cite{pde_bayesian_Giroux24}\footnote{\url{https://github.com/wmdataphys/Bayesian-HR-KAN}} introduced variational Bayesian inference for Higher--Order ReLU--KANs, placing priors on both weights and basis parameters; KL regularization improved robustness on stochastic Poisson and Helmholtz problems.  
Hassan~\cite{Hassan25_Bayesian} proposed a spline-based \emph{Bayesian--KAN}, assigning Gaussian priors \(c_n\!\sim\!\mathcal{N}(\mu_n,\sigma_n^2)\) to the activation coefficients for uncertainty quantification across layers.

A complementary direction is the probabilistic KAN of Polar and Poluektov~\cite{Polar25}\footnote{\url{https://github.com/andrewpolar/pkan}}, which replaces Bayesian priors with an ensemble built by \emph{divisive data re-sorting}.  
This approach learns multiple KANs over recursively refined subsets of the data, capturing full input-dependent output distributions—including multi-modality—while remaining computationally lightweight and highly parallelizable.

\medskip
\noindent\textbf{Initialization and depth-stabilized training.}
Beyond optimizer choice and probabilistic tuning, Rigas et~al.~\cite{Rigas25_deep}\footnote{\url{https://github.com/srigas/RGA-KANs}}
show that deep Chebyshev-based KANs (cPIKANs) fail primarily due to variance blow-up across layers.
They introduce a \emph{basis-agnostic Glorot-like initialization} that preserves forward/backward variance and a
\emph{Residual--Gated Adaptive} block that stabilizes deep gradient flow.  

For spline KANs, Rigas et~al.~\cite{Rigas25_init}\footnote{\url{https://github.com/srigas/KAN_Initialization_Schemes}}
demonstrate that Glorot-style and power-law initializations substantially improve NTK conditioning and convergence
across regression and PDE benchmarks, underscoring initialization as a key design lever for scalable KAN depth.
Related efficiency improvements can also be achieved through feature enrichment strategies such as FEKAN~\cite{Menon2026_FEKAN}, which augment the input representation before the KAN mapping to accelerate convergence without modifying the underlying architecture.

\medskip
\noindent\textbf{Advanced quasi-Newton methods.}
Kiyani et~al.~\cite{Kiyani25} show that self-scaled BFGS/Broyden optimizers markedly outperform Adam and standard BFGS on stiff PDEs.  
Their results reveal two effects: the PDE residual creates a highly ``crinkled'' loss landscape where first-order methods stall, and stronger optimizers cut this optimization error so effectively that \emph{smaller networks} reach equal or better accuracy.  
\emph{One-liner: self-scaled quasi-Newton methods tame the PDE landscape and enable accurate, compact PIKANs.}
A related line of work by Poluektov and Polar ~\cite{Polar25_Newton}\footnote{\url{https://github.com/andrewpolar/det5}} replaces gradient training entirely with a Newton--Kaczmarz update on the spline/Gaussian coefficients, yielding a solver that is less sensitive to initialization and efficient for large datasets.

\medskip
\noindent\textbf{Variational basis-size learning.}
Alesiani et~al.~\cite{Alesiani25a} propose \emph{INFINITYKAN}, which treats the number of basis functions in each univariate map as a latent variable optimized via mean-field variational inference.  
Rather than fixing basis width, Poisson-distributed variables \( \lambda_\ell \) determine layer-wise cardinality, allowing the representational capacity to expand or contract during training.  
Experiments show basis-size selection is a task-dependent and learnable component of KAN optimization.

\noindent
These works highlight the role of probabilistic reasoning for calibration, interpretability, and reliable prediction in settings with randomness or limited data.


\section{Efficiency Improvement}\label{sec:efficiency-improvement}

\noindent
Efficiency gains in KANs arise from two complementary directions:  
(i) exploiting hardware parallelism and compiler support, and  
(ii) reducing algebraic and basis-level complexity.

\subsection{Parallelism, GPU, and JAX engineering}\label{parallel}
\noindent
A major line of work accelerates KANs by making their core operations GPU-friendly.  
Iterative B\hbox{-}spline evaluations are replaced by non-iterative ReLU-power activations~\cite{Qiu25} or reformulated as CUDA-optimized matrix kernels~\cite{Qiu24,KAN_pde_So24}.  
Model-level restructuring further boosts throughput: dual-matrix compression with Random Fourier Features~\cite{Zhang25}, custom backward kernels to reduce memory stalls~\cite{Raffel25}, and GPU-accelerated B\hbox{-}spline evaluation~\cite{Coffman25}.  
Architectural parallelism also plays a role, including multi-branch KAN--MLP hybrids~\cite{Xu25} and domain-decomposition designs for PINNs~\cite{KAN_pde_Shukla24,pde_fbkan_Howard24,pde_jacob24}.  
At the framework level, JAX implementations~\cite{Daryakenari25,pde_Rigas24} leverage \texttt{jit}/XLA and optimized autodiff to deliver substantial end-to-end speedups, while specialized hardware designs such as computing-in-memory architectures using Gaussian-like memory cells further improve energy-efficient KAN inference~\cite{Wen2026}.

\subsection{Matrix optimization and parameter-efficient bases}\label{matrix}
\noindent
A second strategy reduces computational cost by simplifying basis functions and matrix operations.  
Replacing spline bases with faster alternatives—ReLU-power~\cite{Qiu25,Qiu24,KAN_pde_So24},  
Chebyshev/Jacobi/Legendre polynomials~\cite{KAN_pde_Shukla24,Guo24,Mostajeran24,Mostajeran25,pde_Wang24},  
RBFs~\cite{Lin25_geo,pde_Koeing24}, or wavelets~\cite{pde_Patra24}—lowers FLOPs and parameter counts  
while maintaining approximation quality.  
Additional savings come from structural compression: dual-matrix fusion with RFFs~\cite{Zhang25},  
sparsity-promoting regularizers~\cite{Guo25}, hierarchical knot refinement~\cite{Actor25},  
and topology search via Differential Evolution~\cite{Li25_DEKAN}.  
Operator-aware designs further reduce cost by mixing spectral and spatial representations  
to keep matvecs near-diagonal~\cite{Lee25_operator}, limiting mode mixing and memory traffic  
on variable-coefficient PDEs.  
Anant\text{-}Net~\cite{Menon25}\footnote{\url{https://github.com/ParamIntelligence/Anant-Net}}  
extends these ideas with dimension-wise tensor-product sweeps and selective differentiation,  
both of which reduce high-dimensional autodiff complexity and are readily transferable to KANs.  
In a related direction, lookup multivariate KANs (lmKANs) replace univariate inner functions  
with multivariate spline lookup tables and CUDA-efficient evaluation, providing a  
parameter-rich yet computationally light alternative to linear layers~\cite{Pozdnyakov25}%
\footnote{\url{https://github.com/schwallergroup/lmkan}}.

\noindent
Together, these techniques make KANs increasingly competitive in large- and high-dimensional settings by aligning basis, algebra, and implementation with modern hardware.

\section{Sparsity \& Regularization}\label{Regularization}

\noindent
Regularization plays a central role in improving the stability and generalization of KANs, especially in noisy, high-dimensional, or ill-posed regimes. Existing approaches fall into four broad families: sparsity penalties, Lipschitz/complexity control, weight decay, and implicit or dropout-style regularizers.

\medskip
\noindent\textbf{L1 sparsity and entropy balancing.}
Liu et al.~\cite{Liu24} introduce an activation-level $\ell_1$ penalty,
\begin{equation}
\|\varphi_{ij}\|_1 = \tfrac{1}{N_p}\!\sum_{s=1}^{N_p}\!|\varphi_{ij}(x^{(s)})|,
\qquad
\|\Phi\|_1=\sum_{i,j}\|\varphi_{ij}\|_1,
\end{equation}
supplemented by an entropy balancing term,
\begin{equation}
S(\Phi)=-\sum_{i,j}\frac{\|\varphi_{ij}\|_1}{\|\Phi\|_1}
\log\!\Big(\frac{\|\varphi_{ij}\|_1}{\|\Phi\|_1}\Big),
\end{equation}
leading to the regularized objective
\begin{equation}
\mathcal{L}_{\text{total}}
=\mathcal{L}_{\text{pred}}
+\lambda\big(\mu_1\!\sum_\ell\|\Phi_\ell\|_1+\mu_2\!\sum_\ell S(\Phi_\ell)\big).
\end{equation}
Variants such as \emph{EfficientKAN}~\cite{EfficientKAN}\footnote{\url{https://github.com/Blealtan/efficient-kan}}
and \emph{DeepKAN}~\cite{DeepKAN}\footnote{\url{https://github.com/sidhu2690/Deep-KAN}}
apply $\ell_1$ directly to weights; the same scheme appears in symbolic regression~\cite{Wang25} and PDE learning~\cite{Guo25}.  
For Neural ODEs, Koenig et al.~\cite{pde_Koeing24} use a layerwise $\ell_1$ followed by pruning,
\begin{equation}
\mathcal{L}_{\text{total}}
=\mathcal{L}_{\text{fit}}+\lambda_{\text{L1}}\|\theta\|_1,
\end{equation}
eliminating redundant connections without accuracy loss.
Application-driven sparsity can also arise from the data representation itself; for example, SP-KAN employs sparse sinusoidal perception layers aligned with the intrinsic sparsity of infrared targets \cite{Yuan2026}\footnote{\url{https://github.com/xdFai}}.

\medskip
\noindent\textbf{Structured sparsity for symbolic regression.}
Bühler et al.~\cite{Buhler25_regression} propose a composite regularizer for KAN-SR combining:  
(i) a magnitude penalty to suppress unused inputs,  
(ii) row/column entropies to localize attention, and  
(iii) an $\ell_1$ penalty on linear weights.  
This promotes compact, interpretable expressions while preventing the overspecification typical of symbolic regression networks.

\medskip
\noindent\textbf{Bayesian sparsity.}
Zou and Yan~\cite{Zou25_Probabilistic} introduce \emph{Probabilistic KANs (PKAN)} and their sparse counterpart \emph{SSKAN}, using Gaussian-process priors and spike--and--slab priors, respectively.  
These provide principled sparsity and uncertainty quantification, forming a Bayesian alternative to $\ell_1$-based pruning.

\medskip
\noindent\textbf{Lipschitz-driven regularization.}
Li et al.~\cite{Li25_lipschitz} show that KAN generalization is governed by a \emph{Lipschitz complexity} depending on per-layer Lipschitz constants and mixed tensor norms.  
Their \emph{LipKAN} architecture inserts 1-Lipschitz layers and replaces $\ell_1$ with a theoretically motivated $\ell_{1.5}$ penalty, providing tighter complexity control and improved generalization on complex distributions.

\medskip
\noindent\textbf{Low-rank RKHS regularization.}
Zhang and Zhou~\cite{Zhang25_general} establish generalization bounds for KANs whose activations lie in a \emph{low-rank} RKHS (e.g., Wendland functions~\cite{Wendland04}), with complexity scaling in the activation \emph{rank} rather than width.  
This motivates LoRA-style rank-constrained fine-tuning of pre-trained KAN activations: maintaining expressivity while regularizing model capacity.

\vspace{8pt}

\textbf{Feature-conditioning and scale regularization.}
For Gaussian KANs, the scale parameter can also be viewed as an implicit regularizer through the conditioning of the first-layer feature map. 
Since the first layer is evaluated directly on the input variables, its empirical feature matrix $\Phi$ induces the additive kernel
\[
K_0=\Phi\Phi^\top=\sum_{i=1}^{d}\Phi^{(i)}\bigl(\Phi^{(i)}\bigr)^\top .
\]
If $\varepsilon$ is too large, the Gaussian features become nearly constant over the sample set, so $K_0$ approaches a rank-one matrix and distinct inputs may become indistinguishable before entering deeper layers. 
This creates a first-layer bottleneck that cannot be corrected downstream. 
Therefore, constraining $\varepsilon$ to a well-conditioned range provides a practical form of regularization, controlling both feature collapse and numerical instability \cite{Amir_GKAN, Amir_PUGKAN}.

\medskip
\noindent\textbf{L2 weight decay and extensions.}
Shen et al.~\cite{Shen25} incorporate $\ell_2$ weight decay,
\begin{equation}
\mathcal{L}_{\mathrm{base}}
=\mathcal{L}_{\mathrm{pred}}+\lambda_1\mathcal{L}_{\mathrm{L2}},
\end{equation}
along with temporal smoothness and mutual-information terms,
\begin{equation}
\mathcal{L}_{\mathrm{AAKAN}}
=\mathcal{L}_{\mathrm{base}}
+\lambda_2\mathcal{L}_{\mathrm{time}}
+\lambda_3\mathcal{L}_{\mathrm{MI}},
\end{equation}
to stabilize dynamics.  
Small $\ell_2$ penalties are similarly used in physics-informed KANs~\cite{KAN_pde_Shukla24,Toscano25_aivt} to prevent overfitting while preserving optimization stability.

\medskip
\noindent\textbf{Implicit and dropout-style regularizers.}
Implicit regularization arises when activations themselves bound outputs.  
Daryakenari et al.~\cite{Daryakenari25} show that nested activations such as $\tanh(\tanh(x))$ naturally confine outputs, smoothing gradients in Chebyshev-based KANs.  
Altarabichi et al.~\cite{Altarabichi24} propose \emph{DropKAN}, which masks post-activation outputs rather than inputs, injecting noise directly into spline evaluations and offering more stable regularization than standard dropout.

\section{Convergence and Scaling Laws}\label{convergence}

\noindent
Classical approximation theorems guarantee that neural networks can approximate any continuous function, but provide little insight into convergence \emph{rates}. Recent KAN studies fill this gap by deriving Sobolev-space error bounds, analyzing gradient-flow dynamics via NTK theory~\cite{Jacot18,Wang22}, and documenting empirical scaling laws~\cite{Basina24}.

\subsection{Theoretical Approximation Rates and Empirical Scaling Laws}
\label{theory}

\noindent\textbf{Sobolev and Besov rates.}
Wang~\cite{Wang25} showed that deep spline-based KANs achieve a Sobolev approximation
rate that doubles the classical parameter-rate
\(\mathcal{O}(P^{-s/d})\) of standard neural networks and kernel approximators,
yielding \(\mathcal{O}(P^{-2s/d})\) convergence
\cite{DeVore98,DeVore21}.
This accelerated rate reflects the interaction between the compositional structure
of KANs and the expressive power of high-order spline bases.

Formally, the result is obtained by combining the exact reparametrization of
\(\mathrm{ReLU}^k\) networks by KANs (Theorem~\ref{thm:mlp_to_kan}) with sharp
approximation bounds for very deep piecewise-polynomial MLPs on Sobolev spaces
\cite{Wang25}.
Through this equivalence, established approximation theory for deep
\(\mathrm{ReLU}^k\) networks transfers directly to KAN architectures, leading to the
following corollary.

\begin{corollary}[Sobolev approximation rate for deep KANs {\cite{Wang25}}]
\label{cor:KAN_sobolev}
Let \(\Omega\subset\mathbb{R}^d\) be a bounded domain with smooth boundary and
\(s>0\).
There exists a fixed width \(W_0=W_0(d)\) such that for any
\(f\in W^{s}(\Omega)\) and any depth \(L\ge 1\), there exists a KAN \(g\) with width
\(W_0\), depth \(L\), grid size \(G=2\), and \(k\)-th order B-spline basis functions
satisfying
\[
\|f-g\|_{L^{p}(\Omega)} \;\le\; C\,L^{-2s/d},
\]
where \(C\) is a constant independent of \(L\).
\end{corollary}

\noindent
Kratsios et al.~\cite{Kratsios25} extend this result to \emph{Besov}
 spaces (\cite{DeVore88}), showing that spline-based Res-KANs achieve the optimal Besov
approximation rate on Lipschitz and fractal sets with dimension-free sample
complexity.
Complementing these approximation-centric results, Liu et al.~\cite{Liu25_convergence}
establish the first minimax \emph{statistical} rates for KAN regression: spline-based
additive and hybrid KANs achieve the optimal
\[
\mathcal{O}\!\left(n^{-2r/(2r+1)}\right)
\]
for Sobolev-\(r\) functions, together with an optimal knot-scaling rule
\(k_n\asymp n^{1/(2r+1)}\).
These three perspectives jointly show that KANs match the best-known approximation
and estimation rates of classical spline and sieve methods while retaining
neural-network–style compositional structure.

\medskip
\noindent\textbf{Empirical scaling.}
Liu~\cite{Liu24} observed that KAN RMSE follows a power law
\[
\ell \propto N^{-\alpha},
\]
with task-dependent $\alpha>0$. Increasing basis resolution (e.g., spline grid size $G$) consistently improves accuracy more than widening layers, echoing the theoretical depth benefit in~\cite{Wang25} and the statistical spline-resolution rule in~\cite{Liu25_convergence}.  
Together, these findings indicate that KANs achieve fast decay in both theory (Sobolev/Besov/minimax rates) and practice (empirical power laws), though~\cite{Kratsios25} emphasizes that many of these gains reflect favorable optimization dynamics rather than asymptotic expressivity alone.

\subsection{Spectral Bias and Frequency Learning Behavior}\label{bias}

\noindent
A model has \emph{spectral bias} if low frequencies are learned before high ones~\cite{rahaman2019spectral,xu2019frequency,zhang2021understanding}. In NTK terms, this corresponds to steep eigenvalue decay. Flatter spectra indicate more uniform frequency learning.

\medskip
\noindent\textbf{Reduced spectral bias in KANs.}
Wang~\cite{Wang25} showed via Hessian–eigenvalue analysis that shallow KANs possess a well-conditioned landscape, enabling fast learning of oscillatory modes. Grid extension further widens the representable frequency band. Their experiments show KANs capturing high-frequency signals in a few iterations, unlike MLPs.

\noindent
Farea~\cite{Farea25_BasisComp} confirmed this in PDE settings: KANs and learnable-basis PINNs (e.g., Fourier/Jacobi) maintain flatter NTK eigenvalue spectra, indicating weaker spectral bias~\cite{tancik2020fourier,hong2022activation}. However, they also showed that broader spectra can increase curvature, worsen conditioning, and destabilize training~\cite{foret2020sharpness}. Wang~\cite{Wang25} similarly noted that overly fine spline grids may amplify noise sensitivity, whereas modest coarsening can improve generalization.

\smallskip
\noindent
Together, these results show that KANs learn high-frequency features more uniformly~\cite{Wang25,Farea25_BasisComp}, but at the cost of sharper loss landscapes—consistent with kernel-theoretic ill-conditioning phenomena~\cite{Schaback95,Schaback23}. Basis choice must therefore balance spectral reach (e.g., Fourier-like) and stability (e.g., splines).

\subsection{NTK-Based Convergence and Physics-Informed Extensions}\label{ntk}

\noindent\textbf{Gradient-flow convergence.}
Gao~\cite{Gao25} analyzed two-layer KANs under gradient flow, showing that the residual $s(t)$ evolves via
\[
\frac{d}{dt}s(t)=-G\,s(t),
\]
with $G=D^\top D$ the NTK Gram matrix. If $G\succ 0$, the loss decays exponentially:
\[
\mathcal{L}(t)\le\Big(1-\eta\tfrac{\lambda_{\min}}{2}\Big)^{t}\mathcal{L}(0),
\]
giving a rigorous rate under appropriate learning rates. The same holds in expectation for SGD.
Intuitively, a better-conditioned NTK (i.e., a larger and more balanced spectrum of eigenvalues) implies that gradients propagate more uniformly across training directions, leading to faster and more stable convergence in practice.

\medskip
\noindent\textbf{Physics-informed settings.}
For PDE residual losses,
\[
\mathcal{L}_{\mathrm{PDE}}(t)\;\le\;\Big(1-\eta\,\tfrac{\lambda_{\min}}{2}\Big)^{t}\mathcal{L}_{\mathrm{PDE}}(0),
\]
but $\lambda_{\min}$ depends strongly on operator stiffness~\cite{Gao25,Faroughi25}.  
Faroughi~\cite{Faroughi25} showed that Chebyshev-based cKAN/cPIKAN models maintain better NTK conditioning
and converge faster on diffusion, Helmholtz, Allen--Cahn, and beam PDEs. They also demonstrated that
temporal decomposition and hybrid Adam--BFGS schedules improve conditioning, aligning with multi-stage
results in~\cite{Mostajeran25,Mostajeran24}. 
\emph{Initialization can also influence NTK conditioning}: Rigas et al.~\cite{Rigas25_init}\footnote{\url{https://github.com/srigas/KAN_Initialization_Schemes}} 
show that variance-preserving Glorot-style and power-law schemes produce more stable initial NTKs for spline-KANs,
leading to faster convergence.

\subsection{Practical Trade-offs and Convergence Guarantees}\label{practical}

\noindent
KAN convergence is tightly controlled by the conditioning of the NTK Gram matrix:
structured bases often enlarge the smallest eigenvalue $\lambda_{\min}$ relative to
MLPs~\cite{Gao25,Wang25}, accelerating early optimization. Yet the same expressivity
that broadens spectral coverage also increases curvature in the loss landscape,
making training sensitive to basis resolution and learning-rate choices
\cite{Farea25_BasisComp}. Effective practice therefore requires moderating basis
richness to preserve numerical stability.

Physics-informed models highlight this balance.  
Chebyshev-KAN and cPIKAN architectures benefit from well-spread NTK spectra,
producing faster PDE residual decay and more stable training
\cite{Faroughi25,Mostajeran25,Mostajeran24}. But these gains diminish when grids are
refined too aggressively or dimensionality grows, where operators become both costly
to evaluate and harder to condition~\cite{Pal25}. The result is a gradual trade-off
between spectral reach and computational/numerical stability.

Generalization theory reaches the same conclusion from a different angle.  
Zhang and Zhou~\cite{Zhang25_general} show that KAN complexity is governed by
$\ell_1$ coefficient norms and per-layer Lipschitz constants—precisely the terms that
inflate when bases become overly sharp or high-frequency. Li et al.~\cite{Li25_lipschitz}
formalize this perspective through \emph{Lipschitz complexity}, deriving
complexity-dependent generalization bounds and introducing LipKANs, which reduce
Lipschitz growth via Lipschitz-controlled layers and $L^{1.5}$ regularization. Their
results empirically demonstrate that controlling Lipschitz complexity mirrors the
conditioning behavior seen in NTK dynamics: sharper bases yield higher curvature and
poorer generalization.

Taken together, NTK convergence guarantees~\cite{Gao25}, approximation-rate theory
\cite{Wang25,Zhang25_general}, Lipschitz-based generalization analyses
\cite{Li25_lipschitz}, and empirical scaling laws~\cite{Liu24} support a unified
conclusion: KANs are expressive and parameter-efficient, but their practical
performance depends on managing the expressivity–stability trade-off. The most
effective models achieve wide spectral coverage while keeping curvature, conditioning,
and coefficient growth under control~\cite{Wang25,Farea25_BasisComp,Faroughi25}.


\section{Practical ``Choose--Your--KAN'' Guide}\label{sec:choose-kan}
\medskip
\noindent\textbf{When to prefer KANs over MLPs.}
It is important to note that highly optimized MLP implementations remain the preferred choice in many high-throughput production settings where inference latency and hardware efficiency are the primary concerns.
KANs are most advantageous when the learning problem benefits from explicit
control over the underlying function space.
In practice, KANs are preferable to standard MLPs when
(i) data are sparse or expensive and fast convergence is required,
(ii) a shallow architecture is desired for interpretability or stability,
or (iii) prior knowledge about the solution structure is available.
Examples include globally smooth or oscillatory functions (Chebyshev/Fourier
bases), localized or multi-scale features (splines or wavelets), and
nonsmooth or discontinuous behavior (Sinc- or rational-based KANs).
In contrast, when large amounts of data are available and no structural prior
is known, generic MLPs may remain competitive due to their simplicity and
hardware efficiency.

\noindent
The correct configuration of KANs depends on  
(i) the structure of the target function (smooth, oscillatory, discontinuous, periodic, multi-scale),  
(ii) computational constraints (time, memory, precision), and  
(iii) desired properties (accuracy, interpretability).  
A clear seven-step recipe addressing these aspects is given below and is also
presented in algorithmic form as Algorithm~\ref{alg:choose-kan}, synthesizing
Sections~\ref{Basis}--\ref{convergence}.

\medskip
\noindent\textbf{Step 1 — Begin with a stable default.}  
For general regression/classification, start with a \textbf{cubic B\hbox{-}spline KAN} on a uniform grid.  
This setup is mature, stable, and well-conditioned~\cite{Liu24,Liu24b}.  
For PDE or boundary-sensitive tasks, enable \emph{grid extension} and, if needed, apply \emph{post-training refinement}  
(Sec.~\ref{spline}; Fig.~\ref{Spline_Basis}; \cite{Liu24b}).

\medskip
\noindent\textbf{Step 2 — Select a basis, matched to the problem structure.}  
Choose the basis family whose functional properties align with the anticipated solution:
\begin{itemize}
  \item \textbf{Globally smooth / oscillatory:}  
  Chebyshev (ChebyKAN) with per-layer \(\tanh\) normalization~\cite{SS24,ChebyKAN,Yu24,Faroughi25}.
  \item \textbf{Controllable smoothness or boundary shaping:}  
  Jacobi (fKAN, Jacobi-KAN) with fractional/domain parameters~\cite{Aghaei24_fkan,Kashefi25}.
  \item \textbf{Discontinuities, fronts, boundary layers:}  
  SincKAN (bandlimited)~\cite{Yu24},  
  rational rKAN~\cite{Aghaei24_rkan},  
  or DKAN (tanh-jump + smooth background)~\cite{Lei25}.
  \item \textbf{Periodic structure:}  
  FourierKAN or KAF (learnable RFF)~\cite{Xu25_fourier,Zhang25}.
  \item \textbf{Local, multi-scale, or bursty features:}  
  Wavelets (Wav-KAN)~\cite{Bozorgasl24,pde_Patra24}.
\end{itemize}

\medskip
\noindent\textbf{Step 3 — If speed matters, replace heavy components.}  
For wall-clock efficiency, modify the basis or architecture:
\begin{itemize}
  \item \textbf{Fast local bases:}  
  ReLU-KAN/HRKAN (\(\mathrm{ReLU}^m\))~\cite{Qiu24,KAN_pde_So24};  
  or Gaussian RBF variants (FastKAN)~\cite{Li24}.
  \item \textbf{Parallelism and compilers:}  
  JAX \texttt{jit}/XLA, fused CUDA kernels, multi-branch or domain-decomposition architectures~\cite{pde_Rigas24,Daryakenari25,pde_fbkan_Howard24}.
  \item \textbf{Parameter economy:}  
  Prefer orthogonal polynomials or RBFs; prune using \(\ell_1+\)entropy penalties~\cite{KAN_pde_Shukla24,Guo25,Liu24}.
\end{itemize}

\medskip
\noindent\textbf{Step 4 — Stabilize via normalization, grid design, and Lipschitz control.}  
Choose conditioning mechanisms suited to the basis, and reduce Lipschitz complexity where possible:
\begin{itemize}
  \item Chebyshev/Jacobi → apply \(\tanh\) normalization to prevent blow-up of high-order modes~\cite{Yu24,Daryakenari25}.  
  \item RBF/Wavelet → use min–max or sigmoid/tanh scaling to keep oscillatory activations bounded~\cite{Buhler25_regression,Calderon64}.  
  \item Splines → employ extended grids or free-knot refinement near sharp gradients~\cite{Actor25,pde_Rigas24}.  
  \item \textbf{All bases}  optionally insert \(1\)-Lipschitz “Lip’’ layers (tanh/sigmoid/ReLU) between \(\Sigma\) and linear maps to reduce Lipschitz complexity and improve generalization~\cite{Li25_lipschitz}.
\end{itemize}

\medskip
\noindent\textbf{Step 5 — Add physics and adaptive sampling only when needed.}  
For PINNs/PIKANs, form
\[
\mathcal{L}=w_u\mathcal{L}_{\mathrm{data}}+w_f\mathcal{L}_{\mathrm{phys}}+w_b\mathcal{L}_{\mathrm{bc}},
\]
and enhance only when necessary:
\begin{itemize}
  \item Augmented Lagrangian or residual-based attention for conflicting constraints~\cite{Raissi19,pde_Zhang24,Toscano24_kkan}.
  \item Residual-adaptive sampling (RAD) for error concentration~\cite{pde_Rigas24}.  
  \item Coarse\(\to\)fine grid updates as residuals localize~\cite{Actor25}.
\end{itemize}

\medskip
\noindent\textbf{Step 6 — Use the standard optimizer progression.}  
Warm up with Adam/RAdam, then refine with (L\!-\!)BFGS.  
Activate mixed precision and gradient clipping for stiff PDEs~\cite{Daryakenari25,Mostajeran25,KAN_pde_Zeng24,Faroughi25}.

\medskip
\noindent\textbf{Step 7 — Increase capacity \emph{targetedly}, not uniformly.}  
KAN error typically follows \(\ell\propto P^{-\alpha}\), \(\alpha>0\)  
(Sec.~\ref{convergence}; \cite{Liu24,Wang25}).  
The most effective scaling strategies are:
\begin{itemize}
  \item Local refinement (free knots)~\cite{Actor25},
  \item Subdomain PoU decomposition (FBKAN)~\cite{pde_fbkan_Howard24},
  \item Selective basis enrichment (not uniform widening).
\end{itemize}

\medskip
\noindent
This seven-step procedure consolidates the full review into a single, consistent decision pipeline, with one basis-selection step and all relevant citations preserved.

\begin{algorithm}[t]
\caption{Decision Algorithm for Choosing Between MLP and KAN and Configuring KANs}
\label{alg:choose-kan}
\small
\begin{algorithmic}[1]

\State \textbf{Input:} task type, data scale, solution structure, constraints (speed, accuracy, physics)
\State \textbf{Output:} model class and recommended KAN configuration

\medskip
\Statex \textbf{Step 0: Identify task and constraints}
\State Determine task: regression, PDE (PINN/PIKAN), operator learning, or classification
\State Assess data scale (few vs.\ many samples) and structural priors
\State Specify priorities: speed vs.\ accuracy, interpretability, physics constraints

\medskip
\Statex \textbf{Step 1: Choose MLP or KAN}
\If{data are large \textbf{and} no structural prior is available}
    \State Use a standard MLP baseline
\Else
    \State Select a KAN architecture
\EndIf

\medskip
\Statex \textbf{Step 2: Select KAN basis by solution characteristics}
\If{solution is smooth or boundary-sensitive}
    \State Use B-spline KAN (cubic, extended grids)
\ElsIf{solution is periodic or globally oscillatory}
    \State Use FourierKAN or KAF (RFF-based)
\ElsIf{solution is polynomial-like or spectrally smooth}
    \State Use Chebyshev or Jacobi KAN with tanh normalization
\ElsIf{solution has shocks, discontinuities, or boundary layers}
    \State Use SincKAN, rKAN, DKAN, or HRKAN
\ElsIf{solution is multiscale or spatially heterogeneous}
    \State Use Wavelet KAN or FBKAN (PoU decomposition)
\ElsIf{problem is high-dimensional or separable}
    \State Use KKAN, SPIKAN, or KAF (RFF)
\EndIf

\medskip
\Statex \textbf{Step 3: Configure grid and basis resolution}
\State Use coarse grids initially; refine adaptively (multilevel or free knots)
\State Apply residual-adaptive sampling if residuals localize
\State Tune RBF width, wavelet scale, or Sinc step sizes as needed

\medskip
\Statex \textbf{Step 4: Add physics-informed constraints (if applicable)}
\If{physics constraints or PDE residuals are present}
    \State Construct PINN/PIKAN loss with data, PDE, and BC terms
    \State Apply residual reweighting, entropy viscosity, or augmented Lagrangian
\EndIf

\medskip
\Statex \textbf{Step 5: Choose optimization strategy}
\State Warm up with Adam/RAdam
\State Refine with (L-)BFGS or self-scaled quasi-Newton methods
\State Use variance-preserving initialization and moderate regularization

\medskip
\Statex \textbf{Step 6: Optimize for speed or accuracy}
\If{speed is the primary objective}
    \State Prefer ReLU-KAN, HRKAN, FastKAN, LeanKAN
    \State Enable JAX/CUDA kernels and domain decomposition
\Else
    \State Prefer B-spline, Chebyshev/Jacobi, Wavelet KANs
    \State Use adaptive grids, physics losses, and staged optimization
\EndIf

\medskip
\Statex \textbf{Step 7: Final refinement}
\State Apply pruning and sparsity regularization
\State Increase capacity locally rather than uniformly
\State Validate convergence and generalization

\end{algorithmic}
\end{algorithm}

\section{Current Gaps and Path Forward}\label{future}

\noindent
Despite rapid growth in publications and open-source implementations (Tables~\ref{KAN_vs_MLP}, \ref{githubs}), the KAN ecosystem remains fragmented.  
Most studies show strong empirical potential in regression, operator learning, and PDE solving, yet comparisons are inconsistent and rarely grounded in theory.  
The field now requires principled foundations rather than isolated “KAN vs.\ MLP’’ or “basis-vs.-basis’’ experiments.

\subsection*{Limits of Current Comparative Frameworks}

\paragraph{(1) KAN vs.\ MLP benchmarks.}
These comparisons often match parameter counts but ignore essential differences: functional roles of parameters, basis-induced inductive bias, and normalization/grid constraints.  
Outcomes range from clear KAN advantages to near parity~\cite{Yu24_fairer, KAN_pde_Shukla24, Yang24_Comp}, offering little predictive insight into when KANs should win.

\paragraph{(2) Basis-vs.-basis races.}
Comparisons of splines, Chebyshev, Gaussians, etc.~\cite{Farea25_BasisComp} highlight problem dependence but mask deeper structure.  
Each basis involves choices of normalization, grids, and regularization—too many degrees of freedom for a single benchmark to be conclusive.
For fair comparisons, future studies should standardize key variables such as grid resolution, normalization schemes, training settings, and evaluation metrics.

\noindent
Overall, KANs are not a single architecture but a \emph{framework} whose performance depends on principled design choices rooted in approximation theory, numerical conditioning, and optimization geometry.

\paragraph{Case Study: The Many Faces of a Chebyshev KAN:}

A plain Chebyshev KAN can be unstable, but targeted refinements substantially improve behavior:

\begin{itemize}
    \item \textbf{Domain stabilization}: \texttt{tanh} normalization confines inputs to $[-1,1]$ for numerically stable Chebyshev evaluation (Section~\ref{Cheby}).
    \item \textbf{Architectural hybrids}: Linear heads reduce overfitting in inverse/PDE tasks~\cite{Daryakenari25}.
    \item \textbf{Grid-informed adaptivity}: Learnable \texttt{tanh}-based grids add local adaptivity~\cite{Toscano24_kkan}.
    \item \textbf{Principled optimization}: Hybrid Adam--L-BFGS schedules improve spectral conditioning~\cite{Faroughi25, Mostajeran25}.
    \item \textbf{Adaptive polynomial degree}: Making the polynomial degree itself trainable, as in Adaptive PolyKAN~\cite{Attouri25}.

\end{itemize}

\noindent
Similar refinements benefit other bases: adaptive centers for Gaussian KANs~\cite{Abueidda25}, free-knot updates for spline KANs~\cite{Liu24b, Actor25}, and domain decomposition for finite-basis KANs~\cite{pde_fbkan_Howard24}.

\subsection*{A Multi-Pillar Framework for Future KAN Research}

\paragraph{Pillar~1: Robust component library.}
Canonical, stable implementations for Chebyshev, spline, Gaussian/RBF, Fourier, and wavelet bases with standardized normalization and initialization.

\paragraph{Pillar~2: Basis selection theory.}
Principles linking problem structure to basis choice, including spectral alignment, NTK conditioning, and approximation–regularization trade-offs.

\paragraph{Pillar~3: Finite-width optimization theory.}
Beyond NTK predictions~\cite{Gao25, Faroughi25}; characterize finite-width loss geometry, adaptive basis movement, and optimization dynamics.  
Include Extreme Learning Machine–style regimes~\cite{Han06} as an alternative training strategy for KANs.

\paragraph{Pillar~4: Generalization and regularization.}
Joint treatment of parameter count, basis complexity, smoothness, and explicit $\ell_1/\ell_2$ penalties for improved generalization bounds.

\paragraph{Pillar~5: Composition and hierarchy.}
Depth in KANs currently has no formal interpretation.  
Open problems include understanding the functional role of layer composition, the
interactions among heterogeneous bases, and the trade-off between expressivity and
stability under deep compositions (see also the theoretical results of
\cite{Wang25,Zhang25_general} and the empirical observations of
\cite{Pant25}\footnote{\url{https://github.com/geoelements-dev/mlp-kan}}).

\paragraph{Pillar~6: Interpretability and identifiability.}
Uniqueness of learned univariate functions, robustness, and symbolic identifiability.

\paragraph{Pillar~7: RKHS-based formulation.}
Develop RKHS-guided KAN theory using recent formulations~\cite{Zhang25_general} and earlier functional-analytic perspectives on KST~\cite{Braun09a}.

\paragraph{Pillar~8: New basis functions.}
Introduce and test new basis families in KANs, including Newton basis functions~\cite{Muller09, Santin16}, low-order Wendland functions, and Matérn/Sobolev kernels~\cite{Wendland04}.

\paragraph*{Outlook}
A unified framework combining stable components, principled basis selection, finite-width theory, RKHS insights, and exploration of new bases can guide future KAN development.

\section{Conclusion}\label{Conclude}

\noindent
This review has presented a systematic and unified perspective on
Kolmogorov--Arnold Networks (KANs), bridging their theoretical foundations,
architectural principles, and rapidly growing body of practical applications.
By synthesizing insights from the literature together with our own analysis,
we identify several key conclusions that characterize the current state of
KAN research and point toward promising future directions.

\paragraph{Theoretical perspective.}
\noindent
Modern KANs are best understood as architectures \emph{inspired by}, rather
than direct realizations of, the Kolmogorov Superposition Theorem (KST).
The KST provides a conceptual blueprint: its central insight—that any
continuous multivariate function admits a representation as sums of univariate
functions—motivates the characteristic \emph{activate--then--sum} structure
of KANs.
However, contemporary KAN architectures depart fundamentally from the classical
theorem by replacing non-smooth, universal inner functions with smooth,
learnable basis functions such as splines or polynomials.
This departure sacrifices a direct constructive link to KST, but enables
efficient, stable, and data-driven approximation in practical settings.

\paragraph{Relationship to classical and modern methods.}
\noindent
By positioning KANs relative to kernel methods and multilayer perceptrons
(MLPs), this review clarifies their functional role and architectural
advantages.
We showed that shallow, one-dimensional KANs are mathematically equivalent to
classical kernel or basis-expansion methods, differing primarily in their
training procedure.
In contrast, higher-dimensional or deep KANs diverge structurally from kernel
methods by constructing multivariate interactions through additive,
compositional layers rather than explicit tensor-product or radial bases,
thereby mitigating the curse of dimensionality.

\noindent
The introduction of depth further distinguishes KANs from linear
basis-expansion models.
Hidden layers induce both a multiplicative growth in effective polynomial
degree and a nonlinear coupling of the final function coefficients.
This coefficient entanglement underlies the expressive power of deep KANs, but
also explains the optimization challenges observed in practice, including
non-convex loss landscapes, parameter redundancy, and sensitivity to
initialization.

\noindent
From the perspective of neural networks, KANs and MLPs are expressively
equivalent, but differ markedly in how nonlinearity is deployed.
KANs relocate nonlinearity from fixed node-wise activations to learnable
edge-wise functions, replacing the MLP paradigm of \emph{sum--then--activate}
with an \emph{activate--then--sum} architecture.
This structural shift provides a strong inductive bias, improves parameter
efficiency, and enhances interpretability through explicit functional
components.

\paragraph{Practical design and applications.}
\noindent
A central lesson emerging from the expanding KAN ecosystem is that performance
is not intrinsic to the ``KAN'' label itself, but instead arises from informed
design choices.
Among these, basis selection is the dominant design axis.
B-splines naturally encode locality, Chebyshev polynomials offer spectral
accuracy for smooth problems, Fourier bases are well suited for periodic
structures, and wavelets enable multi-scale representations.
No single basis is universally optimal; effective KAN design must align the
chosen basis with the structure of the target function.

\noindent
Beyond basis choice, successful KAN applications consistently employ a rich
toolkit of accuracy and efficiency enhancements.
These include physics-informed loss formulations, adaptive sampling and grid
strategies, domain and function decomposition, and problem-specific
regularization techniques.
Such methods, often inherited from the broader scientific machine learning
literature, are essential for achieving robust and competitive performance.

\paragraph{Outlook.}
\noindent
In summary, Kolmogorov--Arnold Networks represent a shift away from monolithic,
fixed-activation models toward structured, modular, and interpretable
architectures.
Their strength lies not in a single rigid formulation, but in the flexibility
to encode prior knowledge directly into the functional building blocks of the
network.
Future progress will depend on the development of a more unified theoretical
framework that connects basis selection, optimization dynamics, conditioning,
and generalization.
Establishing such principles will be key to fully realizing the potential of
KANs as a general-purpose tool for scientific computing and data-driven
modeling.

\section{Acknowledgments}
This work was supported by the General Research Fund (GRF No. 12301824, 12300922) of Hong Kong Research Grant Council.


\begin{thebibliography}{99}



\bibitem{Liu24}
Z.~Liu, Y.~Wang, S.~Vaidya, F.~Ruehle, J.~Halverson, M.~Solja\v{c}i\'c, T.~Y.~Hou, and M.~Tegmark,
``KAN: Kolmogorov--Arnold Networks,''
in \emph{Proceedings of the International Conference on Learning Representations (ICLR)},
2025.

\bibitem{Liu24b}
Liu, Z., M. Tegmark, P. Ma, W. Matusik, and Y. Wang, Kolmogorov-Arnold Networks Meet Science. Physical Review X, 2025. 15(4): p. 041051. Available: \url{https://github.com/KindXiaoming/pykan}


\bibitem{Howard26}
 A.A. Howard, N. Zolman, B. Jacob, S.L. Brunton and P. Stinism SINDy-KANs: Sparse identification of non-linear dynamics through Kolmogorov-Arnold networks. \textit{arXiv preprint} arXiv:2603.18548, 2026.

\bibitem{Faroughi26_}
S.A. Faroughi,  F. Mostajeran, A. Arzani, and S. Faroughi, Symbolic--KAN: Kolmogorov-Arnold Networks with Discrete Symbolic Structure for Interpretable Learning. \textit{arXiv preprint} arXiv:2603.23854, 2026.


\bibitem{Raissi19}
M. Raissi, P. Perdikaris, and G. E. Karniadakis, 
``Physics-informed neural networks: A deep learning framework for solving forward and inverse problems involving nonlinear partial differential equations,'' 
\textit{Journal of Computational Physics}, vol. 378, pp. 686--707, 2019.



\bibitem{pang2019fpinns}
G. Pang, L. Lu, and G. E. Karniadakis, 
``fPINNs: Fractional physics-informed neural networks,'' 
\textit{SIAM Journal on Scientific Computing}, vol. 41, no. 4, pp. A2603--A2626, 2019.

\bibitem{zhang2019quantifying}
D. Zhang, L. Lu, L. Guo, and G. E. Karniadakis, 
``Quantifying total uncertainty in physics-informed neural networks for solving forward and inverse stochastic problems,'' 
\textit{Journal of Computational Physics}, vol. 397, p. 108850, 2019.


\bibitem{kharazmi2019}
E. Kharazmi, Z. Zhang, and G. E. Karniadakis,  
``Variational physics-informed neural networks for solving partial differential equations,''  
\textit{arXiv preprint} arXiv:1912.00873, 2019.


\bibitem{jagtap2020extended}
A. D. Jagtap and G. E. Karniadakis,  
``Extended physics-informed neural networks (XPINNs): A generalized space–time domain decomposition based deep learning framework for nonlinear partial differential equations,''  
\textit{Communications in Computational Physics}, vol. 28, no. 5, 2020.


\bibitem{Meng20}
X. Meng and G. E. Karniadakis,  
``A composite neural network that learns from multi-fidelity data: Application to function approximation and inverse PDE problems,''  
\textit{Journal of Computational Physics}, vol. 401, p. 109020, 2020.



\bibitem{shin2020convergence}
Y. Shin, J. Darbon, and G. E. Karniadakis,  
``On the convergence of physics-informed neural networks for linear second-order elliptic and parabolic type PDEs,''  
\textit{arXiv preprint} arXiv:2004.01806, 2020.


\bibitem{Wang22}
S. Wang, P. Yu, and P. Perdikaris,  
``When and why PINNs fail to train: A neural tangent kernel perspective,''  
\textit{Journal of Computational Physics}, vol. 449, p. 110768, 2022.



\bibitem{lu2021deepxde}
L. Lu, X. Meng, Z. Mao, and G. E. Karniadakis,  
``DeepXDE: A deep learning library for solving differential equations,''  
\textit{SIAM Review}, vol. 63, no. 1, pp. 208--228, 2021.


\bibitem{wang2023multi}
Y. Wang and C.-Y. Lai,  
``Multi-stage neural networks: Function approximator of machine precision,''  
\textit{arXiv preprint} arXiv:2307.08934, 2023.



\bibitem{mcclenny2020self}
L. McClenny and U. Braga-Neto,  
``Self-adaptive physics-informed neural networks using a soft attention mechanism,''  
\textit{arXiv preprint} arXiv:2009.04544, 2020.


\bibitem{Shukla21}
K. Shukla, A. D. Jagtap, and G. E. Karniadakis,  
``Parallel physics-informed neural networks via domain decomposition,''  
\textit{Journal of Computational Physics}, vol. 447, p. 110683, 2021.
	



\bibitem{hu2024tackling}
Z. Hu, K. Shukla, G. E. Karniadakis, and K. Kawaguchi,  
``Tackling the curse of dimensionality with physics-informed neural networks,''  
\textit{Neural Networks}, vol. 176, p. 106369, 2024.



\bibitem{Sifan25}
S. Wang, S. Sankaran, P. Stinis, and P. Perdikaris,  
``Simulating three-dimensional turbulence with physics-informed neural networks,''  
\textit{arXiv preprint} arXiv:2507.08972, 2025.


\bibitem{Sifan24Pirate}
S. Wang, B. Li, Y. Chen, and P. Perdikaris,  
``Piratenets: Physics-informed deep learning with residual adaptive networks,''  
\textit{Journal of Machine Learning Research}, vol. 25, no. 402, pp. 1--51, 2024.


\bibitem{Rossi05}
F. Rossi and B. Conan-Guez,  
``Functional multi-layer perceptron: A non-linear tool for functional data analysis,''  
\textit{Neural Networks}, vol. 18, no. 1, pp. 45--60, 2005.
\bibitem{Cranmer23}
M. Cranmer,  
``Interpretable machine learning for science with PySR and SymbolicRegression.jl,''  
\textit{arXiv preprint} arXiv:2305.01582, 2023.




\bibitem{Cunningham23}
H. Cunningham, A. Ewart, L. Riggs, R. Huben, and L. Sharkey,  
``Sparse autoencoders find highly interpretable features in language models,''  
\textit{arXiv preprint} arXiv:2309.08600, 2023.




\bibitem{Amir24}
A. Noorizadegan, R. Cavoretto, D. L. Young, and C. S. Chen,  
``Stable weight updating: A key to reliable PDE solutions using deep learning,''  
\textit{Engineering Analysis with Boundary Elements}, vol. 168, p. 105933, 2024.

\bibitem{Amir26_reg}
A. Noorizadegan, Y. C. Hon, D.-L. Young, and C.-S. Chen,  
``Enhancing supervised surface reconstruction through implicit weight regularization,''  
\textit{Engineering Analysis with Boundary Elements}, vol. 180, p. 106439, 2025.  
doi: \url{https://doi.org/10.1016/j.enganabound.2025.106439}.

\bibitem{Amir24a}
A. Noorizadegan, D. L. Young, Y. C. Hon, and C. S. Chen,  
``Power-enhanced residual network for function approximation and physics-informed inverse problems,''  
\textit{Applied Mathematics and Computation}, vol. 480, p. 128910, 2024.





\bibitem{Sifan25a}
S. Wang, A. K. Bhartari, B. Li, and P. Perdikaris,  
``Gradient alignment in physics-informed neural networks: A second-order optimization perspective,''  
\textit{arXiv preprint} arXiv:2502.00604, 2025.



\bibitem{Sifan21}
S. Wang, Y. Teng, and P. Perdikaris,  
``Understanding and mitigating gradient flow pathologies in physics-informed neural networks,''  
\textit{SIAM Journal on Scientific Computing}, vol. 43, no. 5, pp. A3055--A3081, 2021.




\bibitem{Rahaman19}
N. Rahaman, A. Baratin, D. Arpit, F. Draxler, M. Lin, F. Hamprecht, Y. Bengio, and A. Courville,  
``On the spectral bias of neural networks,''  
in \textit{Proceedings of the 36th International Conference on Machine Learning (ICML)}, vol. 97, Proceedings of Machine Learning Research (PMLR), pp. 5301--5310, 2019.



\bibitem{Xu20}
Z. Q. J. Xu, Y. Zhang, T. Luo, Y. Xiao, and Z. Ma,  
``Frequency principle: Fourier analysis sheds light on deep neural networks,''  
\textit{Communications in Computational Physics}, vol. 28, no. 5, pp. 1746--1767, 2020.



\bibitem{Cai19}
W. Cai and Z. Q. J. Xu,  
``Multi-scale deep neural networks for solving high-dimensional PDEs,''  
\textit{arXiv preprint} arXiv:1910.11710, 2019.





\bibitem{Liu25_kanl}
Z. Liu, M. Tegmark, P. Ma, W. Matusik, and Y. Wang,  
``Kolmogorov--Arnold Networks Meet Science,''  
\textit{Physical Review X}, vol.~15, no.~4, Art.~041051, 2025.


\bibitem{Andrade25}
M. Andrade, L. Freitas, and J. Beatrize,  
``Kolmogorov–Arnold Networks for interpretable and efficient function approximation,''  
\textit{Preprints}, 2025.  

\bibitem{Basina24_interp_review}
D. Basina, J.R. Vishal, A. Choudhary, and B. Chakravarthi,  
``KAT to KANs: A review of Kolmogorov–Arnold Networks and the neural leap forward,''  
\textit{arXiv preprint} arXiv:2411.10622, 2024.  

\bibitem{Beatrize25}
J. Beatrize,  
``Scalable and interpretable function-based architectures: A survey of Kolmogorov–Arnold Networks,''  
\textit{engrXiv preprint}, doi:10.31224/4515, 2025. [Online]. Available: \url{https://engrxiv.org/preprint/view/4515/version/6148}  

\red{
\bibitem{Faroughi25_review}
S.A. Faroughi, F. Mostajeran, A. Hamed Mashhadzadeh, and S. Faroughi, Kolmogorov-Arnold networks for data-driven, physics-informed, and deep-operator learning: a review, synthesis, and new analysis. Neural Networks, 200: p. 108791, 2026.}

\bibitem{Kilani25}
B.H. Kilani,  
``Convolutional Kolmogorov–Arnold Networks: A survey,''  
\textit{HAL preprint} hal-05177765, 2025.  

\bibitem{Ji24}
Y. Hou, T. Ji, and D. Zhang,  A. Stefanidis
``Kolmogorov-Arnold Networks: A Critical Assessment of Claims, Performance, and Practical Viability,''  
\textit{arXiv preprint} arXiv:2407.11075, 2024.  

\bibitem{Dutta25_review}
A. Dutta, B. Maheswari, N. Punitha, \textit{et al.},  
``The first two months of Kolmogorov-Arnold Networks (KANs): A survey of the state-of-the-art,''  
\textit{Arch. Computat. Methods Eng.}, 2025. doi: 10.1007/s11831-025-10328-2

\bibitem{Essahraui25}
S. Essahraui, I. Lamaakal, K. E. Makkaoui, I. Ouahbi, M. F. Bouami, and Y. Maleh,  
``Kolmogorov-Arnold Networks: Overview of Architectures and Use Cases,''  
\textit{Proc. Int. Conf. Circuit, Systems and Communication (ICCSC)}, Fez, Morocco, 2025, pp. 1--6. doi: 10.1109/ICCSC66714.2025.11135248

\bibitem{Somvanshi25}
S. Somvanshi, S.A. Javed, M.M. Islam, D. Pandit, and S. Das,  
``A survey on Kolmogorov–Arnold Networks,''  
\textit{ACM Computing Surveys}, 2025.



\bibitem{Yu24_fairer}
R. Yu, W. Yu, and X. Wang,  
``KAN or MLP: A fairer comparison,''  
\textit{arXiv preprint} arXiv:2407.16674, 2024. [Online]. Available: \url{https://github.com/yu-rp/KANbeFair}


\bibitem{KAN_pde_Shukla24}
K. Shukla, J. D. Toscano, Z. Wang, Z. Zou, and G. E. Karniadakis,  
``A comprehensive and fair comparison between MLP and KAN representations for differential equations and operator networks,''  
\textit{Computer Methods in Applied Mechanics and Engineering}, vol. 431, 2024.


\bibitem{Yang24_Comp}
Z. Yang, J. Zhang, X. Luo, Z. Lu, and L. Shen,  
``Activation Space Selectable Kolmogorov-Arnold Networks,''  
\textit{arXiv preprint} arXiv:2408.08338, 2024.


\bibitem{Farea25_BasisComp}
A. Farea and M. S. Celebi,  
``Learnable activation functions in physics-informed neural networks for solving partial differential equations,''  
\textit{Computer Physics Communications}, vol. 315, p. 109753, 2025.  
GitHub: \url{https://github.com/afrah/pinn_learnable_activation}




\bibitem{Ta24}
H.-T. Ta,  
``BSRBF-KAN: A combination of B-splines and radial basis functions in Kolmogorov-Arnold networks,''  
\textit{arXiv preprint} arXiv:2406.11173, 2024. [Online]. Available: \url{https://github.com/hoangthangta/BSRBF_KAN}

\bibitem{He24}
Y. He, Y. Xie, Z. Yuan, and L. Sun,  
``MLP-KAN: Unifying deep representation and function learning,''  
\textit{arXiv preprint} arXiv:2410.03027, 2024.

\bibitem{Pal25}
A. Pal and D. Das,  
``Understanding the limitations of B-spline KANs: Convergence dynamics and computational efficiency,''  
in \textit{NeurIPS 2024 Workshop on Scientific Methods for Understanding Deep Learning}, 2024.

\bibitem{Li25_DEKAN}
D. Li, B. Yan, Q. Long, and B. Wang,  
``DE-KAN: A differential evolution-based optimization framework for enhancing Kolmogorov-Arnold networks in complex nonlinear modeling,''  
in \textit{IEEE Congress on Evolutionary Computation (CEC)}, pp. 1--8, 2025.

\bibitem{pde_Koeing24}
B. C. Koenig, S. Kim, and S. Deng,  
``KAN-ODEs: Kolmogorov-Arnold Network ordinary differential equations for learning dynamical systems and hidden physics,''  
\textit{Computer Methods in Applied Mechanics and Engineering}, vol. 432, p. 117397, 2024.

\bibitem{Mallick25_battery}
S. Mallick, S. Ghosh, and T. Roy,  
``KAN-Therm: A lightweight battery thermal model using Kolmogorov-Arnold Network,''  
\textit{arXiv preprint} arXiv:2509.09145, 2025.



\bibitem{Actor25}
J. A. Actor, G. Harper, B. Southworth, and E. C. Cyr,  
``Leveraging KANs for expedient training of multichannel MLPs via preconditioning and geometric refinement,''  
\textit{arXiv preprint} arXiv:2505.18131, 2025.

\bibitem{KAN_pde_Zeng24}
C. Zeng, J. Wang, H. Shen, and Q. Wang,  
``KAN versus MLP on irregular or noisy functions,''  
\textit{arXiv preprint} arXiv:2408.07906, 2024.

\bibitem{Qiu25}
R. Qiu, Y. Miao, S. Wang, Y. Zhu, L. Yu, and X.-S. Gao,  
``PowerMLP: An efficient version of KAN,''  
\textit{Proceedings of the AAAI Conference on Artificial Intelligence}, vol. 39, no. 19, pp. 20069--20076, 2025. [Online]. Available: \url{https://github.com/Iri-sated/PowerMLP}


\bibitem{Yu24}
T. Yu, J. Qiu, J. Yang, and I. Oseledets,  
``Sinc Kolmogorov-Arnold network and its applications on physics-informed neural networks,''  
\textit{arXiv preprint} arXiv:2410.04096, 2024. [Online]. Available: \url{https://github.com/DUCH714/SincKAN}

\bibitem{Mahmoud25}
A.A. Mahmoud, A. Pester, M.M. Muttardi, F. Andres, S. Tanabe, N. Greneche, and H.H. Ali,  
``Cheby-KANs: Advanced Kolmogorov–Arnold Networks for Applying Geometric Deep Learning in Quantum Chemistry Applications,''  \textit{IEEE Access}, vol. 13, pp. 130525--130534, 2025.


\bibitem{Jiang25_quantum}
J.-C. Jiang, Y.-C. Huang, T. Chen, and H.-S. Goan,  
``Quantum variational activation functions empower Kolmogorov–Arnold networks,''  
\textit{arXiv preprint} arXiv:2509.14026, 2025. [Online]. Available: \url{https://github.com/Jim137/qkan}



\bibitem{Toscano24_kkan}
J. D. Toscano, L.-L. Wang, and G. E. Karniadakis,  
``KKANs: Kurkova-Kolmogorov-Arnold networks and their learning dynamics,''  
\textit{Neural Networks}, vol. 191, p. 107831, 2025.

\bibitem{Wang25}
Y. Wang, J. W. Siegel, Z. Liu, and T. Y. Hou,  
``On the expressiveness and spectral bias of KANs,''  
\textit{arXiv preprint} arXiv:2410.01803, 2024.

\bibitem{Abueidda25}
D. W. Abueidda, P. Pantidis, and M. E. Mobasher,  
``DeepOKAN: Deep operator network based on Kolmogorov-Arnold networks for mechanics problems,''  
\textit{Computer Methods in Applied Mechanics and Engineering}, vol. 436, p. 117699, 2025. 
GitHub: \url{https://github.com/DiabAbu/Dee}

\bibitem{Zhang25_comp}
L. Zhang, L. Chen, F. An, Z. Peng, Y. Yang, T. Peng, Y. Song, and Y. Zhao,  
``A physics-informed neural network for nonlinear deflection prediction of ionic polymer-metal composite based on Kolmogorov-Arnold networks,''  
\textit{Engineering Applications of Artificial Intelligence}, vol. 144, p. 110126, 2025.

\bibitem{Yang25}
S. Yang, K. Lin, and A. Zhou,  
``The KAN-MHA model: A novel physical knowledge based multi-source data-driven adaptive method for airfoil flow field prediction,''  
\textit{Journal of Computational Physics}, vol. 528, p. 113846, 2025.

\bibitem{Guo25}
M.-H. Guo, X. Lü, and Y.-X. Jin,  
``Extraction and reconstruction of variable-coefficient governing equations using Res-KAN integrating sparse regression,''  
\textit{Physica D: Nonlinear Phenomena}, vol. 481, p. 134689, 2025.

\bibitem{Xu25}
Z. Xu and B. Lv,  
``Enhancing physics-informed neural networks with a hybrid parallel Kolmogorov-Arnold and MLP architecture,''  
\textit{arXiv preprint} arXiv:2503.23289, 2025.

\bibitem{Khedr25}
O. Khedr, A. Al-Oufy, A. Saleh, et al.,  
``Physics-informed Kolmogorov-Arnold networks: A superior approach to fluid simulation,''  
\textit{Research Square} (preprint), 2025. doi:10.21203/rs.3.rs-6743344/v1.

\bibitem{Lei25}
G. Lei, D. Exposito, and X. Mao,  
``Discontinuity-aware KAN-based physics-informed neural networks,''  
\textit{arXiv preprint} arXiv:2507.08338, 2025.

\bibitem{Kalesh25}
D. Kalesh, T. Merembayev, S. Omirbekov, and Y. Amanbek,  
``Physics-informed Kolmogorov-Arnold network for two-phase flow model with experimental data,''  
in \textit{ICCSA 2025 Workshops}, LNCS 15888, Springer, 2026.


\bibitem{Mostajeran24}
F. Mostajeran and S. A. Faroughi,  
``EPi-cKANs: Elasto-plasticity informed Kolmogorov–Arnold networks using Chebyshev polynomials,''  
\textit{arXiv preprint} arXiv:2410.10897, 2024.

\bibitem{Mostajeran25}
F. Mostajeran and S. A. Faroughi,  
``Scaled-cPIKANs: Domain scaling in Chebyshev-based physics-informed Kolmogorov–Arnold networks,''  
\textit{arXiv preprint} arXiv:2501.02762, 2025.

\bibitem{Faroughi25}
S. A. Faroughi and F. Mostajeran,  
``Neural tangent kernel analysis to probe convergence in physics-informed neural solvers: PIKANs vs. PINNs,''  
\textit{arXiv preprint} arXiv:2506.07958, 2025.

\bibitem{Daryakenari25}
N. A. Daryakenari, K. Shukla, and G. E. Karniadakis,  
``Representation meets optimization: Training PINNs and PIKANs for gray-box discovery in systems pharmacology,''  
\textit{arXiv preprint} arXiv:2504.07379, 2025.

\bibitem{pde_Zhang24}
Z. Zhang, Q. Wang, Y. Zhang, et al.,  
``Physics-informed neural networks with hybrid Kolmogorov-Arnold network and augmented Lagrangian function for solving partial differential equations,''  
\textit{Scientific Reports}, vol. 15, p. 10523, 2025.



\bibitem{Aghaei24_kantorol}
A. A. Aghaei,  
``KANtrol: A physics-informed Kolmogorov-Arnold network framework for solving multi-dimensional and fractional optimal control problems,''  
\textit{arXiv preprint} arXiv:2409.06649, 2024.

\bibitem{pde_shuai24}
H. Shuai and F. Li,  
``Physics-informed Kolmogorov-Arnold networks for power system dynamics,''  
\textit{arXiv preprint} arXiv:2408.06650, 2024.

\bibitem{pde_Wang24}
Y. Wang, J. Sun, J. Bai, C. Anitescu, M. S. Eshaghi, X. Zhuang, T. Rabczuk, and Y. Liu,  
``Kolmogorov–Arnold-informed neural network: A physics-informed deep learning framework for solving forward and inverse problems based on Kolmogorov–Arnold networks,''  
\textit{Computer Methods in Applied Mechanics and Engineering}, vol. 433, p. 117518, 2025.

\bibitem{Kashefi25}
A. Kashefi,  T. Mukerji,
``Kolmogorov–Arnold PointNet: Deep learning for prediction of fluid fields on irregular geometries,''  
\textit{arXiv preprint} arXiv:2504.06327, 2025.
 [Online]. Available: \url{https://github.com/Ali-Stanford/Physics_Informed_KAN_PointNet}


\bibitem{Zhang25}
J. Zhang, Y. Fan, K. Cai, and K. Wang,  
``Kolmogorov-Arnold Fourier networks,''  
\textit{arXiv preprint} arXiv:2502.06018, 2025. [Online]. Available: \url{https://github.com/kolmogorovArnoldFourierNetwork/KAF}

\bibitem{Yang25_multiScale}
Y.-S. Yang, L. Guo, and X. Ren,  
``Multi-Resolution Training-Enhanced Kolmogorov–Arnold Networks for Multi-Scale PDE Problems,''  
\textit{arXiv preprint} arXiv:2507.19888, 2025.


\bibitem{Xiong25}
X. Xiong, K. Lu, Z. Zhang, Z. Zeng, S. Zhou, Z. Deng, and R. Hu,  
``J-PIKAN: A physics-informed KAN network based on Jacobi orthogonal polynomials for solving fluid dynamics,''  
\textit{Communications in Nonlinear Science and Numerical Simulation}, vol. 152, p. 109414, 2026.


\bibitem{Zhang26_legendre}
Z. Zhang, X. Xiong, S. Zhang, W. Wang, Y. Zhong, C. Yang, and X. Yang,  
``Legend-KINN: A Legendre polynomial-based Kolmogorov–Arnold-informed neural network for efficient PDE solving,''  
\textit{Expert Systems with Applications}, vol. 298, p. 129839, 2026.  
[Online]. Available: \url{https://github.com/zhang-zhuo001/Legend-KINN}


\bibitem{awesomeKAN}
mintisan,  
``awesome-kan,'' GitHub repository, 2024. [Online]. Available: \url{https://github.com/mintisan/awesome-kan}




\bibitem{Hilbert02}
D.~Hilbert,  
``Mathematical problems,''  
\textit{Bull. Amer. Math. Soc.}, vol.~8, pp.~437--479, 1902.

\bibitem{Kolmogorov56}
A.~N.~Kolmogorov,  
``On the representation of continuous functions of several variables as superpositions of continuous functions of a smaller number of variables,''  
\textit{Dokl. Akad. Nauk SSSR}, vol.~108, no.~2, pp.~179--182, 1956. (In Russian.)

\bibitem{Arnold57}	
V.~I.~Arnol'd,  
``On functions of three variables,''  
\textit{Dokl. Akad. Nauk SSSR}, vol.~114, pp.~679--681, 1957. (In Russian.)

\bibitem{Kolmogorov57}
A. N. Kolmogorov,  
``On the representation of continuous functions of many variables by superposition of continuous functions of one variable and addition,''  
\textit{Doklady Akademii Nauk}, vol. 114, pp. 953--956, 1957.


\bibitem{Kahane75}
J.-P.~Kahane,  
``Sur le théorème de superposition de Kolmogorov,''  
\textit{J. Approx. Theory}, vol.~13, no.~3, pp.~229--234, 1975.

\bibitem{Hedberg71}
T.~Hedberg,  
``The Kolmogorov Superposition Theorem,''  
in \textit{Topics in Approximation Theory},  
Lecture Notes in Mathematics, vol.~187, Springer, 1971, Appendix II.


\bibitem{Lorentz62}
G.~G.~Lorentz,  
``Metric entropy, width, and superpositions of functions,''  
\textit{Amer. Math. Monthly}, vol.~69, pp.~469--485, 1962.

\bibitem{Sprecher65}
D.~A.~Sprecher,  
``On the structure of representations of continuous functions of several variables,''  
\textit{Trans. Amer. Math. Soc.}, vol.~115, pp.~340--355, 1965.


\bibitem{Arnold58}
V.~I.~Arnol'd,  
``The representation of functions of several variables,''  
\textit{Mat. Prosvesch.}, vol.~3, pp.~41--61, 1958. (In Russian.)


\bibitem{Ismailov24}
A.~Ismayilova and V.~E.~Ismailov,  
``On the Kolmogorov neural networks,''  
\textit{Neural Networks}, vol.~176, p.~106333, 2024.

\bibitem{Samadi24_smooth}
M.~E.~Samadi, Y.~Müller, and A.~Schuppert,  
``Smooth Kolmogorov--Arnold networks enabling structural knowledge representation,''  
\textit{arXiv preprint arXiv:2405.11318}, 2024.


\bibitem{Sternfeld79}
Y.~Sternfeld,  
``Superpositions of continuous functions,''  
\textit{J. Approx. Theory}, vol.~25, pp.~360--368, 1979.



\bibitem{Vitushkin54_nonSmooth}
A.~G.~Vitushkin,  
``On Hilbert's thirteenth problem,''  
\textit{Dokl. Akad. Nauk SSSR}, vol.~95, pp.~701--704, 1954.

\bibitem{Vitushkin77_nonSmooth}
A.~G.~Vitushkin,  
``On representation of functions by means of superpositions and related topics,''  
\textit{L’Enseignement Mathématique}, 1977.

\bibitem{Henkin64_nonSmooth}
G.~M.~Henkin,  
``Linear superpositions of continuously differentiable functions,''  
\textit{Dokl. Akad. Nauk SSSR}, vol.~157, pp.~288--290, 1964. (In Russian.)



\bibitem{Girosi89}
F.~Girosi and T.~Poggio,  
``Representation properties of networks: Kolmogorov’s theorem is irrelevant,''  
\textit{Neural Comput.}, vol.~1, pp.~465--469, 1989.

\bibitem{Kurkova91}
V.~Kůrková,  
``Kolmogorov’s theorem is relevant,''  
\textit{Neural Comput.}, vol.~3, pp.~617--622, 1991.

\bibitem{Kurkova92}
V.~Kůrková,  
``Kolmogorov’s theorem and multilayer neural networks,''  
\textit{Neural Networks}, vol.~5, no.~3, pp.~501--506, 1992.



\bibitem{Fridman67}
B.~L.~Fridman,  
``Improvement in the smoothness of functions in the Kolmogorov superposition theorem,''  
\textit{Dokl. Akad. Nauk SSSR}, vol.~177, pp.~1019--1022, 1967.  
English transl.: \textit{Soviet Math. Dokl.}, vol.~8, pp.~1550--1553, 1967.


\bibitem{Sprecher72}
D.~A.~Sprecher,  
``An improvement in the superposition theorem of Kolmogorov,''  
\textit{J. Math. Anal. Appl.}, vol.~38, no.~1, pp.~208--213, 1972.



\bibitem{Actor18}
J.~Actor,  
\textit{Computation for the Kolmogorov Superposition Theorem},  
PhD Dissertation, 2018.


\bibitem{Actor19}
J.~Actor and M.~G.~Knepley,  
``An algorithm for computing Lipschitz inner functions in Kolmogorov's superposition theorem,''  
\textit{arXiv preprint} arXiv:1712.08286, 2017.


\bibitem{Sprecher96}
D.~A.~Sprecher,  
``A numerical implementation of Kolmogorov’s superpositions,''  
\textit{Neural Networks}, vol.~9, pp.~765--772, 1996.

\bibitem{Sprecher97}
D.~A.~Sprecher,  
``A numerical implementation of Kolmogorov’s superpositions II,''  
\textit{Neural Networks}, vol.~10, no.~3, pp.~447--457, 1997.



\bibitem{Koppen02}
M.~Köppen,  
``On the training of a Kolmogorov network,''  
in \textit{Lecture Notes in Computer Science}, vol.~2415, ICANN 2002, pp.~474--479.


\bibitem{Braun09}
J.~Braun and M.~Griebel,  
``On a constructive proof of Kolmogorov’s superposition theorem,''  
\textit{Constr. Approx.}, vol.~30, pp.~653--675, 2009.  
doi: 10.1007/s00365-009-9054-2

\bibitem{Braun09a}
J.~Braun,  
\textit{An Application of Kolmogorov's Superposition Theorem to Function Reconstruction in Higher Dimensions},  
PhD Dissertation, University of Bonn, 2009.


\bibitem{Nakamura93}
M.~Nakamura, R.~Mines, and V.~Kreinovich,  
``Guaranteed intervals for Kolmogorov’s theorem (and their possible relation to neural networks),''  
\textit{Interval Comput.}, vol.~3, pp.~183--199, 1993.

\bibitem{Nees94}
M.~Nees,  
``Approximative versions of Kolmogorov's superposition theorem, proved constructively,''  
\textit{J. Comput. Appl. Math.}, vol.~54, no.~2, pp.~239--250, 1994.

\bibitem{Nees96}
M.~Nees,  
``Chebyshev approximation by discrete superposition: Application to neural networks,''  
\textit{Adv. Comput. Math.}, vol.~5, no.~1, pp.~137--151, 1996.



\bibitem{Figueiredo80}
R.~D.~Figueiredo,  
``Implications and applications of Kolmogorov's superposition theorem,''  
\textit{IEEE Trans. Autom. Control}, vol.~25, pp.~1227--1231, 1980.

\bibitem{Nielsen87}
R.~Hecht-Nielsen,  
``Kolmogorov’s mapping neural network existence theorem,''  
in \textit{Proc. IEEE Int. Conf. Neural Networks}, 1987, pp.~11--13.


\bibitem{Igelnik03}
B.~Igelnik and N.~Parikh,  
``Kolmogorov’s spline network,''  
\textit{IEEE Trans. Neural Netw.}, vol.~14, no.~4, pp.~725--733, 2003.


\bibitem{Coppejans05}
M.~Coppejans,  
``On Kolmogorov’s representation of functions of several variables by functions of one variable,''  
\textit{J. Econometrics}, vol.~123, pp.~1--31, 2004.


\bibitem{Fakhoury22}
D.~Fakhoury, E.~Fakhoury, and H.~Speleers,  
``ExSpliNet: An interpretable and expressive spline-based neural network,''  
\textit{Neural Networks}, vol.~152, pp.~332--346, 2022.

\bibitem{Fakhoury25}
D.~Fakhoury and H.~Speleers,  
``On the expressivity of the ExSpliNet KAN model,''  
\textit{Journal of Computational and Applied Mathematics}, vol.~476, 117053, 2026.

\bibitem{Deventer22}
H.~van~Deventer, P.~J.~van~Rensburg, and A.~Bosman,  
``KASAM: Spline additive models for function approximation,''  
\textit{arXiv preprint}, arXiv:2205.06376, 2022.




\bibitem{Polar21}
A.~Polar and M.~Poluektov,  
``A deep machine learning algorithm for construction of the Kolmogorov–Arnold representation,''  
\textit{Eng.\ Appl.\ Artif.\ Intell.}, vol.~99, 104137, 2021.



\bibitem{Frisch89}
H.~L.~Frisch, C.~Borzi, G.~Ord, J.~K.~Percus, and G.~O.~Williams,  
``Approximate representation of functions of several variables in terms of functions of one variable,''  
\textit{Phys. Rev. Lett.}, vol.~63, no.~9, pp.~927--929, 1989.


\bibitem{Sprecher93}
D.~A.~Sprecher,  
``A universal mapping for Kolmogorov superposition theorem,''  
\textit{Neural Networks}, vol.~6, pp.~1089--1094, 1993.


\bibitem{Schaback05}
R.~Schaback,  
``Multivariate interpolation by polynomials and radial basis functions,''  
\textit{Constructive Approximation}, vol.~21, no.~3, pp.~293--317, 2005.


\bibitem{Wendland04}
H.~Wendland,  
\textit{Scattered Data Approximation}.  
Cambridge, UK: Cambridge University Press, 2004.



\bibitem{TorchKAN}
S. S. Bhattacharjee,  
``TorchKAN: Simplified KAN Model with Variations,''  
2024. [Online]. Available: \url{https://github.com/1ssb/torchkan/}


\bibitem{DeepKAN}
S. Sidharth,  
``Deep-KAN,'' GitHub repository, 2024--2025. [Online]. Available: \url{https://github.com/sidhu2690/Deep-KAN}


\bibitem{RBFKAN}
S. Sidharth,  
``RBF-KAN,'' GitHub repository, 2024--2025. [Online]. Available: \url{https://github.com/sidhu2690/RBF-KAN}


\bibitem{EfficientKAN}
Blealtan,  
``Efficient-KAN: Efficient Kolmogorov-Arnold networks,'' GitHub repository, 2024. [Online]. Available: \url{https://github.com/Blealtan/efficient-kan}


\bibitem{pde_Rigas24}
S. Rigas, M. Papachristou, T. Papadopoulos, F. Anagnostopoulos, and G. Alexandridis,  
``Adaptive training of grid-dependent physics-informed Kolmogorov-Arnold networks,''  
\textit{arXiv preprint} arXiv:2407.17611, 2024. [Online]. Available: \url{https://github.com/srigas/jaxKAN}



\bibitem{Li24}
Z. Li,  
``Kolmogorov-Arnold networks are radial basis function networks,''  
\textit{arXiv preprint} arXiv:2405.06721, 2024. [Online]. Available: \url{https://github.com/ZiyaoLi/fast-kan}



\bibitem{Athanasios2024}
A. Delis,  
``FasterKAN,'' GitHub repository, 2024. [Online]. Available: \url{https://github.com/AthanasiosDelis/faster-kan}



\bibitem{liu2024kan}
Indoxer,  
``LKAN,'' GitHub repository, 2024.  
Available at: \url{https://github.com/Indoxer/LKAN}



\bibitem{pde_fbkan_Howard24}
A. A. Howard, B. Jacob, S. H. Murphy, A. Heinlein, and P. Stinis,  
``Finite basis Kolmogorov-Arnold networks: Domain decomposition for data-driven and physics-informed problems,''  
\textit{arXiv preprint} arXiv:2406.19662, 2024. [Online]. Available: \url{https://github.com/pnnl/neuromancer/tree/feature/fbkans/examples/KANs}




\bibitem{Qiu24}
Q. Qiu, T. Zhu, H. Gong, L. Chen, and H. Ning,  
``ReLU-KAN: New Kolmogorov-Arnold networks that only need matrix addition, dot multiplication, and ReLU,''  
\textit{arXiv preprint} arXiv:2406.02075, 2024. [Online]. Available: \url{https://github.com/quiqi/relu_kan}



\bibitem{Coffman25}
C. Coffman and L. Chen,  
``MatrixKAN: Parallelized Kolmogorov-Arnold network,''  
\textit{arXiv preprint} arXiv:2502.07176, 2025.


\bibitem{FourierKAN}
GistNoesis,  
``FourierKAN,'' GitHub repository, 2024. [Online]. Available: \url{https://github.com/GistNoesis/FourierKAN}

\bibitem{FusedFourierKAN}
GistNoesis,  
``FusedFourierKAN,'' GitHub repository, 2024. [Online]. Available: \url{https://github.com/GistNoesis/FusedFourierKAN}

\bibitem{Aghaei24_fkan}
A. A. Aghaei,  
``fKAN: Fractional Kolmogorov-Arnold networks with trainable Jacobi basis functions,''  
\textit{arXiv preprint} arXiv:2406.07456, 2024. [Online]. Available: \url{https://github.com/alirezaafzalaghaei/fKAN}



\bibitem{Aghaei24_rkan}
A. A. Aghaei,  
``rKAN: Rational Kolmogorov-Arnold networks,''  
\textit{arXiv preprint} arXiv:2406.14495, 2024. [Online]. Available: \url{https://github.com/alirezaafzalaghaei/rKAN}

\bibitem{Wolff25}
M. Wolff, F. Eilers, and X. Jiang,  
``CVKAN: Complex-valued Kolmogorov-Arnold networks,''  
\textit{arXiv preprint} arXiv:2502.02417, 2025. [Online]. Available: \url{https://github.com/M-Wolff/CVKAN}


\bibitem{ChebyKAN}
SynodicMonth,  
``ChebyKAN,'' GitHub repository. [Online]. Available: \url{https://github.com/SynodicMonth/ChebyKAN}




\bibitem{OrthogPolyKANs}
Boris-73-TA,  
``OrthogPolyKANs,'' GitHub repository, 2024. [Online]. Available: \url{https://github.com/Boris-73-TA/OrthogPolyKANs}



\bibitem{kaf_act}
kolmogorovArnoldFourierNetwork,  
``kaf\_act,'' GitHub repository, 2024. [Online]. Available: \url{https://github.com/kolmogorovArnoldFourierNetwork/kaf_act}



\bibitem{KAN_pde_So24}
C. C. So and S. P. Yung,  
``Higher-order ReLU-KANs (HRKANs) for solving physics-informed neural networks (PINNs) more accurately, robustly, and faster,''  
\textit{arXiv preprint} arXiv:2409.14248, 2024.


\bibitem{Xu25_fourier}
J. Xu, Z. Chen, J. Li, S. Yang, W. Wang, X. Hu, and E. C. H. Ngai,  
``FourierKAN-GCF: Fourier Kolmogorov-Arnold network—An effective and efficient feature transformation for graph collaborative filtering,''  
\textit{arXiv preprint} arXiv:2406.01034, 2024. [Online]. Available: \url{https://github.com/Jinfeng-Xu/FKAN-GCF}


\bibitem{MLP-KAN}
Zhangyanbo,  
``MLP-KAN,'' GitHub repository, 2024. [Online]. Available: \url{https://github.com/Zhangyanbo/MLP-KAN}



\bibitem{Yang25_transformer}
X. Yang and X. Wang,  
``Kolmogorov-Arnold transformer,''  
\textit{arXiv preprint} arXiv:2409.10594, 2024. [Online]. Available: \url{https://github.com/Adamdad/kat}

\bibitem{Dong25_FAN}
Y. Dong,  
``FAN: Fourier analysis network,'' GitHub repository, 2024. [Online]. Available: \url{https://github.com/YihongDong/FAN}

\bibitem{Seydi24a}
S. T. Seydi,  
``Exploring the potential of polynomial basis functions in Kolmogorov-Arnold networks: A comparative study of different groups of polynomials,''  
\textit{arXiv preprint} arXiv:2406.02583, 2024.

\bibitem{Bozorgasl24}
Z. Bozorgasl and H. Chen,  
``Wav-KAN: Wavelet Kolmogorov-Arnold networks,''  
\textit{arXiv preprint} arXiv:2405.12832, 2024. [Online]. Available: \url{https://github.com/zavareh1/Wav-KAN}

\bibitem{Liu25_convergence}
W. Liu, E. Chatzi, and Z. Lai,  
``On the rate of convergence of Kolmogorov-Arnold Network regression estimators,''  
\textit{arXiv preprint} arXiv:2509.19830, 2025. [Online]. Available: \url{https://github.com/liouvill/KAN-Converge}


\bibitem{pde_bayesian_Giroux24}
J. Giroux and C. Fanelli,  
``Uncertainty quantification with Bayesian higher order ReLU-KANs,''  
\textit{arXiv preprint} arXiv:2410.01687, 2024.  [Online]. Available: \textit{GitHub repository:} \url{https://github.com/wmdataphys/Bayesian-HR-KAN},



\bibitem{Koenig25}
B. C. Koenig, S. Kim, and S. Deng,  
``LeanKAN: A parameter-lean Kolmogorov-Arnold network layer with improved memory efficiency and convergence behavior,''  
\textit{arXiv preprint} arXiv:2502.17844, 2025. 
 [Online]. Available: \url{https://github.com/DENG-MIT/LeanKAN}


\bibitem{Zhange25_bubble}
Y. Zhang, L. Cheng, A. Gnanaskandan, and A. D. Jagtap,  
``BubbleONet: A Physics-Informed Neural Operator for High-Frequency Bubble Dynamics,''  
\textit{arXiv preprint} arXiv:2508.03965, 2025.


\bibitem{Sifan24}
S. Wang, J. H. Seidman, S. Sankaran, H. Wang, G. J. Pappas, and P. Perdikaris, 
``CViT: Continuous vision transformer for operator learning,'' 
\textit{arXiv preprint} arXiv:2405.13998, 2024.

\bibitem{Toscano24g}
J. D. Toscano, V. Oommen, A. J. Varghese, Z. Zou, N. A. Daryakenari, C. Wu, and G. E. Karniadakis, 
``From PINNs to PIKANs: Recent advances in physics-informed machine learning,'' 
\textit{arXiv preprint} arXiv:2410.13228, 2024.

\bibitem{Cuomo22}
S. Cuomo, V. S. Di Cola, F. Giampaolo, G. Rozza, M. Raissi, and F. Piccialli, 
``Scientific machine learning through physics-informed neural networks: Where we are and what’s next,'' 
\textit{Journal of Scientific Computing}, vol. 92, no. 3, p. 88, 2022.

\bibitem{raissi2018hiddenfluidmechanicsnavierstokes}
M. Raissi, A. Yazdani, and G. E. Karniadakis, 
``Hidden fluid mechanics: A Navier–Stokes informed deep learning framework for assimilating flow visualization data,'' 
\textit{arXiv preprint} arXiv:1808.04327, 2018.


\bibitem{cai2019multi}
W. Cai and Z. Q. J. Xu, 
``Multi-scale deep neural networks for solving high dimensional PDEs,'' 
\textit{arXiv preprint} arXiv:1910.11710, 2019.


\bibitem{lu2019deeponet}
L. Lu, P. Jin, and G. E. Karniadakis,  
``DeepONet: Learning nonlinear operators for identifying differential equations based on the universal approximation theorem of operators,''  
\textit{arXiv preprint} arXiv:1910.03193, 2019.

\bibitem{jagtap2020locally}
A. D. Jagtap, K. Kawaguchi, and G. E. Karniadakis,  
``Locally adaptive activation functions with slope recovery for deep and physics-informed neural networks,''  
\textit{Proceedings of the Royal Society A}, vol. 476, no. 2239, p. 20200334, 2020.

\bibitem{yang2021b}
L. Yang, X. Meng, and G. E. Karniadakis,  
``B-PINNs: Bayesian physics-informed neural networks for forward and inverse PDE problems with noisy data,''  
\textit{Journal of Computational Physics}, vol. 425, p. 109913, 2021.

\bibitem{Wang2020_Fourier_nets}
S. Wang, H. Wang, and P. Perdikaris,  
``On the eigenvector bias of Fourier feature networks: From regression to solving multi-scale PDEs with physics-informed neural networks,''  
\textit{arXiv preprint} arXiv:2012.10047, 2020.

\bibitem{Mattey22}
R. Mattey and S. Ghosh,  
``A novel sequential method to train physics-informed neural networks for Allen–Cahn and Cahn–Hilliard equations,''  
\textit{Computer Methods in Applied Mechanics and Engineering}, vol. 390, p. 114474, 2022.

\bibitem{cho2023sep}
J. Cho, S. Nam, H. Yang, S.-B. Yun, Y. Hong, and E. Park,  
``Separable physics-informed neural networks,''  
\textit{arXiv preprint} arXiv:2306.15969, 2023.

\bibitem{zhang2023artificial}
Q. Zhang, C. Wu, A. Kahana, Y. Kim, Y. Li, G. E. Karniadakis, and P. Panda,  
``Artificial to spiking neural networks conversion for scientific machine learning,''  
\textit{arXiv preprint} arXiv:2308.16372, 2023.

\bibitem{anagnostopoulos2023residual}
S. J. Anagnostopoulos, J. D. Toscano, N. Stergiopulos, and G. E. Karniadakis,  
``Residual-based attention and connection to information bottleneck theory in PINNs,''  
\textit{arXiv preprint} arXiv:2307.00379, 2023.

\bibitem{howard2023stacked}
A. A. Howard, S. H. Murphy, S. E. Ahmed, and P. Stinis,  
``Stacked networks improve physics-informed training: Applications to neural networks and deep operator networks,''  
\textit{arXiv preprint} arXiv:2311.06483, 2023.

\bibitem{Hou23}
J. Hou, Y. Li, and S. Ying,  
``Enhancing PINNs for solving PDEs via adaptive collocation point movement and adaptive loss weighting,''  
\textit{Nonlinear Dynamics}, vol. 111, no. 16, pp. 15233--15261, 2023.



\bibitem{BarNatan}
D. Bar-Natan,  
``Dessert: Hilbert’s 13th problem, in full colour,'' [Online]. 



\bibitem{Guo25_Fourier}
H. Guo, L. Zheng, S. Lin, M. Dong, X. Cao, and H. Zheng,  
``A novel and robust Fourier-based Kolmogorov–Arnold Network in early warning of rockbursts from microseismic data,''  
\textit{Stochastic Environmental Research and Risk Assessment}, 2025.

\bibitem{Hassan25_Bayesian}
M. M. Hassan,  
``Bayesian Kolmogorov–Arnold Networks: Uncertainty-aware interpretable modeling through probabilistic spline decomposition,''  
\textit{Physica A: Statistical Mechanics and its Applications}, vol. 680, p. 131041, 2025.



\bibitem{Lai21}
M.-J. Lai and Z. Shen,  
``The Kolmogorov superposition theorem can break the curse of dimensionality when approximating high dimensional functions,''  
\textit{arXiv preprint} arXiv:2112.09963, 2021.


\bibitem{Leni13}
P.-E. Leni, Y. D. Fougerolle, and F. Truchetet,  
``The Kolmogorov spline network for image processing,''  
in \textit{Image Processing: Concepts, Methodologies, Tools, and Applications}, pp. 54--78, IGI Global, 2013.

\bibitem{Schoots25}
N. Schoots and M. J. Villani,  
``Relating piecewise linear Kolmogorov-Arnold networks to ReLU networks,''  
\textit{arXiv preprint} arXiv:2503.01702, 2025.

\bibitem{Gao25}
Y. Gao and V. Y. Tan,  
``On the convergence of (stochastic) gradient descent for Kolmogorov–Arnold networks,''  
\textit{IEEE Transactions on Information Theory}, 2025 (early access).


\bibitem{Bodner24}
A.~D. Bodner, A.~S. Tepsich, J.~N. Spolski, and S. Pourteau,  
``Convolutional Kolmogorov-Arnold Networks,''  
\textit{arXiv preprint} arXiv:2406.13155, 2024.





\bibitem{Sun23_opt}
Y. Sun, U. Sengupta, and M. Juniper,  
``Physics-informed deep learning for simultaneous surrogate modeling and PDE-constrained optimization of an airfoil geometry,''  
\textit{Computer Methods in Applied Mechanics and Engineering}, vol. 411, p. 116042, 2023.

\bibitem{Chen19_ode}
R. T. Q. Chen, Y. Rubanova, J. Bettencourt, and D. Duvenaud,  
``Neural Ordinary Differential Equations,''  
\textit{arXiv preprint} arXiv:1806.07366, 2019.

\bibitem{Eghbalian23}
M. Eghbalian, M. Pouragha, and R. Wan,  
``A physics-informed deep neural network for surrogate modeling in classical elasto-plasticity,''  \textit{Computers and Geotechnics}, vol. 159, p. 105472, 2023.

\bibitem{Song25}
S. Song, T. Mukerji, and D. Zhang,  
``Physics-informed multigrid neural operator: Theory and an application to porous flow simulation,''  
\textit{Journal of Computational Physics}, vol. 520, p. 113438, 2025.

\bibitem{Yu24_res}
R. C. Yu, S. Wu, and J. Gui,  
``Residual Kolmogorov-Arnold network for enhanced deep learning,''  
\textit{arXiv preprint} arXiv:2410.05500, 2024.  
GitHub: \url{https://github.com/withray/residualKAN}

\bibitem{AutoEncoder}
SekiroRong,  
``KAN-AutoEncoder,''  
GitHub repository, 2024.  
Available at: \url{https://github.com/SekiroRong/KAN-AutoEncoder}


\bibitem{Li24b}
C. Li, X. Liu, W. Li, C. Wang, H. Liu, and Y. Yuan,  
``U-KAN makes strong backbone for medical image segmentation and generation,''  
\textit{arXiv preprint} arXiv:2406.02918, 2024.

\bibitem{Seydi24b}
S. T. Seydi,  
``Unveiling the power of wavelets: A wavelet-based Kolmogorov-Arnold network for hyperspectral image classification,''  
\textit{arXiv preprint} arXiv:2406.07869, 2024.

\bibitem{Cheon24}
M. Cheon,  
``Kolmogorov-Arnold Network for satellite image classification in remote sensing,''  
\textit{arXiv preprint} arXiv:2406.00600, 2024.


\bibitem{Chen24}
Y. Chen, Z. Zhu, S. Zhu, L. Qiu, B. Zou, F. Jia, Y. Zhu, C. Zhang, Z. Fang, F. Qin, \emph{et al.},  
``SCKansformer: Fine-grained classification of bone marrow cells via Kansformer backbone and hierarchical attention mechanisms,''  
\textit{arXiv preprint} arXiv:2406.09931, 2024.


\bibitem{timeSeries_Zhou24}
Q. Zhou, C. Pei, F. Sun, J. Han, Z. Gao, D. Pei, H. Zhang, G. Xie, and J. Li,  
``KAN-AD: Time series anomaly detection with Kolmogorov-Arnold Networks,''  
\textit{arXiv preprint} arXiv:2411.00278, 2024.



\bibitem{Amir_PUGKAN}
A. Noorizadegan,  
``Partition-of-Unity Gaussian Kolmogorov--Arnold Networks,''  
\textit{arXiv preprint} arXiv:2604.23599, 2026.

\bibitem{Amir_GKAN}
A. Noorizadegan and S. Wang,  
``Scaling of Gaussian Kolmogorov--Arnold Networks,''  
\textit{arXiv preprint} arXiv:2604.21174, 2026.

\bibitem{timeSeries_Rubio24}
C.~J. Vaca-Rubio, L. Blanco, R. Pereira, and M. Caus,  
``Kolmogorov-Arnold Networks (KANs) for time series analysis,''  
\textit{arXiv preprint} arXiv:2405.08790, 2024.

\bibitem{Xu24}
K. Xu, L. Chen, and S. Wang,  
``Kolmogorov-Arnold Networks for time series: Bridging predictive power and interpretability,''  
\textit{arXiv preprint} arXiv:2406.02496, 2024.

\bibitem{Genet24a}
R. Genet and H. Inzirillo,  
``TKAN: Temporal Kolmogorov-Arnold Networks,''  
\textit{arXiv preprint} arXiv:2405.07344, 2024.

\bibitem{Genet24b}
R. Genet and H. Inzirillo,  
``A temporal Kolmogorov-Arnold transformer for time series forecasting,''  
\textit{arXiv preprint} arXiv:2406.02486, 2024.

\bibitem{GraphKAN_WillHua}
W. Hua,  
``GraphKAN: Graph Kolmogorov–Arnold Networks,'' GitHub repository, 2024.  
Available at: \url{https://github.com/WillHua127/GraphKAN-Graph-Kolmogorov-Arnold-Networks}

\bibitem{GraphKAN_LiuYue}
Y. Liu,  
``KAN4Graph,'' GitHub repository, 2023.  
Available at: \url{https://github.com/yueliu1999/KAN4Graph}




\bibitem{Kiamari24}
M. Kiamari, M. Kiamari, and B. Krishnamachari,  
``GKAN: Graph Kolmogorov-Arnold Networks,''  
\textit{arXiv preprint} arXiv:2406.06470, 2024.



\bibitem{Azam24}
B. Azam and N. Akhtar,  
``Suitability of Kolmogorov-Arnold Networks for computer vision: A preliminary investigation,''  
\textit{arXiv preprint} arXiv:2406.09087, 2024.

\bibitem{Cheon24a}
M. Cheon,  
``Demonstrating the efficacy of Kolmogorov-Arnold Networks in vision tasks,''  
\textit{arXiv preprint} arXiv:2406.14916, 2024.





\bibitem{Basina24}
D. Basina, J. R. Vishal, A. Choudhary, and B. Chakravarthi,  
``KAT to KANs: A review of Kolmogorov-Arnold networks and the neural leap forward,''  
\textit{arXiv preprint} arXiv:2411.10622, 2024.

\bibitem{Lin25_geo}
S. Lin, M. Dong, H. Guo, L. Zheng, K. Zhao, and H. Zheng,  
``Early warning system for risk assessment in geotechnical engineering using Kolmogorov-Arnold networks,''  
\textit{Journal of Rock Mechanics and Geotechnical Engineering}, 2025 (in press).

\bibitem{pde_Howard24}
A. A. Howard, B. Jacob, and P. Stinis,
``Multifidelity Kolmogorov-Arnold networks,''
\textit{Machine Learning: Science and Technology}, 6(3):035038, 2025.

\bibitem{pde_jacob24}
B. Jacob, A. A. Howard, and P. Stinis,  
``SPIKANs: Separable physics-informed Kolmogorov-Arnold networks,''  
\textit{arXiv preprint} arXiv:2411.06286, 2024.
Available at: \url{https://github.com/pnnl/spikans}

\bibitem{pde_Patra24}
S. Patra, S. Panda, B. K. Parida, M. Arya, K. Jacobs, D. I. Bondar, and A. Sen,  
``Physics-informed Kolmogorov-Arnold neural networks for dynamical analysis via Efficient-KAN and Wav-KAN,''  
\textit{arXiv preprint} arXiv:2407.18373, 2024.

\bibitem{pde_Ranasinghe24}
N. Ranasinghe, Y. Xia, S. Seneviratne, and S. Halgamuge,  
``GINN-KAN: Interpretability pipelining with applications in physics-informed neural networks,''  
\textit{arXiv preprint} arXiv:2408.14780, 2024.

\bibitem{Raffel25}
M. Raffel and L. Chen,  
``FlashKAT: Understanding and addressing performance bottlenecks in the Kolmogorov-Arnold transformer,''  
\textit{arXiv preprint} arXiv:2505.13813, 2025.

\bibitem{Wang25old}
Y. Wang, C. Zhu, S. Zhang, C. Xiang, Z. Gao, G. Zhu, ... and X. Shen,  
``Accurately models the relationship between physical response and structure using Kolmogorov–Arnold network,''  
\textit{Advanced Science}, vol. 12, no. 12, p. 2413805, 2025.



\bibitem{Shen25}
Y. Shen, J. Du, Y. He, Y. Wang, T. Hao, X. Niu, and Q. Jiang,  
``AAKAN-WGAN: Predicting mechanical properties of magnesium alloys based on generative design and adaptive activation of Kolmogorov-Arnold networks,''  
\textit{Materials Today Communications}, vol. 47, p. 113198, 2025.


\bibitem{SS24}
SS, Sidharth, Keerthana AR, and Anas KP  
``Chebyshev polynomial-based Kolmogorov-Arnold networks: An efficient architecture for nonlinear function approximation,''  
\textit{arXiv preprint} arXiv:2405.07200, 2024.

\bibitem{Guo24}
C. Guo, L. Sun, S. Li, and Z. Yuan, C. Wang,  
``Physics-informed Kolmogorov-Arnold network with Chebyshev polynomials for fluid mechanics,''  
\textit{arXiv preprint} arXiv:2411.04516, 2024.


\bibitem{Schaback95}
R. Schaback,  
``Error estimates and condition numbers for radial basis function interpolation,''  
\textit{Advances in Computational Mathematics}, vol. 3, no. 3, pp. 251--264, 1995.

\bibitem{Schaback23}
R. Schaback,  
``Small errors imply large evaluation instabilities,''  
\textit{Advances in Computational Mathematics}, vol. 49, no. 2, p. 25, 2023.


\bibitem{Amir22}
A. Noorizadegan, C.-S. Chen, D. L. Young, and C. S. Chen,  
``Effective condition number for the selection of the RBF shape parameter with the fictitious point method,''  
\textit{Applied Numerical Mathematics}, vol. 178, pp. 280--295, 2022.

\bibitem{Amir23}
C.-S. Chen, A. Noorizadegan, D. L. Young, and C. S. Chen,  
``On the selection of a better radial basis function and its shape parameter in interpolation problems,''  
\textit{Applied Mathematics and Computation}, vol. 442, p. 127713, 2023.

\bibitem{Larsson24}
E. Larsson and R. Schaback,  
``Scaling of radial basis functions,''  
\textit{IMA Journal of Numerical Analysis}, vol. 44, no. 2, pp. 1130--1152, 2024.


\bibitem{Tizian1}
T. Wenzel,  
``Sharp inverse statements for kernel interpolation,''  
\textit{Mathematics of Computation}, vol. 95, no. 359, pp. 1389--1413, 2026.


\bibitem{Tizian2}
T. Wenzel and G. Santin,  
``On the optimal shape parameter for kernel methods: Sharp direct and inverse statements,''  
\textit{arXiv preprint} arXiv:2601.14070, 2026.


\bibitem{Tizian3}
T. Wenzel and A. Iske,  
``Spectral alignment of kernel matrices and applications,''  
\textit{SIAM Journal on Matrix Analysis and Applications}, vol. 47, no. 1, pp. 265--281, 2026.

\bibitem{Ling06}
J. Wertz, E. J. Kansa, and L. Ling,  
``The role of the multiquadric shape parameters in solving elliptic partial differential equations,''  
\textit{Computers \& Mathematics with Applications}, vol. 51, no. 8, pp. 1335--1348, 2006.


\bibitem{Ling20}
S. N. Chiu, L. Ling, and M. McCourt,  
``On variable and random shape Gaussian interpolations,''  
\textit{Applied Mathematics and Computation}, vol. 377, p. 125159, 2020.

\bibitem{Amir_s24}
A. Noorizadegan and R. Schaback,  
``Introducing the evaluation condition number: A novel assessment of conditioning in radial basis function methods,''  
\textit{Engineering Analysis with Boundary Elements}, vol. 166, p. 105827, 2024.

\bibitem{Cavoretto21AD}
R. Cavoretto, A. De Rossi, M. S. Mukhametzhanov, and Y. D. Sergeyev,  
``On the search of the shape parameter in radial basis functions using univariate global optimization methods,''  
\textit{Journal of Global Optimization}, vol. 79, no. 2, pp. 305--327, 2021.

\bibitem{Fasshauer07}
G. E. Fasshauer and J. G. Zhang,  
``On choosing `optimal' shape parameters for RBF approximation,''  
\textit{Numerical Algorithms}, vol. 45, no. 1, pp. 345--368, 2007.



\bibitem{Alesiani25a}
F.~Alesiani, H.~Christiansen, and F.~Errica,  
``Variational Kolmogorov--Arnold Network,''  
arXiv:2507.02466, 2025.

\bibitem{Alesiani25b}
F.~Alesiani, T.~Maruyama, H.~Christiansen, and V.~Zaverkin,  
``Geometric Kolmogorov--Arnold Superposition Theorem,''  
arXiv:2502.16664, 2025.



\bibitem{Gong25}
Y. Gong, Y. He, Y. Mei, X. Zhuang, F. Qin, and T. Rabczuk,  
``Physics-Informed Kolmogorov–Arnold Networks for multi-material elasticity problems in electronic packaging,''  
\textit{arXiv preprint} arXiv:2508.16999, 2025.

\bibitem{Wei26_battery}
C. Wei, H. Pang, T. Huang, Z. Quan, and Z. Qian,  
``Predicting lithium-ion battery health using attention mechanism with Kolmogorov–Arnold and physics-informed neural networks,''  
\textit{Expert Systems with Applications}, vol. 296, p. 128969, 2026.


\bibitem{Grossmann84}
A. Grossmann and J. Morlet,  
``Decomposition of Hardy functions into square integrable wavelets of constant shape,''  
\textit{SIAM Journal on Mathematical Analysis}, vol. 15, no. 4, pp. 723--736, 1984.

\bibitem{Calderon64}
A. Calderón,  
``Intermediate spaces and interpolation, the complex method,''  
\textit{Studia Mathematica}, vol. 24, no. 2, pp. 113--190, 1964.

\bibitem{Pratyush24}
P. Pratyush, C. Carrier, S. Pokharel, H. D. Ismail, M. Chaudhari, and D. B. KC,  
``CaLMPhosKAN: Prediction of general phosphorylation sites in proteins via fusion of codon aware embeddings with amino acid aware embeddings and wavelet-based Kolmogorov-Arnold network,''  
\textit{bioRxiv}, 2024.




\bibitem{Moseley23}
B. Moseley, A. Markham, and T. Nissen-Meyer,  
``Finite basis physics-informed neural networks (FBPINNs): A scalable domain decomposition approach for solving differential equations,''  
\textit{Advances in Computational Mathematics}, vol. 49, no. 4, p. 62, 2023.

\bibitem{Dolean24}
V. Dolean, A. Heinlein, S. Mishra, and B. Moseley,  
``Finite basis physics-informed neural networks as a Schwarz domain decomposition method,''  
in \textit{DDM in Science and Engineering XXVII}, pp. 165--172, Springer, 2024.

\bibitem{Dolean24b}
V. Dolean, A. Heinlein, S. Mishra, and B. Moseley,  
``Multilevel domain decomposition-based architectures for physics-informed neural networks,''  
\textit{Computer Methods in Applied Mechanics and Engineering}, vol. 429, p. 117116, 2024.

\bibitem{Heinlein24}
A. Heinlein, A. A. Howard, D. Beecroft, and P. Stinis,  
``Multifidelity domain decomposition–based physics-informed neural networks for time-dependent problems,''  
\textit{arXiv preprint} arXiv:2401.07888, 2024.

\bibitem{Roberto1}
R. Cavoretto, A. De Rossi, and E. Perracchione,  
``Efficient computation of partition of unity interpolants through a block-based searching technique,''  
\textit{Computers \& Mathematics with Applications}, vol. 71, no. 12, pp. 2568--2581, 2016.

\bibitem{Roberto2}
G. Garmanjani, R. Cavoretto, and M. Esmaeilbeigi,  
``A RBF partition of unity collocation method based on finite difference for initial–boundary value problems,''  
\textit{Computers \& Mathematics with Applications}, vol. 75, no. 11, pp. 4066--4090, 2018.

\bibitem{Roberto3}
R. Cavoretto, S. De Marchi, A. De Rossi, E. Perracchione, and G. Santin,  
``Partition of unity interpolation using stable kernel-based techniques,''  
\textit{Applied Numerical Mathematics}, vol. 116, pp. 95--107, 2017.

\bibitem{Anagnostopoulos24}
S. J. Anagnostopoulos, J. D. Toscano, N. Stergiopulos, and G. E. Karniadakis,  
``Residual-based attention in physics-informed neural networks,''  
\textit{Computer Methods in Applied Mechanics and Engineering}, vol. 421, p. 116805, 2024.

\bibitem{Wang19}
Z. Wang, M. S. Triantafyllou, Y. Constantinides, and G. E. Karniadakis,  
``An entropy-viscosity large eddy simulation study of turbulent flow in a flexible pipe,''  
\textit{Journal of Fluid Mechanics}, vol. 859, pp. 691--730, 2019.

\bibitem{Guermond11}
J.-L. Guermond, R. Pasquetti, and B. Popov,  
``Entropy viscosity method for nonlinear conservation law,''  
\textit{Journal of Computational Physics}, vol. 230, no. 11, pp. 4248--4267, 2011.

\bibitem{Toscano25_aivt}
J. D. Toscano, et al.,  
``AIVT: Inference of turbulent thermal convection from measured 3D velocity data by physics-informed Kolmogorov-Arnold networks,''  
\textit{Science Advances}, vol. 11, p. eads5236, 2025.

\bibitem{Zienkiewicz89}
O. Zienkiewicz, J. Zhu, and N. Gong,  
``Effective and practical h–p-version adaptive analysis procedures for the finite element method,''  
\textit{International Journal for Numerical Methods in Engineering}, vol. 28, no. 4, pp. 879--891, 1989.

\bibitem{Nguyen21}
V. M. Nguyen-Thanh, C. Anitescu, N. Alajlan, T. Rabczuk, and X. Zhuang,  
``Parametric deep energy approach for elasticity accounting for strain gradient effects,''  
\textit{Computer Methods in Applied Mechanics and Engineering}, vol. 386, p. 114096, 2021.

\bibitem{Altarabichi24}
M. G. Altarabichi,  
``DropKAN: Regularizing KANs by masking post-activations,''  
\textit{arXiv preprint} arXiv:2407.13044, 2024.

\bibitem{Jacot18}
A. Jacot, F. Gabriel, and C. Hongler,  
``Neural tangent kernel: Convergence and generalization in neural networks,''  
in \textit{Advances in Neural Information Processing Systems (NeurIPS)}, vol. 31, 2018.
\textit{Physics}, vol. 449, p. 110768, 2022.

\bibitem{DeVore98}
R. A. DeVore,  
``Nonlinear approximation,''  
\textit{Acta Numerica}, vol. 7, pp. 51--150, 1998.

\bibitem{DeVore21}
R. DeVore, B. Hanin, and G. Petrova,  
``Neural network approximation,''  
\textit{Acta Numerica}, vol. 30, pp. 327--444, 2021.

\bibitem{Chen19}
R. T. Q. Chen, Y. Rubanova, J. Bettencourt, and D. Duvenaud,  
``Neural ordinary differential equations,''  
\textit{arXiv preprint} arXiv:1806.07366, 2019.

\bibitem{Tsitouras11}
C. Tsitouras,  
``Runge–Kutta pairs of order 5(4) satisfying only the first column simplifying assumption,''  
\textit{Computers \& Mathematics with Applications}, vol. 62, no. 2, pp. 770--775, 2011.

\bibitem{Udrescu20}
S.-M. Udrescu and M. Tegmark,  
``AI Feynman: A physics-inspired method for symbolic regression,''  
\textit{Sci. Adv.}, vol. 6, eaay2631, 2020. doi: 10.1126/sciadv.aay2631

\bibitem{Buhler25_regression}
M. A. Bühler and G. Guillén-Gosálbez,  
``KAN-SR: A Kolmogorov-Arnold Network Guided Symbolic Regression Framework,''  
\textit{arXiv preprint} arXiv:2509.10089, 2025.

\bibitem{Lee22_complex}
C. Y. Lee, H. Hasegawa, and S. C. Gao,  
``Complex-valued neural networks: A comprehensive survey,''  
\textit{IEEE/CAA J. Autom. Sinica}, vol. 9, no. 8, pp. 1406--1426, Aug. 2022. doi: 10.1109/JAS.2022.105743

\bibitem{Chen94}
S. Chen, S. McLaughlin, and B. Mulgrew,  
``Complex-valued radial basic function network, part I: network architecture and learning algorithms,''  
\textit{Signal Process.}, vol. 35, no. 1, pp. 19--31, 1994.

\bibitem{Che25_complexKAN}
R. Che, L. af Klinteberg, and M. Aryapoor,  
``Improved Complex-Valued Kolmogorov–Arnold Networks with Theoretical Support,''  
in \textit{Proc. 24th EPIA Conf. on Artificial Intelligence (EPIA)}, Faro, Portugal, Oct. 2025, Part I, pp. 439--451. Springer, Heidelberg. doi: 10.1007/978-3-032-05176-9\_34

\bibitem{Chen25_weak}
Z. Chen, Z. Zeng, P. Hu, and Y. Zhu,  
``Weak Collocation Networks: A deep learning approach to reconstruct stochastic dynamics from aggregate data,''  
\textit{Commun. Nonlinear Sci. Numer. Simul.}, 2025.


\bibitem{Gong25_sensors}
Y. Gong, G. Shi, J. Zhu, S. Tang, H. Liu, S. Hu, and S. Chen,  
``Sparse sensor measurement for indoor physical field reconstruction with physics-inspired K-means and Kolmogorov-Arnold networks,''  
\textit{J. Build. Eng.}, vol. 113, p. 114115, 2025.

\bibitem{Lee25_operator}
J. Lee, Z. Liu, X. Yu, Y. Wang, H. Jeong, M. Y. Niu, and Z. Zhang,  
``KANO: Kolmogorov-Arnold Neural Operator,''  
\textit{arXiv preprint} arXiv:2509.16825, 2025.

\bibitem{Li20_FNO}
Z. Li, N. Kovachki, K. Azizzadenesheli, B. Liu, K. Bhattacharya, A. Stuart, and A. Anandkumar,  
``Fourier neural operator for parametric partial differential equations,''  
\textit{arXiv preprint} arXiv:2010.08895, 2020.


\bibitem{Sen25_time}
A. Sen, I. V. Lukin, K. Jacobs, L. Kaplan, A. G. Sotnikov, and D. I. Bondar,  
``Physics-informed time series analysis with Kolmogorov-Arnold Networks under Ehrenfest constraints,''  
\textit{arXiv preprint} arXiv:2509.18483, 2025.

\bibitem{Ehrenfest1927}
P. Ehrenfest,  
``Bemerkung über die angenäherte Gültigkeit der klassischen Mechanik innerhalb der Quantenmechanik,''  
\textit{Z. Phys.}, vol. 45, pp. 455--457, 1927.

\bibitem{Wang25_Hamiltonian}
F. Wang, L. Chen, and J. Ding,  
``Symplectic physics-embedded learning via Lie groups Hamiltonian formulation for serial manipulator dynamics prediction,''  
\textit{Sci. Rep.}, vol. 15, p. 33179, 2025. doi: 10.1038/s41598-025-17935-w

\bibitem{Duong24_Hamiltonian}
T. P. Duong, A. Altawaitan, J. Stanley, and N. Atanasov,  
``Port-Hamiltonian neural ODE networks on Lie groups for robot dynamics learning and control,''  
\textit{IEEE Trans. Rob.}, vol. 40, pp. 3695--3715, 2024.

\bibitem{Zou25_Probabilistic}
Q. Zou and H. Yan,  
``Probabilistic Kolmogorov-Arnold Networks via sparsified deep Gaussian processes with additive kernels,''  
in \textit{Proc. IEEE 21st Int. Conf. Automation Science and Engineering (CASE)}, Los Angeles, USA, Aug. 2025, pp. 2889--2894. IEEE.

\bibitem{Guo25_equation}
G. Guo, Z. Tang, Z. Cui, C. Li, H. Wang, and H. You,  
``Accurate analytic equation generation for compact modeling with physics-assisted Kolmogorov-Arnold networks,''  
\textit{ACM Trans. Des. Autom. Electron. Syst.}, Sept. 2025. doi: 10.1145/3765904

\bibitem{jacot2018neural}
A. Jacot, F. Gabriel, and C. Hongler, 
``Neural tangent kernel: Convergence and generalization in neural networks,'' 
\textit{Adv. Neural Inf. Process. Syst.}, vol. 31, 2018.

\bibitem{lee2019wide}
J. Lee, L. Xiao, S. S. Schoenholz, Y. Bahri, R. Novak, J. Sohl\textendash Dickstein, and J. Pennington, 
``Wide neural networks of any depth evolve as linear models under gradient descent,'' 
\textit{Adv. Neural Inf. Process. Syst.}, vol. 32, 2019.

\bibitem{rahaman2019spectral}
N. Rahaman, A. Baratin, D. Arpit, F. Draxler, M. Lin, F. Hamprecht, Y. Bengio, and A. Courville, 
``On the spectral bias of neural networks,'' 
in \textit{Proc. Int. Conf. Machine Learning (ICML)}, PMLR, vol. 97, pp. 5301--5310, 2019.

\bibitem{xu2019frequency}
Z.\textendash Q. J. Xu, Y. Zhang, T. Luo, Y. Xiao, and Z. Ma, 
``Frequency principle: Fourier analysis sheds light on deep neural networks,'' 
\textit{arXiv:1901.06523}, 2019.

\bibitem{ghorbani2019investigation}
B. Ghorbani, S. Krishnan, and Y. Xiao, 
``An investigation into neural net optimization via Hessian eigenvalue density,'' 
in \textit{Proc. Int. Conf. Machine Learning (ICML)}, PMLR, vol. 97, pp. 2232--2241, 2019.

\bibitem{foret2020sharpness}
P. Foret, A. Kleiner, H. Mobahi, and B. Neyshabur, 
``Sharpness\textendash aware minimization for efficiently improving generalization,'' 
in \textit{Proc. Int. Conf. Learning Representations (ICLR)}, 2021.

\bibitem{zhang2021understanding}
C. Zhang, S. Bengio, M. Hardt, B. Recht, and O. Vinyals, 
``Understanding deep learning (still) requires rethinking generalization,'' 
\textit{Commun. ACM}, vol. 64, no. 3, pp. 107--115, 2021.

\bibitem{tancik2020fourier}
M. Tancik, P. Srinivasan, B. Mildenhall, S. Fridovich\textendash Keil, N. Raghavan, U. Singhal, R. Ramamoorthi, J. Barron, and R. Ng, 
``Fourier features let networks learn high frequency functions in low dimensional domains,'' 
\textit{Adv. Neural Inf. Process. Syst.}, vol. 33, pp. 7537--7547, 2020.

\bibitem{hong2022activation}
Q. Hong, J. W. Siegel, Q. Tan, and J. Xu, 
``On the activation function dependence of the spectral bias of neural networks,'' 
\textit{arXiv:2208.04924}, 2022.

\bibitem{Luo25_DAE}
K. Luo, J. Tang, M. Cai, X. Zeng, M. Xie, and M. Yan,  
``DAE-KAN: A Kolmogorov–Arnold Network model for high-index differential–algebraic equations,''  
\textit{arXiv:2504.15806}, 2025.


\bibitem{Clafa25}
M. Calafà, T. Andriollo, A. P. Engsig-Karup, and C. H. Jeong,  
``A holomorphic Kolmogorov-Arnold network framework for solving elliptic problems on arbitrary 2D domains,''  
\textit{arXiv preprint} arXiv:2507.22678, 2025.  
GitHub: \url{https://github.com/teocala/pihnn}


\bibitem{Menon25}
S. S. Menon and A. D. Jagtap,  
``Anant-Net: Breaking the curse of dimensionality with scalable and interpretable neural surrogate for high-dimensional PDEs,''  
\textit{Computer Methods in Applied Mechanics and Engineering}, vol. 447, p. 118403, 2025.  
GitHub: \url{https://github.com/ParamIntelligence/Anant-Net}

\bibitem{Kundu24_quantom}
A. Kundu, A. Sarkar, and A. Sadhu, 
``KANQAS: Kolmogorov-Arnold Network for Quantum Architecture Search,''  
\textit{EPJ Quantum Technology}, vol. 11, p. 76, 2024.  
GitHub: \url{https://github.com/Aqasch/KANQAS_code}

\bibitem{Polar25}
A. Polar and M. Poluektov,  
``Probabilistic Kolmogorov–Arnold Network: An Approach for Stochastic Modelling Using Divisive Data Re-Sorting,''  
\textit{Modelling}, vol. 6, no. 3, p. 88, 2025.  
GitHub: \url{https://github.com/andrewpolar/pkan}



\bibitem{pde_Rigas24f}
S. Rigas, M. Papachristou, T. Papadopoulos, F. Anagnostopoulos, and G. Alexandridis,  
``Adaptive Training of Grid-Dependent Physics-Informed Kolmogorov-Arnold Networks,''  
\textit{IEEE Access}, vol. 12, pp. 176982--176998, 2024.  
GitHub: \url{https://github.com/srigas/jaxKAN}

\bibitem{pde_Rigas24a}
S. Rigas, M. Papachristou, T. Papadopoulos, F. Anagnostopoulos, and G. Alexandridis,  
``jaxKAN: A unified JAX framework for Kolmogorov-Arnold Networks,''  
\textit{Journal of Open Source Software}, vol. 10, no. 108, p. 7830, 2025.  
GitHub: \url{https://github.com/srigas/jaxKAN}

\bibitem{Guilhoto24}
L. F. Guilhoto and P. Perdikaris,  
``Deep Learning Alternatives of the Kolmogorov Superposition Theorem,''  
\textit{Proc. ICLR}, 2025.

\bibitem{Rigas25_deep}
S. Rigas, F. Anagnostopoulos, M. Papachristou, and G. Alexandridis,  
``Towards deep physics-informed Kolmogorov-Arnold networks,''  
\textit{arXiv preprint} arXiv:2510.23501, 2025.  
GitHub: \url{https://github.com/srigas/RGA-KANs}

\bibitem{Kiyani25}
E. Kiyani, K. Shukla, J. F. Urbán, J. Darbon, and G. E. Karniadakis,  
``Optimizing the optimizer for physics-informed neural networks and Kolmogorov-Arnold networks,''  
\textit{Computer Methods in Applied Mechanics and Engineering}, vol. 446, p. 118308, 2025.

\bibitem{Polar25_Newton}
M. Poluektov and A. Polar,  
``Construction of the Kolmogorov-Arnold networks using the Newton-Kaczmarz method,''  
\textit{Machine Learning}, vol. 114, p. 185, 2025.

\bibitem{Lai25_kst}
M.~J.~Lai and Z.~Shen,  
``Optimal linear B-spline approximation via Kolmogorov superposition theorem and its applications,''  
\textit{Sampling Theory, Signal Processing, and Data Analysis}, vol.~23, 11, 2025.  
[Online]. Available: \url{https://doi.org/10.1007/s43670-025-00107-2}

\bibitem{Lai25_kst_a}
M.~J.~Lai and Z.~Shen,  
``The Kolmogorov superposition theorem can break the curse of dimensionality when approximating high-dimensional functions,''  
\textit{arXiv preprint} arXiv:2112.09963, 2021.

\bibitem{Pant25}
R.~Pant, S.~Li, X.~Li, H.~Iqbal, and K.~Kumar,  
``MLPs and KANs for data-driven learning in physical problems: A performance comparison,''  
\textit{arXiv preprint} arXiv:2504.11397, 2025.  
[Online]. Available: \url{https://github.com/geoelements-dev/mlp-kan}

\bibitem{Pozdnyakov25}
S.~Pozdnyakov and P.~Schwaller,  
``Lookup multivariate Kolmogorov-Arnold Networks,''  
\textit{arXiv preprint} arXiv:2509.07103, 2025.  
[Online]. Available: \url{https://github.com/schwallergroup/lmkan}

\bibitem{Cui25_flow}
S.~Cui, M.~Cao, Y.~Liao, and J.~Wu,  
``Physics-informed Kolmogorov–Arnold networks: Investigating architectures and hyperparameter impacts for solving Navier–Stokes equations,''  
\textit{Physics of Fluids}, vol.~37, no.~3, 037159, 2025.  
[Online]. Available: \url{https://doi.org/10.1063/5.0257677}

\bibitem{Meshir25_NTK}
J.~D.~Meshir, A.~Palafox, and E.~A.~Guerrero,  
``On the study of frequency control and spectral bias in wavelet-based Kolmogorov Arnold networks: A path to physics-informed KANs,''  
\textit{arXiv preprint} arXiv:2502.00280, 2025.

\bibitem{Kratsios25}
A.~Kratsios and T.~Furuya,  
``Kolmogorov-Arnold Networks: Approximation and learning guarantees for functions and their derivatives,''  
\textit{arXiv preprint} arXiv:2504.15110, 2025.

\bibitem{DeVore88}
R.~A.~DeVore and V.~A.~Popov,  
``Interpolation of approximation spaces,''  
in \textit{Constructive Theory of Functions} (Varna, 1987).  
Sofia: Publ.\ House Bulgar.\ Acad.\ Sci., 1988, pp.~110--119.

\bibitem{Zhang25_general}
X.~Zhang and H.~Zhou,  
``Generalization bounds and model complexity for Kolmogorov–Arnold networks,''  
\textit{arXiv preprint} arXiv:2410.08026, 2024.

\bibitem{Li25_lipschitz}
P.~Li, L.~Ding, J.~Fu, G.~Wang, and Y.~Yuan,  
``Generalization Bounds for Kolmogorov-Arnold Networks (KANs) and Enhanced KANs with Lower Lipschitz Complexity,''  
in \textit{Advances in Neural Information Processing Systems}, NeurIPS~2025.



\bibitem{Rigas25_init}
S.~Rigas, D.~Verma, G.~Alexandridis, and Y.~Wang,  
``Initialization schemes for Kolmogorov-Arnold networks: An empirical study,''  
\textit{arXiv preprint} arXiv:2509.03417, 2025.  
 \url{https://github.com/srigas/KAN_Initialization_Schemes}

\bibitem{Muller09}
S.~Müller and R.~Schaback,  
``A Newton basis for kernel spaces,''  
\textit{Journal of Approximation Theory}, vol.~161, no.~2, pp.~645--655, 2009.

\bibitem{Han06}
F.~Han and D.-S.~Huang,  
``Improved extreme learning machine for function approximation by encoding a priori information,''  
\textit{Neurocomputing}, vol.~69, no.~16, pp.~2369--2373, 2006.

\bibitem{Santin16}
G.~Santin and R.~Schaback,  
``Approximation of eigenfunctions in kernel-based spaces,''  
\textit{Advances in Computational Mathematics}, vol.~42, no.~4, pp.~973--993, 2016.

\bibitem{Attouri25}
K.~Attouri, M.~Mansouri, and A.~Kouadri,  
``Adaptive PolyKAN-based autoencoder for fault detection and classification in wind and solar power systems,''  
\textit{Ain Shams Engineering Journal}, vol.~17, no.~1, 103884, 2026.


\bibitem{Yuan2026}
S. Yuan, Y. Liu, X. Zhang, X. Yan, H. Qin, and N. Akhtar,  
``SP-KAN: Sparse-sine perception Kolmogorov–Arnold networks for infrared small target detection,''  
\textit{ISPRS Journal of Photogrammetry and Remote Sensing}, vol. 234, pp. 1--19, 2026.  
\url{https://github.com/xdFai}

\bibitem{LiC2026}
T. Li, C. Sun, Z. Zhao, T. Liu, X. Chen, and R. Yan,  
``A Kolmogorov–Arnold-Informed Interpretable Graph Wavelet Activation Network for Machine Fault Diagnosis,''  
\textit{IEEE Transactions on Systems, Man, and Cybernetics: Systems}, vol. 56, no. 3, pp. 1693--1705, 2026.  
\url{https://github.com/HazeDT/GWAN}


\bibitem{Wen2026}
Z. Wen, Q. Zhang, J. Chen, et al.,  
``Computing-in-memory architecture for Kolmogorov-Arnold networks based on tunable Gaussian-like memory cells,''  
\textit{Nature Communications}, 2026.  

\bibitem{Ta2026}
H.-T. Ta, D.-Q. Thai, A. B. S. Rahman, G. Sidorov, and A. Gelbukh,  
``FC-KAN: Function combinations in Kolmogorov-Arnold networks,''  
\textit{Information Sciences}, vol. 736, p. 123103, 2026.  
\url{https://github.com/hoangthangta/FC_KAN}

\bibitem{Cerardi2026}
N. Cerardi, E. Tolley, and A. Mishra,  
``Solving the cosmological Vlasov–Poisson equations with physics-informed Kolmogorov–Arnold networks,''  
\textit{Monthly Notices of the Royal Astronomical Society}, vol. 545, no. 4, 2026.  

\bibitem{Cerardi2026}
N. Cerardi, E. Tolley, and A. Mishra,  
``Solving the cosmological Vlasov–Poisson equations with physics-informed Kolmogorov–Arnold networks,''  
\textit{Monthly Notices of the Royal Astronomical Society}, vol. 545, no. 4, 2026.  

\bibitem{Liu2026_PI}
Z. Liu, Y. Yuan, B. Luo, X. Xu, and S. Dubljevic,  
``Efficient operation strategy for solar membrane distillation process based on the coupled physics-informed Kolmogorov–Arnold network,''  
\textit{Desalination}, vol. 627, p. 120028, 2026.  


\bibitem{Liu2026_TS}
L. Liu, C. Zhu, Z. Zhao, Y. Ma, D. Liu, and F. Liu,  
``A physics-informed Temporal–Spatial gated Kolmogorov–Arnold network for real-time response prediction of floating structures,''  
\textit{Ocean Engineering}, vol. 353, part 2, p. 124767, 2026.  

\bibitem{Liu2026}
J. Liu, F. Yang, and K. Yan,  
``Multi-Scale Data Fusion and AdaptiveLoss Kolmogorov–Arnold Network for multivariate time series forecasting,''  
\textit{Information Fusion}, vol. 125, p. 103454, 2026.  

\bibitem{Han2026}
Z. Han, W. Xia, W. Shen, Q. Zhu, H. Liu, and C. Zhang,  
``Simulation-to-real transfer learning for bearing fault diagnosis across working conditions: A hybrid approach combining physical modeling and data-driven techniques,''  
\textit{Advanced Engineering Informatics}, vol. 69, part C, p. 103998, 2026.  

\bibitem{Wang2026}
D. Wang, R. Chai, Q. Zhou, X. Yang, D. Hao, C. Zhou, and J. Cui,  
``Two stages framework based on foundation model and Kolmogorov Cross-Domain Decoupled Network for spinal disease analysis,''  
\textit{Biomedical Signal Processing and Control}, vol. 115, p. 109412, 2026.  


\bibitem{Xu2026}
X. Xu and C. Fu,  
``Quantum-driven neural network with masked self-attention for multi-modal driving fatigue detection,''  
\textit{Engineering Applications of Artificial Intelligence}, vol. 171, p. 114192, 2026.  

\bibitem{Hu2026}
J. Hu, Y. Wang, R. Liu, H. Feng, and Y. Bai,  
``PI-MBKAN: Physics-Informed Multi Branch Kolmogorov–Arnold Network for high-precision Chiller Power Prediction,''  
\textit{Energy}, vol. 348, p. 140558, 2026.  


\bibitem{Menon2026_FEKAN}
S. S. Menon and A. D. Jagtap,  
``FEKAN: Feature-Enriched Kolmogorov–Arnold Networks,''  
\textit{arXiv preprint} arXiv:2602.16530, 2026.

\bibitem{Alireza2026}
A. A. Aghaei and M. A. Zaky,  
``An efficient physics-informed Kolmogorov–Arnold network approach for solving distributed order fractional differential equations,''  
\textit{Engineering Analysis with Boundary Elements}, vol. 186, p. 106709, 2026.  

\bibitem{Shamim2026}
M. A. Shamim, E. A. F. Reinhardt, T. A. Chowdhury, S. Gleyzer, and P. T. Araujo,  
``Probing quantum spin systems with Kolmogorov-Arnold neural network quantum states,''  
\textit{Physical Review B}, vol. 113, no. 4, p. 045157, 2026.  

\bibitem{Li2026}
J. Li et al.,  
``DWSen: Dual-Path Wavelet-Attention KAN for Joint Activity and Indoor Location Sensing,''  
\textit{IEEE Transactions on Mobile Computing}, 2026.  

\bibitem{Liang2026}
P. Liang, Q. Zeng, B. Liang, H. Huang, Y. Zhang, B. Pu, and J. Chen,  
``WKPNet: A novel wavelet-KAN-POLA network for medical image segmentation,''  
\textit{Biomedical Signal Processing and Control}, vol. 113, part B, p. 108988, 2026.  

\bibitem{Chao2026}
Z. Chao, X. Liu, Z. Wu, and X. Li,  
``RBF-KAN: Radial Basis Function-Kolmogorov-Arnold Networks,''  
\textit{IEEE Internet of Things Journal}, 2026.  

\bibitem{Rigas2026}
S. Rigas, T. Papaioannou, P. Trakadas, and G. Alexandridis,  
``A Dynamic Framework for Grid Adaptation in Kolmogorov-Arnold Networks,''  
\textit{arXiv preprint} arXiv:2601.18672, 2026.



\bibitem{Zheng2026}
L. N. Zheng, W. E. Zhang, L. Yue, M. Xu, O. Maennel, and W. Chen,  
``Free-knots Kolmogorov-Arnold Network: On the analysis of spline knots and advancing stability,''  
\textit{arXiv preprint} arXiv:2501.09283, 2025.  
\url{https://github.com/IcurasLW/FR-KAN}

\bibitem{Chiu2026}
S. T. Chiu, S. W. Cheung, U. Braga-Neto, C. S. Lee, and R. P. Li,  
``Free-RBF-KAN: Kolmogorov-Arnold Networks with Adaptive Radial Basis Functions for Efficient Function Learning,''  
\textit{arXiv preprint} arXiv:2601.07760, 2026.



\end{thebibliography}
\end{document}